\documentclass[10pt,twocolumn,letterpaper]{article}

\usepackage[T1]{fontenc}

\usepackage[pagenumbers]{cvpr}

\usepackage{cuted} %

\usepackage{currfile} %

\usepackage{caption} %

\DeclareMathOperator*{\argmin}{arg\,min}

\usepackage{amsthm}
\usepackage{mathtools}
\usepackage{float}
\newfloat{algorithm}{tbp}{loa}
\floatname{algorithm}{Algorithm}
\newcounter{algline}

\AtBeginDocument{\crefname{algorithm}{Algorithm}{Algorithms}\Crefname{algorithm}{Algorithm}{Algorithms}}

\newtheorem{proposition}{Proposition}
\newtheorem{lemma}{Lemma}

\AtEndPreamble{%
  \crefname{suppsec}{Suppl.}{Suppl.}\Crefname{suppsec}{Supplementary}{Supplementary}%
}

\usepackage{colortbl}
\definecolor{oursrow}{HTML}{F6E4DD}
\definecolor{pairedrow}{HTML}{E4EAF0}
\definecolor{markgrey}{HTML}{7A756F}
\definecolor{famOurs}{HTML}{C4553A}
\definecolor{famPaired}{HTML}{6F86A3}
\newcommand{\colOurs}[1]{\textcolor{famOurs}{#1}}
\newcommand{\colPaired}[1]{\textcolor{famPaired}{#1}}
\newcommand{\sd}[1]{{\scriptsize$\pm$#1}}
\newcommand{\sig}{\textsuperscript{*}}

\newcommand{\Xspace}{\mathcal{X}}          %
\newcommand{\Yspace}{\mathcal{Y}}          %
\newcommand{\Zspace}{\mathcal{Z}}          %
\newcommand{\Qs}{Q}                        %
\newcommand{\Ps}{P}                        %
\newcommand{\pushfwd}{\#}                  %

\newcommand{\Enc}[1]{E_{#1}}
\newcommand{\Dec}[1]{D_{#1}}
\newcommand{\Proj}[1]{\Pi_{\phi,#1}}       %
\newcommand{\Flow}[3]{\Phi^{#1}_{#2 \leftarrow #3}}   %
\newcommand{\Mstep}[1]{M^{Y}_{#1}}         %

\newcommand{\Loss}[1]{\mathcal{L}_{\mathrm{#1}}}
\newcommand{\LossProj}{\mathcal{L}_{\Pi}}  %
\newcommand{\MMD}{\operatorname{MMD}}

\newcommand{\Wass}{W_1}

\newcommand{\Norm}[1]{\left\lVert #1 \right\rVert}
\newcommand{\Expect}{\mathbb{E}}
\newcommand{\Reals}{\mathbb{R}}
\newcommand{\Normal}{\mathcal{N}}

\renewcommand{\dbltopfraction}{0.95}
\renewcommand{\dblfloatpagefraction}{0.95}

\usepackage{xr-hyper}
\definecolor{cvprblue}{rgb}{0.21,0.49,0.74}
\usepackage[pagebackref,breaklinks,colorlinks,allcolors=cvprblue]{hyperref}

\AtEndPreamble{\externaldocument[S-]{main}[]\externaldocument[M-]{main}[]}

\def\confName{CVPR}
\def\confYear{2027}

\title{PHANTESA: Projection Harmonizes Anatomical Non-paired Trajectories,\\
Enforcing Subject-specific Alignment for Synthesizing Missing Scans from Brain MRI}

\author{Yu Deng\\
{\tt\small yu.deng@lih.lu}
\and
Salah Ghamizi\\
{\tt\small salah.ghamizi@lih.lu}
\and
Daniele Perlo\\
{\tt\small daniele.perlo@lih.lu}
\end{tabular}\\[0.5em]\begin{tabular}[t]{c}
Georgia Kanli\\
{\tt\small georgia.kanli@lih.lu}
\and
Byung-Woo Hong\\
{\tt\small hong@cau.ac.kr}
\and
Olivier Keunen\\
{\tt\small olivier.keunen@lih.lu}
}

\hypersetup{
  pdftitle={PHANTESA: Projection Harmonizes Anatomical Non-paired Trajectories, Enforcing Subject-specific Alignment for Synthesizing Missing Scans from Brain MRI},
  pdfauthor={Yu Deng, Salah Ghamizi, Daniele Perlo, Georgia Kanli, Byung-Woo Hong, Olivier Keunen}
}

\begin{document}
\maketitle
\begin{abstract}
Synthesizing one type of scan from another is usually learned from paired scans of the same subjects, which few cohorts provide. Without pairs, a model can fit the target population but cannot decide which target scan each subject receives. Accordingly, we propose PHANTESA. It learns one time-indexed projection between two flow-based generative models, each trained on one scan type alone, held to the target population and pulled toward a pseudo-target built from the source scan and the target population's statistics alone. We apply PHANTESA where a patient's record lacks one of the scans that a clinical decision reads together, and synthesize the missing scan from the routine MRI for three scan types: amyloid PET to confirm amyloid pathology, T2-weighted MRI (T2) to delineate the edema and infiltrating tumor around a glioma's enhancing core, and a perfusion map to help distinguish treatment effects from tumor progression. We test whether each synthetic scan yields the imaging biomarker of the real scan. On T2 from BraTS, a tumor segmenter reads the whole tumor from PHANTESA's synthetic scan almost as well as from the paired models', and on perfusion from UPENN-GBM its map follows the tumor's blood volume as closely as theirs. Yet on amyloid PET from ADNI and OASIS-3, no model, paired or unpaired, reads amyloid burden from MRI. Even trained on each subject's real target, the projection stays well below the paired models on T2 and perfusion, where the limit is the projection's objective and application. On amyloid PET it comes close to them, and every model is held back because structural MRI does not show the amyloid uptake. Synthesis can therefore grow through a better projection where the routine MRI contains the information of the features that the target scan shows and through an input beyond the MRI where it does not, measured by imaging biomarkers beside image fidelity.
\end{abstract}

\section{Introduction}
\label{sec:intro}

\begin{figure*}[t]
  \centering
  \includegraphics[width=\linewidth]{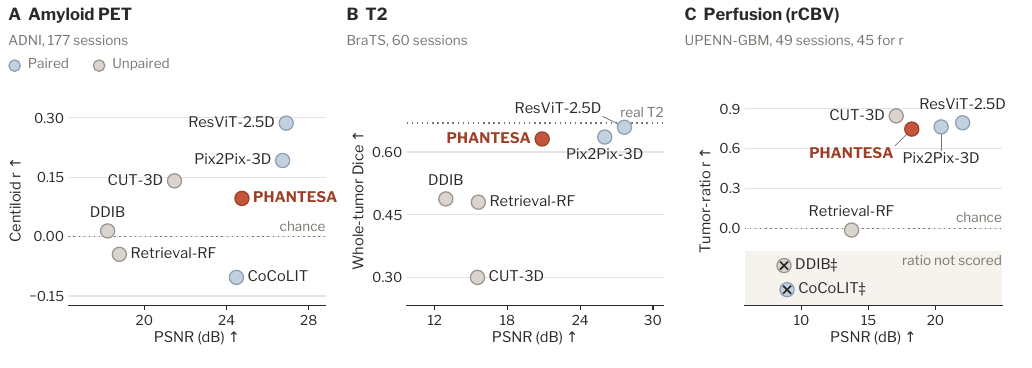}
  \caption{\textbf{Image fidelity against the analytical verification of imaging biomarkers, test set.} PHANTESA leads the unpaired models in PSNR on every task, yet PSNR does not predict the biomarkers. Dotted lines mark chance or the real scan. In C, $^{\ddagger}$ marks the two models that fail on perfusion and have no scored ratio (\cref{S-sec:sup-verification}).}
  \label{fig:teaser}
\end{figure*}

Synthesizing one type of scan from another is usually learned from paired scans of the same subjects. This paper asks whether it can be learned without pairs, for scans that a patient's record often lacks. A record lacks a scan for clinical, logistic or technical reasons. Amyloid positron emission tomography (PET) is recommended only when the amyloid status is expected to guide management~\cite{rabinovici2025auc}, some centers omit perfusion imaging from glioma protocols~\cite{thust2018glioma}, and even the standard T2-weighted (T2) sequence~\cite{ellingson2015btip} can be lost to time limits or motion~\cite{li2023brasyn}. Yet each scan answers a clinical question of its own. Amyloid PET agrees with autopsy~\cite{clark2012cerebral} and confirms the amyloid pathology that anti-amyloid therapy requires among its eligibility criteria~\cite{cummings2023lecanemab,rabinovici2025auc}. T2 shows the edema and non-enhancing tumor around a glioma's enhancing core~\cite{ellingson2015btip,deverdier2024brats}, and the relative cerebral blood volume (rCBV) of a perfusion map helps to distinguish treatment effects from tumor progression~\cite{thust2018glioma,boxerman2020dsc}. The decisions they serve combine them. Anti-amyloid therapy needs a confirmed amyloid status and a recent magnetic resonance image (MRI)~\cite{cummings2023lecanemab}, and the response of a treated IDH-mutated glioma can be assessed on contrast-enhanced T1-weighted (T1) images together with T2~\cite{wen2023rano2}, so a partial record leaves such a decision without one of its inputs. T1 and T2 name these weighted images throughout, and we call each kind of image a scan type, so T1 and T2 are two scan types of one modality.

Synthesizing the missing scan from the routine MRI meets three difficulties. First, the routine MRI need not determine the missing scan, because brains that look alike on MRI can differ in their amyloid~\cite{baron2026whenbrains}. We therefore ask, task by task, how much of the missing scan a synthetic scan recovers and what limits it (\cref{fig:teaser}). Second, synthesis models from MRI to PET and between MRI sequences train on registered pairs, whether adversarial networks after Pix2Pix~\cite{isola2017image}, transformer networks such as ResViT~\cite{dalmaz2022resvit}, diffusion and latent-diffusion models such as PASTA and CoCoLIT~\cite{li2026pasta,sargood2025cocolit}, diffusion bridges such as SelfRDB and TGH-DB~\cite{arslan2025selfrdb,pang2026tghdb} or flows such as PETFlux and Spectral Consistent Flow~\cite{yin2026petflux,li2026scflow}. Yet few cohorts acquire both scans~\cite{theodorou2025mri2pet}, limited pairs hinder large volumetric networks~\cite{li2026pasta}, and small single-center training sets limit generalization~\cite{dayarathna2024synthesis}. Third, training without pairs can use every scan of a scan type, yet it leaves each subject's target undetermined.

Without pairs, synthesis is under-constrained~\cite{zhu2017unpaired,shrestha2025diversified}, because the target population decides which outputs are admissible but not which output each subject receives. Unpaired synthesis models therefore tie each output to its own source, through cycle consistency in CycleGAN and SynDiff~\cite{zhu2017unpaired,ozbey2023unsupervised}, a patch-wise contrastive term in CUT and UNSB~\cite{park2020contrastive,kim2024unpaired}, or retrieval of similar real targets in a frozen feature space in RAFM and PRISM~\cite{zhou2026rafm,yoshai2026prism}. Other work couples the two domains inside one jointly trained model, AlignFlow, UniCycleFlow and IBCD~\cite{grover2020alignflow,zhou2026unicycleflow,lee2025ibcd}, or at one meeting point of separately trained models, DDIB and SSB~\cite{su2023dual,liu2026ssb}. Yet among these ties, one that rewards local similarity to the source, as the contrastive term does, has a copy of the source as its trivial solution. Where the target reverses the tissues' brightness order, as T2 does for T1's fluid, gray matter and white matter, such a tie pulls the output toward the source's contrast. Removing the tie does not help, because distribution matching alone hallucinates image features that can lead to misdiagnosis~\cite{cohen2018distribution}.

\begin{figure*}[t]
  \centering
  \includegraphics[width=\linewidth]{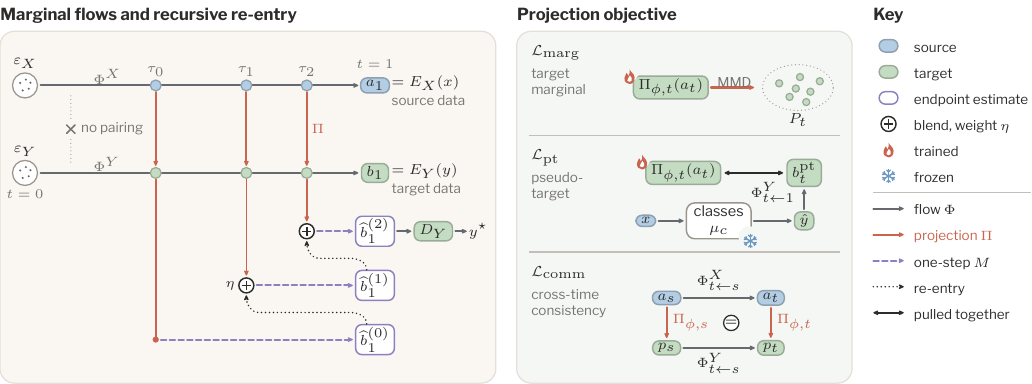}
  \caption{\textbf{PHANTESA joins two generative models, each trained on one scan type alone, through one projection trained without pairs.} Left, inference, where $\Pi$ sends the source state to the target flow and estimates re-enter with weight $\eta$. Right, the terms that train $\Pi$, against real target states ($\Loss{marg}$), to the pseudo-target ($\Loss{pt}$) and across flow times ($\Loss{comm}$).}
  \label{fig:overview}
\end{figure*}

We therefore need a tie that keeps the subject's anatomy without copying the source's contrast, and we present PHANTESA (\cref{fig:overview}). For each scan type, an autoencoder and a Rectified Flow~\cite{liu2023rectified}, both trained on that type's scans alone, carry Gaussian noise to the type's latents along a path indexed by a flow time from 0 to 1, whose points are states. A time-indexed projection moves a source state onto the target flow at the same flow time, so the two models meet at every flow time, whereas DDIB joins them at the noise alone. To tie the projection to its subject, a pseudo-target clusters the source's intensities into tissue classes and paints each class with the target population's intensity for that class. It takes only the anatomy from the source, so the tie does not copy the source's contrast. No training step uses a pair or a label, so each generative model can draw on every scan of its type.

To test this design, we compare PHANTESA with three paired and three unpaired synthesis models on amyloid PET, T2 and perfusion and test whether each prediction yields the real scan's imaging biomarker, which we call the analytical verification: Centiloid for amyloid PET, a tumor segmenter for T2, and the tumor's blood volume against its mirror region for perfusion. We also train the unpaired models on three sizes of unpaired data, to test whether they improve as their data grow (\cref{fig:scaling}).

Our contributions are:
\begin{itemize}[leftmargin=*]
  \item PHANTESA joins two flow-based generative models, each trained on one scan type alone, through a time-indexed projection trained without pairs by a population term, a pseudo-target term and a consistency term.
  \item PHANTESA leads the compared unpaired models in PSNR on every task but stays below the best paired models (\cref{fig:teaser}).
  \item In the analytical verification PHANTESA is almost level with the paired models on T2 and perfusion at a far lower PSNR, yet no model recovers amyloid burden from MRI. Image fidelity alone therefore does not show which synthetic scan can replace the real one.
  \item We locate the limit task by task, in the projection where the routine MRI contains the information of the features that the target scan shows, and in the input where it does not.
\end{itemize}

\section{Method}
\label{sec:method}

\subsection{Problem formulation}
\label{sec:formulation}

Let $\Xspace$ and $\Yspace$ be the source and target image spaces, with population distributions $\Qs_X$ and $\Ps_Y$. Source and target images are drawn independently, so none is paired. We seek a deterministic map $G:\Xspace\rightarrow\Yspace$ that sends a subject's source image to that subject's target image. Without pairs a model observes only the two populations, so the problem is under-constrained~\cite{zhu2017unpaired}. We therefore impose two constraints on $G$. The \emph{marginal constraint} is set by the two populations, and the \emph{per-subject constraint} decides each subject's output.

The marginal constraint requires
\begin{equation}
  G_{\pushfwd}\Qs_X = \Ps_Y ,
  \label{eq:marginal-constraint}
\end{equation}
where the push-forward $G_{\pushfwd}\Qs_X$ is the distribution of $G(X)$ for $X\sim\Qs_X$ (\cref{S-sec:prereq-pushforward}). It fixes the distribution of the outputs but not which output a given source receives, because composing $G$ with any map that preserves $\Ps_Y$, such as one that exchanges the outputs of two subjects, satisfies it as well (\cref{S-lem:nonident}).

The per-subject constraint needs each subject's target scan, which is unknown, so a \emph{pseudo-target} $\hat{y}(X)$ stands in for it. The pseudo-target is an image built from the source scan and the statistics of the target population alone, with the source's anatomy and the target's contrast. The per-subject constraint then selects, among the maps that satisfy \cref{eq:marginal-constraint}, the map whose output stays closest to the pseudo-target under a per-subject cost $\ell$,
\begin{equation}
  G^{\star} \in \argmin_{G\,:\,G_{\pushfwd}\Qs_X=\Ps_Y}\;
  \Expect_{X\sim \Qs_X}\!\left[\, \ell\!\left(G(X), \hat{y}(X)\right)\right].
  \label{eq:objective}
\end{equation}
Each constraint needs the other. The marginal constraint alone leaves the correspondence free, and the per-subject cost alone returns the pseudo-target itself.

To meet both constraints without pairs, PHANTESA keeps one generative model per scan type, trained on that type's scans alone, and learns a single projection between the two (\cref{fig:overview}). Only the projection sees both populations, so both constraints are imposed on it, the marginal constraint by a population term and the per-subject constraint by a pseudo-target term.

\subsection{Independent marginal flows}
\label{sec:flows}

Each scan type is modeled in a latent space by a Rectified Flow~\cite{liu2023rectified} trained by flow matching~\cite{lipman2023flow}. Its autoencoder, trained on that type's scans alone, encodes an image into $\Zspace=\Reals^{d}$ with $\Enc{X}$ or $\Enc{Y}$ and decodes it with $\Dec{X}$ or $\Dec{Y}$. With $A_1=\Enc{X}(X)\sim \Qs_1$ and $\varepsilon_X\sim\Normal(0,I)$, the source path $A_t=(1-t)\varepsilon_X+tA_1$ runs from noise at $t=0$ to the encoded scan at $t=1$. We call $t$ the flow time and the state at $t=1$ the endpoint of the path. The source velocity $u_\alpha$ minimizes
\begin{equation}
  \Loss{X}(\alpha)=\Expect\!\left[\Norm{u_\alpha(A_t,t)-(A_1-\varepsilon_X)}_2^2\right].
  \label{eq:loss-x}
\end{equation}
The target velocity $v_\theta$ is trained in the same way and independently, on the path $B_t=(1-t)\varepsilon_Y+tB_1$ with $B_1=\Enc{Y}(Y)\sim \Ps_1$. A trained velocity defines an ordinary differential equation whose solution matches the distribution of the path at every flow time (\cref{S-sec:prereq-rf}). We write $\Flow{X}{t}{s}$ and $\Flow{Y}{t}{s}$ for the flow maps that integrate the two equations from time $s$ to time $t$, and $\Qs_t$ and $\Ps_t$ for the distributions of $A_t$ and $B_t$. A subject's source state at time $t$ is $a_t=\Flow{X}{t}{1}(\Enc{X}(x))$, distributed as $\Qs_t$, and a target state is $b_t=\Flow{Y}{t}{1}(\Enc{Y}(y))$, distributed as $\Ps_t$.

\begin{figure}[t]
  \centering
  \includegraphics[width=\linewidth]{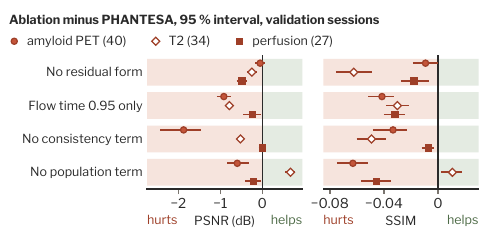}
  \caption{\textbf{Ablations of the projection's training.} Each lowers SSIM on every task but the population term's on T2. No residual form trains $\Proj{t}(a)=f_\phi(a,t)$; flow time 0.95 only trains at 0.95.}
  \label{fig:ablations}
\end{figure}

\begin{figure*}[t]
  \centering
  \includegraphics[width=0.49\linewidth]{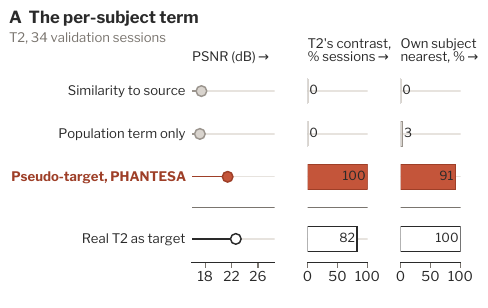}\hfill
  \includegraphics[width=0.49\linewidth]{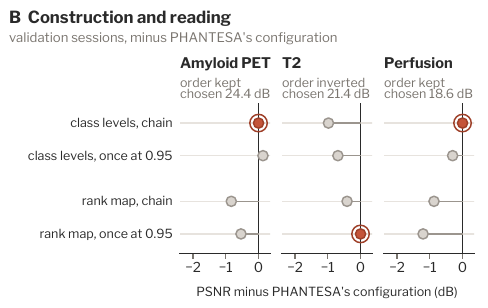}
  \caption{\textbf{The per-subject term and the pseudo-target's construction.} Only the pseudo-target term gives an output its subject's anatomy in T2's contrast (A), and the class order picks its construction (B). A, each row trains the projection with one per-subject term, the last with the subject's real T2 as a reference. The columns give the PSNR and the shares of sessions whose output has T2's order of tissue brightness and correlates best with its own subject's real T2. B, each mark is one construction of the pseudo-target under one reading of the projection, as its PSNR minus that of PHANTESA's choice (ringed). The chain reads three projections with re-entry.}
  \label{fig:pseudo}
\end{figure*}

\subsection{Same-time projection and the marginal constraint}
\label{sec:projection}

The projection $\Proj{t}:\Reals^{d}\rightarrow\Reals^{d}$ carries a source state $a_t\sim\Qs_t$ to a target state $p_t=\Proj{t}(a_t)$ at the same flow time, and its states must be distributed as the real target states at that time,
\begin{equation}
  \left(\Proj{t}\right)_{\pushfwd}\Qs_t = \Ps_t ,
  \label{eq:pushforward-condition}
\end{equation}
which, with exact flows and decoder, gives the marginal constraint of \cref{eq:marginal-constraint} (\cref{S-prop:marginal}). It is the residual map $\Proj{t}(a)=a+f_\phi(a,t)$, with $f_\phi$ one UNet~\cite{ronneberger2015unet} with 3D convolutions for all flow times, which takes $t$ as an embedding. The residual form starts every $\Proj{t}$ at the identity, which satisfies \cref{eq:pushforward-condition} as $t\rightarrow 0$, where the flows share a Gaussian base, and without it SSIM falls (\cref{fig:ablations}).

\paragraph{Population term.}
\label{sec:marginal}
$\Loss{marg}$ imposes \cref{eq:pushforward-condition} through the maximum mean discrepancy (MMD)~\cite{gretton2012kernel}. The MMD is a kernel distance between sample sets that vanishes when their distributions agree (\cref{S-sec:prereq-mmd}). The term compares the $n$ projected states of a training step at flow time $t$ with $m$ real target states, which are states of target training scans, computed once before training at twenty fixed flow times. The step reads the real states at the time $\bar{t}$ nearest to $t$ (\cref{S-sec:sup-reservoir}),
\begin{equation}
  \Loss{marg} = \Expect_t\left[
  \MMD^2\!\left(\{\Proj{t}(a^i_t)\}_{i=1}^{n},\,\{b^j_{\bar{t}}\}_{j=1}^{m}\right)\right].
  \label{eq:loss-marg}
\end{equation}

\begin{figure*}[t]
  \centering
  \includegraphics[width=\linewidth]{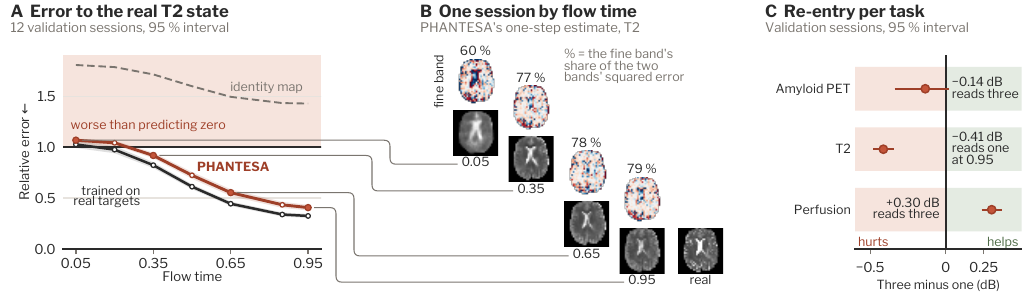}
  \caption{\textbf{The projection reaches the real target state only late in the flow, where the remaining error is mostly fine detail.} A, the projected state's squared distance to the real target state on T2 over that state's squared norm, which is 1 for a projection that returns zero. The lower curve trains the projection on real targets. B, one session's error at four flow times above its estimate. The fine band is the part of the error that a blur of 3 voxels removes. C, PHANTESA's PSNR with three projections minus its PSNR with one at flow time 0.95, labeled with each task's reading.}
  \label{fig:flowtime}
\end{figure*}

\subsection{The pseudo-target and the per-subject constraint}
\label{sec:pseudo}

The usual per-subject term in unpaired synthesis compares the output with the subject's own source scan~\cite{park2020contrastive}. The source itself minimizes such a comparison, so the output keeps the source's contrast where the target reverses the tissues' brightness order (\cref{fig:pseudo}A). PHANTESA instead compares the output with the pseudo-target.

Clustering the intensities inside the brain mask of a source $x$ gives each voxel $r$ one of three tissue classes $c(r)$, from darkest to brightest, which in T1 are roughly fluid, gray matter and white matter. The same clustering of target training scans gives each target class $c$ its mean intensity $\mu_c$ and standard deviation $\sigma_c$. The class order $\gamma$ assigns each source class its target class, and so records whether tissue that is brighter in the source is brighter or darker in the target. It is one fact per task about the two scan types' contrast, checked on 27 to 40 paired validation sessions (\cref{S-sec:sup-pseudo-build}). As SynthSeg paints each label of a segmentation with an intensity drawn for it~\cite{billot2023synthseg}, the pseudo-target paints each voxel with the intensity of its target class,
\begin{equation}
  \hat{y}(x)(r) = \Bigl[g_{w} \ast \bigl(\mu_{\gamma(c(\cdot))}
     + \sigma_{\gamma(c(\cdot))}\,\xi(\cdot)\bigr)\Bigr](r),
  \label{eq:pseudo-target}
\end{equation}
with noise $\xi\sim\Normal(0,I)$ and a Gaussian blur $g_w$. We call this construction the painted class levels. Where $\gamma$ inverts the order, the pseudo-target instead maps each voxel by the rank of its intensity, so that the brightest source voxel takes the darkest target intensity. This is the histogram matching reviewed for MRI intensity normalization~\cite{shinohara2014statistical}, with the order reversed,
\begin{equation}
  \hat{y}(x)(r) = \bigl[g_{w} \ast F_{Y}^{-1}\bigl(1-F_{x}(x(\cdot))\bigr)\bigr](r),
  \label{eq:pseudo-map}
\end{equation}
with $F_x$ the cumulative distribution of the source's intensities and $F_Y^{-1}$ the quantile function of the target training scans' intensities. We call this construction the rank map (\cref{fig:pseudo}B). In both constructions a lesion's voxels take the level that their intensity gives them.

The pseudo-target is encoded once, $b^{\mathrm{pt}}_1(x)=\Enc{Y}(\hat{y}(x))$, and the frozen target flow carries it to the sampled time, $b^{\mathrm{pt}}_t(x)=\Flow{Y}{t}{1}(b^{\mathrm{pt}}_1(x))$. The pseudo-target term draws the projected state toward this state,
\begin{equation}
  \Loss{pt} = \Expect_{t,x}\left[
  \frac{\Norm{\Proj{t}(a_t)-b^{\mathrm{pt}}_t(x)}_2^2}{\Norm{b^{\mathrm{pt}}_t(x)}_2^2}\right],
  \label{eq:loss-pt}
\end{equation}
where dividing by the squared norm of the pseudo-target state gives the term one scale at every flow time. On T2, only with this term does the output have the target's contrast and lie nearest its own subject (\cref{fig:pseudo}A).

\begin{figure*}[t]
  \centering
  \includegraphics[width=\linewidth]{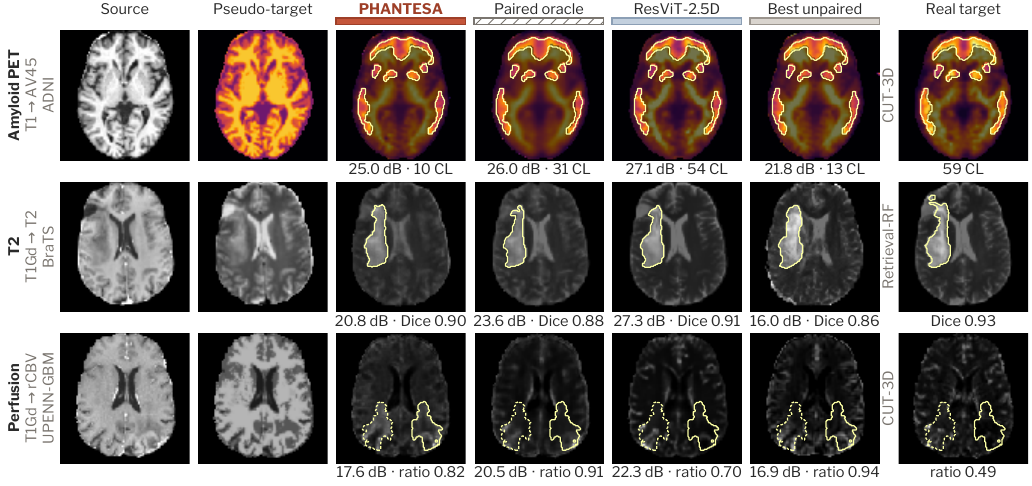}
  \caption{\textbf{One test session per task, with the region each biomarker is read from.} PHANTESA keeps the subject's anatomy in the target's contrast, and the judge segments its T2 tumor about as well as ResViT-2.5D's at a far lower PSNR. The brain outside the region is darkened. On T2 it is the expert label on the real target and the judge's segmentation on a prediction; on perfusion it is the tumor (solid) and its mirror region (dashed). Tiles give PSNR and the biomarker value. The paired oracle trains PHANTESA's projection on real targets.}
  \label{fig:qualitative}
\end{figure*}

\subsection{Cross-time consistency}
\label{sec:consistency}

Inference combines projections of one source at several flow times (\cref{fig:flowtime}C), and projections learned independently at each time could send it to different endpoints. We therefore require the family $\{\Proj{t}\}$ to be induced by one map between the encoded scans (\cref{S-eq:induced-projection}), which is equivalent to the commutation relation
\begin{equation}
  \Proj{t}\circ\Flow{X}{t}{s} = \Flow{Y}{t}{s}\circ\Proj{s}
  \label{eq:commutation}
\end{equation}
for all $0<s\le t\le 1$. Projecting before or after advancing along the flows then gives the same state, so a projection at any time reaches the same endpoint. PiFM likewise makes a flow over two time parameters independent of their order~\cite{tellez2026pifm}. The consistency term imposes \cref{eq:commutation} at sampled times $s<t$, which bounds the disagreement between endpoints reached from different times (\cref{S-sec:sup-commute}),
\begin{align}
  \Loss{comm} = \Expect_{s<t,\,a_s}
  \Big\lVert &\Proj{t}\!\left(\Flow{X}{t}{s}(a_s)\right) \nonumber\\
             &- \Flow{Y}{t}{s}\!\left(\Proj{s}(a_s)\right)\Big\rVert_1 .
  \label{eq:loss-comm}
\end{align}

\paragraph{Projection objective.}
\label{sec:objective}
The three terms combine into
\begin{equation}
  \boxed{\;
  \LossProj{} = \lambda_{\mathrm{marg}}\Loss{marg}
               + \lambda_{\mathrm{pt}}\Loss{pt}
               + \lambda_{\mathrm{comm}}\Loss{comm}\;} .
  \label{eq:loss-projection}
\end{equation}
No term uses a source and a target of the same subject.

\subsection{Inference by recursive projected re-entry}
\label{sec:reentry}
\label{sec:inference}

Inference projects a subject's source state onto the target flow and then needs the endpoint of the target path from that state. From a target state $b$ at time $t$, the one-step estimate
\begin{equation}
  \Mstep{t}(b) = b + (1-t)\,v_\theta(b,t)
  \label{eq:onestep}
\end{equation}
equals the conditional expectation of the endpoint at the optimal velocity. At $t=0$ it returns the population mean (\cref{S-sec:prereq-onestep}), and a bound on its error falls as $(1-t)^2$ toward $t=1$ (\cref{S-sec:sup-contraction}).

This makes the flow time at which inference projects a trade-off (\cref{fig:flowtime}). Near $t=0$ both state distributions are close to the shared Gaussian, so their distributions are easiest to match but the one-step estimate is close to the population mean. Near $t=1$ the estimate is sharp but the two scan types' states differ most. A schedule can therefore project first at a low flow time and recover detail by re-entry, which re-noises the endpoint estimate to a later flow time, blends in that time's projection and estimates the endpoint again.

For a source $x$ and an increasing schedule $0<\tau_0<\cdots<\tau_{K-1}<1$, write $a_{\tau}(x)=\Flow{X}{\tau}{1}(\Enc{X}(x))$ and $p_{\tau}(x)=\Proj{\tau}(a_{\tau}(x))$ for its source and projected states. The recursion starts from $\widehat{b}_1^{(0)}=\Mstep{\tau_0}(p_{\tau_0}(x))$. Each later step $k=1,\dots,K-1$ re-noises the current endpoint estimate to time $\tau_k$, blends it with the projected state at $\tau_k$, and takes the one-step estimate from the blend:
\begin{align}
  \overline{b}^{(k)}_{\tau_k} &= \tau_k\,\widehat{b}^{(k-1)}_1 + (1-\tau_k)\,\varepsilon_Y,
    \quad \varepsilon_Y\sim\Normal(0,I), \label{eq:reentry-stoch}\\
  \tilde{b}^{(k)}_{\tau_k} &= (1-\eta_k)\,\overline{b}^{(k)}_{\tau_k} + \eta_k\,p_{\tau_k}(x),
    \label{eq:relax}\\
  \widehat{b}^{(k)}_1 &= \tilde{b}^{(k)}_{\tau_k} + (1-\tau_k)\,v_\theta\!\left(\tilde{b}^{(k)}_{\tau_k},\tau_k\right),
    \label{eq:update}
\end{align}
with $0<\eta_k<1$. The output is $y^{\star}=\Dec{Y}(\widehat{b}^{(K-1)}_1)$. Each draw re-noises with fresh noise, and the prediction is the mean of eight draws, which stay as near the target population as one projection (\cref{S-tab:sup-marginal}). The schedule is chosen per task on validation sessions (\cref{fig:flowtime}C), and $K=1$ is one deterministic projection.

\section{Experiment settings}
\label{sec:setup}
\label{sec:experiments}

\paragraph{Tasks and data.} Each of the three tasks synthesizes a target scan type from a source scan type of the routine MRI. The tasks are amyloid PET from T1-weighted MRI, and T2 and the rCBV map of perfusion imaging from contrast-enhanced T1. To train without pairs, each task divides its subjects into two disjoint groups. The source subjects give the unpaired models only their source scans and the target subjects only their target scans, so an unpaired model never sees both scans of one subject. A session is one imaging visit. On amyloid PET the two groups are 345 subjects of ADNI~\cite{jack2008adni} with 635 T1 sessions and 316 subjects of OASIS-3~\cite{lamontagne2019oasis3} with 422 PET sessions. On T2 they are the subjects of the GE sites and of the Siemens sites of the BraTS 2024 post-treatment glioma cohort~\cite{deverdier2024brats}, with 656 and 587 sessions, and on perfusion two disjoint halves of UPENN-GBM~\cite{bakas2022upenn}, with 189 sessions each. The paired models train on the sessions that hold both scans, from the source subjects on amyloid PET and from both groups elsewhere. Source subjects held out from all training give sessions with both scans, which form the validation sessions and the test set of 177, 60 and 49 sessions. The main paper reports the benchmark on the test set and each design choice on validation sessions, and the Supplement repeats the benchmark on the independent test set of held-out target subjects (\cref{S-sec:sup-data}). Each volume is cropped to its brain mask, resampled to a $128^3$ grid and mapped to $[-1,1]$.

\paragraph{Evaluation.} The primary evaluation is the analytical verification of each scan type's imaging biomarker, which tests whether a synthetic scan yields the real scan's value. Amyloid PET is read as Centiloid~\cite{klunk2015centiloid,navitsky2018standardization}, compared with the real PET by the Pearson correlation $r$. For T2, the synthetic scan replaces the real T2 among the four sequences read by a tumor segmenter~\cite{myronenko2019segresnet}, which we call the segmentation judge, and the other three stay real. Its whole-tumor Dice against the expert label is compared with the Dice it reaches with the real T2. Perfusion is read as the tumor's mean blood volume over that of its mirror region across the mid-sagittal plane, compared with the real map by the Pearson correlation of this ratio. Image fidelity is SSIM~\cite{wang2004image} and PSNR inside the brain mask. One pipeline scores every model on the same sessions with paired tests and Holm correction~\cite{holm1979simple} (\cref{S-sec:sup-settings}).

\paragraph{Baselines.} Three baselines are trained on paired scans. ResViT-2.5D~\cite{dalmaz2022resvit} is a residual vision-transformer network that synthesizes one slice at a time, and Pix2Pix-3D~\cite{isola2017image} a conditional adversarial network on volumes. CoCoLIT~\cite{sargood2025cocolit} is a latent diffusion model conditioned on the MRI. Its training failed on perfusion, where its autoencoders never left the mean image. Its T2 results are in the Supplement alone (\cref{S-sec:sup-baselines}). The other three train without pairs, on PHANTESA's data. CUT-3D~\cite{park2020contrastive} is a contrastive adversarial synthesis model on volumes. Retrieval-RF, which we build for this benchmark, trains a conditional Rectified Flow~\cite{liu2023rectified} on pseudo-pairs, each source matched to the target scan nearest in age, sex and head size. DDIB~\cite{su2023dual} carries the source latent to noise with PHANTESA's source flow and back with its target flow, and trains nothing. The paired models use both scans of every training session, so they train on twice as many scans as the unpaired models on T2 and perfusion and 1.2 times on amyloid PET (\cref{S-sec:sup-data}).

\paragraph{Unpaired training data at three sizes.} Training without pairs can use any scan, so a scaling experiment tests whether the unpaired models improve as their data grow (\cref{fig:scaling}). It trains PHANTESA, CUT-3D and Retrieval-RF from scratch at three sizes, runs DDIB on PHANTESA's flows of each size and scores each on the same test set. The middle size is the benchmark's training data, so PHANTESA at this size is the model of every other result. The smallest keeps a random third of the subjects of each group. The largest adds the additional subjects, who belong to the same datasets and to neither group. Perfusion has no additional subjects, so it has two sizes. On amyloid PET the additional subjects are ADNI's, so the largest size adds PET scans of the test set's cohort to the target scans, which grow from 141 to 422 and 839 sessions (\cref{S-sec:sup-scaling-settings}).

\paragraph{Implementation details.}
\label{sec:training}
The class order inverts the tissue contrast on T2 and keeps the source's on amyloid PET and perfusion (\cref{S-sec:sup-pseudo-build}), so the pseudo-target is the rank map on T2 and the painted class levels elsewhere. With the autoencoders and flows frozen, the projection minimizes \cref{eq:loss-projection} for 2,000 steps with $\lambda_{\mathrm{marg}}=\lambda_{\mathrm{pt}}=\lambda_{\mathrm{comm}}=1$. Inference reads one deterministic projection at flow time 0.95 on T2. Elsewhere it reads three at $\tau=(0.35,0.65,0.85)$ with $\eta_1=\eta_2=0.5$ and averages eight draws (\cref{S-sec:sup-training}).

\section{Results and discussion}
\label{sec:results}

\paragraph{PHANTESA leads the unpaired models in PSNR, but PSNR does not predict the biomarkers.} Among the unpaired models PHANTESA has the highest PSNR on every task, by 1.2 to 5.3\,dB, and it stays 2.2 to 6.8\,dB below the best paired model (\cref{fig:teaser}; \cref{S-tab:benchmark}). Yet on T2 the segmentation judge reaches a whole-tumor Dice of 0.63 with PHANTESA's scan, against 0.67 with the real T2, 0.66 and 0.64 with the paired models' and 0.30 to 0.49 with the other unpaired models'. On perfusion PHANTESA's blood-volume ratio correlates with the real one at $r=0.75$. The tests separate it neither from the paired models at 0.79 and 0.76 nor from CUT-3D at 0.85, whose PSNR is 1.2\,dB below PHANTESA's. On amyloid PET no model's Centiloid $r$ exceeds 0.29 (\cref{S-sec:sup-benchmark}). Image fidelity therefore does not tell which synthetic scan can replace the real one.

\begin{figure}[t]
  \centering
  \includegraphics[width=\linewidth]{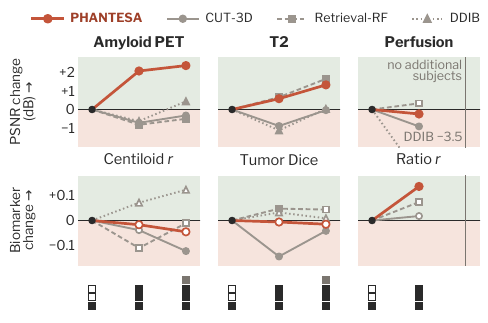}
  \caption{\textbf{Unpaired training data at three sizes, test set.} PHANTESA's PSNR rises with more data on amyloid PET and T2, and of its biomarkers only the perfusion ratio does. Lines show the change from a third of the subjects; a filled marker is separated from the next smaller size.}
  \label{fig:scaling}
\end{figure}

\paragraph{More unpaired data raise PHANTESA's PSNR on two tasks.} Over the three sizes PHANTESA's PSNR rises from 22.7 to 24.7 and 25.0\,dB on amyloid PET and from 20.3 to 20.9 and 21.6\,dB on T2, and the tests separate every step (\cref{fig:scaling}; \cref{S-sec:sup-scaling}). On perfusion the PSNR falls by 0.2\,dB. Of PHANTESA's biomarkers only the perfusion ratio rises, from $r=0.61$ to 0.75 on the test set, whereas Centiloid $r$ and the whole-tumor Dice do not.

\paragraph{No model recovers a biomarker whose signal the routine MRI does not display.} The tasks differ in how much of a biomarker's signal the source displays. Contrast-enhanced T1 outlines the tumor in tissue classes whose brightness order T2 reverses, and the tumor's enhancing part marks where the blood volume is high (\cref{S-sec:prereq-scantypes}), so a synthetic scan recovers both whole-tumor biomarkers. By contrast, how high the blood volume is inside the enhancing tumor is displayed less directly. There the ratio's $r$ is 0.63 and 0.56 for the paired models against at most 0.37 for the unpaired on the test set, yet at 41 sessions the tests separate the paired models from neither PHANTESA nor CUT-3D (\cref{S-sec:sup-benchmark}). Structural MRI does not display amyloid uptake at all. The Centiloid of PHANTESA's scans accordingly has a standard deviation of 8 on the test set, against 19 and 27 for the two best paired models' and 54 for the real scans. Another input is needed where the MRI does not display a biomarker's signal.

\paragraph{Ablations.} On T2, without the pseudo-target the population term returns one population-average image for every subject, and a term that rewards similarity to the source returns the source's contrast. Neither output lies nearest its own subject's real T2 in more than 3\% of validation sessions, against 91\% with the pseudo-target, so there the pseudo-target ties an output to its subject (\cref{fig:pseudo}A). On amyloid PET and perfusion the output without it lies nearest its own subject about as often as with it, yet loses 1.3 and 3.0\,dB (\cref{S-sec:sup-selector}). Used as the prediction, the pseudo-target image comes close to PHANTESA on both whole-tumor biomarkers (\cref{S-sec:sup-benchmark}). The projection adds 1.0 to 1.6\,dB over this image on amyloid PET and perfusion, where the painted class levels hold no detail within a class, and none on T2 (\cref{S-sec:sup-pseudo-info}). Three further choices set the fidelity. First, the class order picks the construction, because the painted class levels score 0.8 to 0.9\,dB higher where the target keeps the source's order and the rank map 0.7\,dB higher where T2 inverts it (\cref{fig:pseudo}B). Second, the residual form, the consistency term and training over all flow times each hold fidelity, as removing one lowers SSIM by up to 0.06 on every task (\cref{fig:ablations}). Without the consistency term the mismatch of \cref{eq:loss-comm} ends training 1.4 to 1.6 times as high. The population term draws the outputs to the target population, because without it they lie farther from it on every task and amyloid PET's mean Centiloid falls by 25. Its removal lowers SSIM by 0.05 to 0.06 on amyloid PET and perfusion, yet raises PSNR by 0.7\,dB on T2 (\cref{S-sec:sup-ablation}). Third, the projection reaches the real target state only late in the flow (\cref{fig:flowtime}A), so re-entry can help only where one projection at flow time 0.95 is far from it (\cref{S-sec:sup-oracle}). Against that one projection, re-entry with three gains 0.3\,dB on perfusion, ties on amyloid PET and costs 0.4\,dB on T2, so T2 uses one projection (\cref{fig:flowtime}C; \cref{S-sec:sup-reading}).

\paragraph{Most of PHANTESA's PSNR shortfall lies in the projection.} PHANTESA can at most reach the PSNR of a real target scan passed through the target autoencoder. Of its distance to this limit, 4.3 to 6.4\,dB opens in the projection, over 70\% on every task, whereas the generative models lose less than 0.05\,dB (\cref{S-fig:sup-ledger}), so the projection is the part to improve. The paired oracle of \cref{fig:qualitative} trains the projection on each subject's real target in place of the pseudo-target, and its PSNR is 1.1 to 2.5\,dB above PHANTESA's. The oracle stays 1.0\,dB below the best paired model on amyloid PET. There the paired models read no amyloid burden either, so only an input beyond the MRI can help. On T2 and perfusion it stays 4.3 and 1.8\,dB below the best paired model, which recovers the biomarkers. The remaining gap thus lies in the projection's objective and application (\cref{S-tab:sup-oracle}).

\paragraph{Limitations.} The pseudo-target is engineered for brain scans, from tissue classes and a class order, which limits PHANTESA's use on other anatomy. Nor is it optimized, because it holds no texture within a class and no lesion that the source intensities do not separate. Future work should learn the pseudo-target, add data and clinical tasks such as CT from MRI, and have clinicians verify synthetic scans on trial data.

\section{Conclusion}
\label{sec:conclusion}

PHANTESA synthesizes a missing scan from the routine MRI without pairs, through a projection between per-scan-type flow models that a pseudo-target ties to the subject. It leads the compared unpaired models in PSNR, and its imaging biomarkers are almost level with the paired models' on T2 and perfusion, yet no model reads amyloid burden from MRI. The limit lies in the projection's objective and application where the routine MRI contains the information of the target's features, and in the input where it does not, and progress on either front should be measured by imaging biomarkers beside image fidelity.

\section*{LLM Usage Statement}
\label{sec:llm}

We used Claude (Anthropic), a large language model, as a writing and engineering assistant. In the writing, it edited and rewrote the text of the paper and the supplementary material from the authors' drafts and outlines, checked statements and numbers against the experiment outputs, and checked the bibliography against the cited papers. In the research, it helped write the model code, port the baselines and run the experiments, and it wrote and revised the scripts that build the figures and tables from the experiment outputs. It worked interactively under the authors' supervision, and the authors made and double-checked every critical decision and every reading of a result. The authors designed the study and take full responsibility for the paper.

\section*{Acknowledgments}

This work was supported by the Luxembourg National Research Fund (FNR), grant C24/\allowbreak BM/\allowbreak 18940711/\allowbreak \mbox{PREDIPET-AD}.

Data collection and sharing for the Alzheimer's Disease Neuroimaging Initiative (ADNI) is funded by the National Institute on Aging (National Institutes of Health Grant U19 AG024904). The grantee organization is the Northern California Institute for Research and Education. In the past, ADNI has also received funding from the National Institute of Biomedical Imaging and Bioengineering, the Canadian Institutes of Health Research, and private sector contributions through the Foundation for the National Institutes of Health (FNIH) including generous contributions from the following: AbbVie, Alzheimer's Association; Alzheimer's Drug Discovery Foundation; Araclon Biotech; BioClinica, Inc.; Biogen; Bristol-Myers Squibb Company; CereSpir, Inc.; Cogstate; Eisai Inc.; Elan Pharmaceuticals, Inc.; Eli Lilly and Company; EuroImmun; F.~Hoffmann-La Roche Ltd and its affiliated company Genentech, Inc.; Fujirebio; GE Healthcare; IXICO Ltd.; Janssen Alzheimer Immunotherapy Research \& Development, LLC.; Johnson \& Johnson Pharmaceutical Research \& Development LLC.; Lumosity; Lundbeck; Merck \& Co., Inc.; Meso Scale Diagnostics, LLC.; NeuroRx Research; Neurotrack Technologies; Novartis Pharmaceuticals Corporation; Pfizer Inc.; Piramal Imaging; Servier; Takeda Pharmaceutical Company; and Transition Therapeutics.

Data were provided in part by OASIS-3: Longitudinal Multimodal Neuroimaging: Principal Investigators: T.~Benzinger, D.~Marcus, J.~Morris; NIH P30 AG066444, P50 AG00561, P30 NS09857781, P01 AG026276, P01 AG003991, R01 AG043434, UL1 TR000448, R01 EB009352. AV-45 doses were provided by Avid Radiopharmaceuticals, a wholly owned subsidiary of Eli Lilly.

{
    \small
    \bibliographystyle{ieeenat_fullname}
    \bibliography{main}
}

\clearpage

\appendix
\counterwithin*{equation}{section}
\renewcommand{\theequation}{\Alph{section}.\arabic{equation}}
\setcounter{figure}{0}
\setcounter{table}{0}
\renewcommand{\thetable}{S\arabic{table}}
\renewcommand{\thefigure}{S\arabic{figure}}
\renewcommand{\thelemma}{S\arabic{lemma}}
\renewcommand{\theproposition}{S\arabic{proposition}}
\def\theHequation{supp.\Alph{section}.\arabic{equation}}
\def\theHtable{supp.\arabic{table}}
\def\theHfigure{supp.\arabic{figure}}
\def\theHlemma{supp.\arabic{lemma}}
\def\theHproposition{supp.\arabic{proposition}}

\crefname{suppsec}{Sec.}{Secs.}\Crefname{suppsec}{Section}{Sections}

\setcounter{dbltopnumber}{2}
\renewcommand{\dbltopfraction}{0.95}
\renewcommand{\dblfloatpagefraction}{0.95}
\setcounter{topnumber}{2}
\setcounter{bottomnumber}{0}
\setcounter{totalnumber}{2}
\renewcommand{\topfraction}{0.5}
\renewcommand{\floatpagefraction}{0.95}
\makeatletter
\let\supp@addtodblcol\@addtodblcol
\def\@addtodblcol{%
  \ifx\@dbltoplist\@empty
    \supp@addtodblcol
    \ifx\@dbltoplist\@empty\else
      \global\@dbltoproom\dimexpr0.5\paperheight-\ht\@currbox-\dblfloatsep\relax
    \fi
  \else
    \supp@addtodblcol
  \fi}
\def\supp@capcolroom{%
  \ifx\@dbltoplist\@empty\else
    \ifx\@toplist\@empty
      \@tempdima\dimexpr0.5\paperheight-\textheight+\@colht-\textfloatsep\relax
      \ifdim\@tempdima<\@toproom \global\@toproom\@tempdima\fi
    \fi
  \fi}
\let\supp@addtocurcol\@addtocurcol
\def\@addtocurcol{\supp@capcolroom\supp@addtocurcol}
\let\supp@addtotoporbot\@addtotoporbot
\def\@addtotoporbot{\supp@capcolroom\supp@addtotoporbot}
\makeatother

\maketitlesupplementary
\sloppy

\section{Preliminaries}
\label[suppsec]{sec:prereq}

\subsection{Notation and state space}
\label[suppsec]{sec:prereq-notation}

Let $\Xspace$ and $\Yspace$ be the source and target image spaces, for example T1-weighted MRI volumes and amyloid PET volumes. Source distributions are written $\Qs$ and target distributions $\Ps$, with subscripts for the space or time they live on. Encoders $\Enc{X}\colon\Xspace\to\Zspace$, $\Enc{Y}\colon\Yspace\to\Zspace$ and decoders $\Dec{X},\Dec{Y}$ map each image space to and from $\Zspace=\Reals^{d}$. The two autoencoders are trained separately on separate data, so equal dimensionality carries no shared meaning of coordinates. A given coordinate of $\Reals^{d}$ is a different feature for each scan type.

\subsection{Push-forward and non-identifiability}
\label[suppsec]{sec:prereq-pushforward}

For a measurable map $G$ and a distribution $\Qs$, the push-forward $G_{\pushfwd}\Qs$ is the law of $G(X)$ for $X\sim\Qs$, and push-forwards compose one map at a time,
\begin{equation}
  (R\circ G)_{\pushfwd}\Qs = R_{\pushfwd}\!\left(G_{\pushfwd}\Qs\right).
  \label{eq:composition}
\end{equation}
The proof of the target marginal applies \cref{eq:composition} at each stage of PHANTESA's pipeline.

Composing a map that has the correct output distribution with any map that preserves this distribution gives another map with the correct output distribution.

\begin{lemma}
Let $G:\Xspace\rightarrow\Yspace$ satisfy $G_{\pushfwd}\Qs_X=\Ps_Y$ and let a self-map $R:\Yspace\rightarrow\Yspace$ of the target space satisfy $R_{\pushfwd}\Ps_Y=\Ps_Y$. Then $R\circ G$ also satisfies $(R\circ G)_{\pushfwd}\Qs_X=\Ps_Y$.\label{lem:nonident}
\end{lemma}

\begin{proof}
By \cref{eq:composition}, $(R\circ G)_{\pushfwd}\Qs_X=R_{\pushfwd}(G_{\pushfwd}\Qs_X) =R_{\pushfwd}\Ps_Y=\Ps_Y$.
\end{proof}

For example, take two source and two target samples with $\Qs_X(x_1)=\Qs_X(x_2)=\tfrac12$ and $\Ps_Y(y_1)=\Ps_Y(y_2)=\tfrac12$. Both $G_1$, sending $x_1\mapsto y_1$ and $x_2\mapsto y_2$, and $G_2$, sending $x_1\mapsto y_2$ and $x_2\mapsto y_1$, produce the correct target marginal. Letting $R$ swap $y_1$ and $y_2$ gives $G_2=R\circ G_1$ with $R_{\pushfwd}\Ps_Y=\Ps_Y$, so the marginal constraint cannot distinguish them.

The same holds beyond two samples. By \cref{lem:nonident}, every self-map of the target space that preserves $\Ps_Y$ generates another map with the correct marginal, and for a rich target space that family is enormous. Hence the marginal constraint fixes only that the outputs collectively follow the target distribution, and an additional criterion is required to select which output each subject receives. PHANTESA supplies that criterion as the per-subject pseudo-target.

\subsection{Rectified Flow and flow maps}
\label[suppsec]{sec:prereq-rf}

Both flows are Rectified Flow models~\cite{liu2023rectified} trained by flow matching~\cite{lipman2023flow}. For a data sample $Z_1\sim p_1$ and an independent $\varepsilon\sim p_0=\Normal(0,I)$, each uses the linear path
\begin{equation}
  Z_t = (1-t)\,\varepsilon + t\,Z_1,\qquad t\in[0,1],
  \label{eq:rf-path}
\end{equation}
and trains a velocity network $w_\vartheta$ by
\begin{equation}
  \Loss{RF}(\vartheta) =
  \Expect\!\left[\,\Norm{w_\vartheta(Z_t,t)-(Z_1-\varepsilon)}_2^2\right],
  \label{eq:rf-loss}
\end{equation}
whose minimizer is the conditional mean velocity
\begin{equation}
  w^{\star}(z,t)=\Expect\!\left[Z_1-\varepsilon\mid Z_t=z\right].
  \label{eq:sup-optimum}
\end{equation}
The solution of the ODE of $w^{\star}$ started from noise has the distribution of $Z_t$ at every $t$~\cite{liu2023rectified}.

PHANTESA relies on this agreement of distributions at every flow time, and on no other property of the flows. The agreement leaves open which noise sample belongs to which scan, so a flow delivers the distributions $\Qs_t$ and $\Ps_t$ alone. By \cref{lem:nonident} the marginal constraint cannot identify which target belongs to which source either. The projection $\Proj{t}$ therefore has to supply this correspondence.

For an ODE $\dot z_\tau=w(z_\tau,\tau)$ we write $\Flow{}{t}{s}(z_s)=z_t$ for integration from time $s$ to time $t$. Where solutions are unique, $\Flow{}{r}{t}\circ\Flow{}{t}{s}=\Flow{}{r}{s}$ and $\Flow{}{s}{t}=(\Flow{}{t}{s})^{-1}$. We write $\Flow{X}{t}{s}$ and $\Flow{Y}{t}{s}$ for the flow maps of the source and target models.

\subsection{One-step endpoint estimate}
\label[suppsec]{sec:prereq-onestep}

Inference takes the endpoint of the target path from a state at an intermediate flow time in one step. Let $B_t=(1-t)\varepsilon_Y+tB_1$ be the target path, with velocity $v_\theta$ and population optimum $v^{\star}$. Then $B_1=B_t+(1-t)(B_1-\varepsilon_Y)$, and taking the conditional expectation given $B_t=b$ with \cref{eq:sup-optimum} at the population optimum gives $\Expect[B_1\mid B_t=b]=b+(1-t)v^{\star}(b,t)$~\cite{liu2023rectified}. Under $v_\theta$ this is the estimate $\Mstep{t}(b)=b+(1-t)v_\theta(b,t)$ of \cref{M-eq:onestep}, exact when $v_\theta=v^{\star}$.

At $t=0$ the estimate carries no subject-specific content, because $B_0=\varepsilon_Y$ is independent of $B_1$ during training, so $v^{\star}(\varepsilon_Y,0)=\Expect[B_1]-\varepsilon_Y$ and hence
\begin{equation}
  \Mstep{0}(\varepsilon_Y)=\varepsilon_Y+v^{\star}(\varepsilon_Y,0)=\Expect[B_1].
  \label{eq:sup-degenerate}
\end{equation}
One step from pure noise therefore returns the population mean of the target latents, with no individual variation. Integrating the source flow backward from a subject's latent does give a noise sample that is specific to that subject. Yet the target velocity is trained without pairs, so it gives that sample no meaning at $t=0$. Accordingly, the subject reaches the target flow only through the projection at later flow times and through re-entry.

\subsection{Maximum mean discrepancy}
\label[suppsec]{sec:prereq-mmd}

The population term compares the projected states with real target states by the unbiased estimate of the squared maximum mean discrepancy~\cite{gretton2012kernel}, which for a characteristic kernel, such as a Gaussian, is zero in expectation exactly when the two distributions agree. The kernel is a sum of Gaussians,
\begin{equation*}
  k(a,b)=\sum_{c\in\{1/4,1/2,1,2,4\}}\exp\!\left(-\Norm{a-b}_2^2/(2c^2\sigma^2)\right),
\end{equation*}
where $\sigma$ follows the median heuristic of~\cite{gretton2012kernel}, as the median distance between points of both samples together, and is held constant when the gradient is taken. In training each estimate compares the 8 projected states of a step with the 128 real target states drawn for it.

\subsection{Wasserstein-1 distance and contracting kernels}
\label[suppsec]{sec:prereq-wasserstein}

The convergence of re-entry is stated in the Wasserstein-1 distance $\Wass$ between distributions with a finite first moment. A Markov kernel $K$ with $\Wass(\mu K,\nu K)\le\rho\,\Wass(\mu,\nu)$ for some $\rho<1$ has exactly one invariant distribution $\pi$, and $\Wass(\mu K^{n},\pi)\le\rho^{n}\,\Wass(\mu,\pi)$ from any initial $\mu$~\cite{ollivier2009ricci}.

\subsection{The scan types}
\label[suppsec]{sec:prereq-scantypes}

The three tasks use two source scan types. T1-weighted MRI is the source for amyloid PET, and contrast-enhanced T1 (T1Gd) is the source for T2 and for perfusion. T1-weighted MRI shows anatomy. After gadolinium contrast it also shows enhancing tumor tissue as hyperintense signal and necrosis or cysts as dark regions. T2-weighted MRI shows the surrounding non-enhancing hyperintensity, which includes edema, infiltrating tumor, gliosis and radiation-related change~\cite{deverdier2024brats}.

Amyloid PET with florbetapir measures cortical tracer uptake and is read for amyloid status~\cite{clark2012cerebral}. Tracer burden is not observed by structural MRI, which sees anatomy that correlates with it.

Dynamic susceptibility contrast (DSC) perfusion MRI yields the relative cerebral blood volume (rCBV). UPENN-GBM provides it as a map that its authors call an automated proxy of the rCBV~\cite{bakas2022upenn}. Contrast enhancement and tumor geometry locate high perfusion but do not quantify it.

\section{Details of the method}
\label[suppsec]{sec:sup-training}

\subsection{Training data without pairs}
\label[suppsec]{sec:regimes}

Because no term of the objective uses a source scan and a target scan of the same subject, the training data may take in any subject for whom either scan type exists, across cohorts, sites and scanners. Paired acquisitions are scarce and concentrated in a few studies, whereas each scan type alone is abundant. Each flow can therefore train on every available scan of its type, whereas a paired model is limited to the sessions that hold both scans. The source and target subjects are disjoint. They come from different datasets on amyloid PET, from different site groups of one dataset on T2 and from two halves of one dataset on perfusion. The pairs a cohort provides score the held-out sessions and train the paired models.

Target volumes differ by scanner, tracer and reconstruction pipeline, so a target distribution $\Ps_Y$ drawn from many sites is a \emph{mixture} over acquisition conditions. \Cref{M-eq:marginal-constraint} reproduces that mixture, site effects included. The target flow takes no site or scanner as a condition, which keeps its training data as large as possible. A synthetic scan therefore carries the acquisition characteristics of the whole set of target scans rather than those of one chosen site. For example, the T2 target scans come from the Siemens sites of BraTS, and a synthetic T2 follows the mixture of those sites.

\subsection{Autoencoders and flows}
\label[suppsec]{sec:sup-generative}

\paragraph{Autoencoders.} The two autoencoders are trained separately, each on its own scan type, under an intensity reconstruction loss. The encoder outputs a mean and a variance for every latent value, and the latent of a scan is this mean divided by the standard deviation of that scan type's training latents, which is one scalar per scan type. The flows run in this latent space because a velocity field over a full volume needs too much memory at voxel resolution, and an autoencoder that downsamples each axis by a factor $f$ reduces the number of grid positions by $f^3$. The latent keeps a few channels on this coarser grid, so the flows and the projection take the latent of a whole volume at once. Every term of \cref{M-eq:loss-projection} compares latent states, and the pseudo-target image is encoded once before training, so training the projection never calls a decoder. Setting the encoders and decoders to the identity gives the voxel-space form of the method, for which every statement of this section holds unchanged.

\paragraph{Flows.} \Cref{M-eq:loss-x} and its target counterpart are minimized independently for 30,000 steps. Inference and the projection's training integrate the flows from $t=1$ back to intermediate flow times, so the numerical integration must not alter the latents. A round trip integrates a target latent from $t=1$ back to flow time 0.35 and forward again, with four Heun steps on each of the chain's three intervals in each direction (\cref{sec:sup-inference}). On the target scans of every task and both test sets it lowers the PSNR of the decoded scan by at most 0.014\,dB on average and by at most 0.054\,dB in any session, so the integration adds no measurable error.

\subsection{Proof of the target marginal}
\label[suppsec]{sec:sup-marginal}

The synthesis map returns scans whose distribution equals the target distribution when the following assumptions hold. The assumptions are that the projection meets the push-forward condition at the flow time at which it is read, that both flows transport their distributions exactly, and that the target decoder maps the latent distribution onto the image distribution. Composing the projection with the two flows and the target decoder gives the synthesis map
\begin{equation}
  G_{\phi,t} = \Dec{Y}\circ\Flow{Y}{1}{t}\circ\Proj{t}\circ\Flow{X}{t}{1}\circ\Enc{X},
  \label{eq:generator}
\end{equation}
which reads the projection at the flow time $t$.

\begin{proposition}[Target marginal]
If \cref{M-eq:pushforward-condition} holds, both flows transport their marginals exactly, and the decoder satisfies $(\Dec{Y})_{\pushfwd}\Ps_1=\Ps_Y$, then $(G_{\phi,t})_{\pushfwd}\Qs_X = \Ps_Y$.\label{prop:marginal}
\end{proposition}

\begin{proof}[Proof of \cref{prop:marginal}]
Let $X\sim \Qs_X$. Encoding gives $\Enc{X}(X)\sim \Qs_1$; transporting along the source flow gives $\Flow{X}{t}{1}(\Enc{X}(X))\sim \Qs_t$; the push-forward condition \cref{M-eq:pushforward-condition} gives $\Proj{t}(\Flow{X}{t}{1}(\Enc{X}(X)))\sim \Ps_t$; transporting along the target flow gives a sample of $\Ps_1$; and the decoder assumption gives $\Ps_Y$. Written with \cref{eq:composition} applied at each stage,
\begin{align}
  (G_{\phi,t})_{\pushfwd}\Qs_X
   &= (\Dec{Y})_{\pushfwd}(\Flow{Y}{1}{t})_{\pushfwd}(\Proj{t})_{\pushfwd}
      \nonumber\\
   &\qquad (\Flow{X}{t}{1})_{\pushfwd}(\Enc{X})_{\pushfwd}\Qs_X \nonumber\\
   &= (\Dec{Y})_{\pushfwd}(\Flow{Y}{1}{t})_{\pushfwd}
      (\Proj{t})_{\pushfwd}\Qs_t \nonumber\\
   &= (\Dec{Y})_{\pushfwd}(\Flow{Y}{1}{t})_{\pushfwd}\Ps_t \nonumber\\
   &= (\Dec{Y})_{\pushfwd}\Ps_1 \nonumber\\
   &= \Ps_Y .
  \label{eq:sup-chain}
\end{align}
\end{proof}

The proposition concerns the distribution of all outputs and says nothing about which output a given subject receives, which the pseudo-target term decides. Trained flows and decoders are approximate, so the equality holds up to their error.

\subsection{Building the pseudo-target}
\label[suppsec]{sec:sup-pseudo-build}

The pseudo-target of \cref{M-eq:pseudo-target} is built once per training source, at data preparation. Three-cluster k-means on the source's intensities inside its brain mask gives the tissue classes $c(r)\in\{0,1,2\}$, numbered from darkest to brightest, and the class order $\gamma$ sends each to its target class. The target levels $\mu_c$ and $\sigma_c$ are the means and standard deviations of the same clustering applied to 64 target training scans, averaged over them, so they come from target scans alone. Each voxel takes its class's level with Gaussian noise at the class's standard deviation, drawn once, and the image is blurred by a Gaussian of width $w=0.6$ voxel. This is the form of SynthSeg's generator~\cite{billot2023synthseg}, which draws each voxel from the Gaussian of its label and blurs the result. SynthSeg samples the Gaussians' parameters at random for each batch, whereas the pseudo-target uses the measured target levels. It is then set to the target background outside the mask and clipped to $[-1,1]$. A tumor's voxels take the class that their source intensity falls in. The image is encoded once by the target autoencoder, and at every step the frozen target flow carries its latent to the sampled time with four steps of Heun's method, without gradient.

The class levels come from the target training scans alone. The class order is the one fact per task that the pseudo-target needs beyond them. It states how the two scan types' contrasts relate, which is knowledge of the scan types and of no subject, and it is checked on the task's validation sessions, 34 on T2, 40 on amyloid PET and 27 on perfusion. For each source class the real target's mean intensity inside that class is averaged over these sessions, and the class takes the target rank of that mean. On T2 the means fall from the darkest source class to the brightest, so the order inverts, with fluid brightest and white matter darkest, as the known contrast of the two sequences predicts. On amyloid PET the means rise from $-0.18$ through $-0.03$ to $0.16$, so the order of T1 is kept, and it holds in all 40 sessions. On perfusion the means are $-0.04$, $0.00$ and $0.14$, so the order of contrast-enhanced T1 is kept, because its brightest class holds enhancing tumor and vessels, whose blood volume is highest. Across the 27 sessions the brightest class lies above the darkest in 85\% and all three classes follow the kept order in 37\%, against one in six for a random order. The two darker classes lie within 0.04 of each other on the $[-1,1]$ scale, so exchanging them barely changes the pseudo-target. Because T2's order is a monotone inversion, its pseudo-target is mapped from the intensity ranks (\cref{M-eq:pseudo-map}) with no added noise. Its target quantile function takes 20,000 voxels from each of the same 64 target training scans and keeps 1,001 quantiles. On the other tasks the pseudo-target is painted from the class levels.

\subsection{Cross-time consistency}
\label[suppsec]{sec:sup-commute}

\paragraph{The induced family commutes.} A family of projections that all come from one map between the encoded scans satisfies the commutation relation of \cref{M-eq:commutation}. Let $T$ send each encoded source scan to an encoded target scan. The family $\{\Proj{t}\}$ is induced by $T$ when
\begin{equation}
  \Proj{t} = \Flow{Y}{t}{1}\circ T\circ\Flow{X}{1}{t}.
  \label{eq:induced-projection}
\end{equation}
so the projection at time $t$ carries the source state to the end of the source flow, applies $T$ and returns along the target flow to time $t$. Substituting \cref{eq:induced-projection} and using the composition rule $\Flow{}{r}{t}\circ\Flow{}{t}{s}=\Flow{}{r}{s}$ of the flow maps,
\begin{align}
  \Proj{t}\circ\Flow{X}{t}{s}
    &= \Flow{Y}{t}{1}\circ T\circ\Flow{X}{1}{t}\circ\Flow{X}{t}{s}
       \nonumber\\
    &= \Flow{Y}{t}{1}\circ T\circ\Flow{X}{1}{s},
  \label{eq:sup-commute-lhs}
\end{align}
while
\begin{align}
  \Flow{Y}{t}{s}\circ\Proj{s}
    &= \Flow{Y}{t}{s}\circ\Flow{Y}{s}{1}\circ T\circ\Flow{X}{1}{s}
       \nonumber\\
    &= \Flow{Y}{t}{1}\circ T\circ\Flow{X}{1}{s}.
  \label{eq:sup-commute-rhs}
\end{align}
The two agree, establishing \cref{M-eq:commutation}.

\paragraph{Commutation implies endpoint consistency.} Applying \cref{M-eq:commutation} with $t=1$ gives $\Flow{Y}{1}{t}\circ\Proj{t}=\Proj{1}\circ\Flow{X}{1}{t}$, so for any source endpoint $a_1$,
\begin{align}
  \Flow{Y}{1}{t}\circ\Proj{t}\circ\Flow{X}{t}{1}(a_1)
    &= \Proj{1}\circ\Flow{X}{1}{t}\circ\Flow{X}{t}{1}(a_1) \nonumber\\
    &= \Proj{1}(a_1),
  \label{eq:sup-endpoint}
\end{align}
independent of $t$. Projection at any time reaches the same endpoint. The first identity also gives $\Proj{t}=\Flow{Y}{t}{1}\circ\Proj{1}\circ\Flow{X}{1}{t}$, which is \cref{eq:induced-projection} with $T=\Proj{1}$, so a commuting family is induced by one map and the two statements are equivalent.

\paragraph{Local form and a residual bound.} Differentiating $b_t=\Proj{t}(a_t)$ along a source trajectory with $\dot a_t=u_\alpha(a_t,t)$, and requiring $b_t$ to follow the target flow, yields the differential form of \cref{M-eq:commutation}, $\partial_t\Proj{t}(a) + J_a\Proj{t}(a)\,u_\alpha(a,t) = v_\theta(\Proj{t}(a),t)$, with $J_a\Proj{t}$ the Jacobian in the state argument. Defining the residual
\begin{equation}
  r_\phi(a,t) = \partial_t\Proj{t}(a) + J_a\Proj{t}(a)u_\alpha(a,t)
                - v_\theta(\Proj{t}(a),t),
  \label{eq:sup-residual}
\end{equation}
A Lipschitz constant of a map is the largest factor by which the map lengthens the distance between two points. Write $L_{v,t}$ for the Lipschitz constant of the target velocity $v_\theta(\cdot,t)$ at flow time $t$, and $L_v$ for its largest value over $t$. Gr\"onwall's inequality then gives
\begin{align}
  &\Norm{\Flow{Y}{1}{t}(\Proj{t}(a_t)) - \Proj{1}(a_1)} \nonumber\\
  &\qquad \le \int_t^{1} e^{L_v(1-s)}\Norm{r_\phi(a_s,s)}\,ds .
  \label{eq:sup-gronwall}
\end{align}
Reducing the local residual therefore reduces the disagreement between the endpoints that projections at different times reach. The bound is informative when the target velocity changes slowly with the state, because its factor grows exponentially with $L_v$. The consistency term reaches this residual without the Jacobian. Expanding both sides of \cref{M-eq:loss-comm} to first order in $t-s$ shows that they differ by $(t-s)\,r_\phi(a_s,s)$ plus terms of order $(t-s)^2$, so comparing the two states at sampled times $s<t$ penalizes the residual through one integration of each flow and two evaluations of the projection.

\subsection{Training the projection}
\label[suppsec]{sec:sup-projection-training}

Both flows are frozen and the projection's weights $\phi$ alone are updated. Each step samples 8 source latents and two flow times $s<t$ in $[0.05,0.95]$. It forms the source states and the pseudo-target states with four steps of Heun's method and the two branches of the consistency term with two steps each. It then minimizes \cref{M-eq:loss-projection} with $\lambda_{\mathrm{marg}}=\lambda_{\mathrm{pt}}=\lambda_{\mathrm{comm}}=1$, for 2,000 steps at learning rate $10^{-4}$ with mixed precision.

\paragraph{Real target states.}\label[suppsec]{sec:sup-reservoir} Real target states need no gradient, so they are computed once before training rather than at every step. The flow time is divided into twenty slots of width $0.05$, and the target flow carries every target training latent to the center $t=(k+0.5)/20$ of each slot $k$, which gives one table of real target states per slot. A step at flow time $t$ draws $128$ states without replacement from the slot that contains $t$, and the population term of \cref{M-eq:loss-marg} compares them with the step's 8 projected states. Because the 128 real states far outnumber the 8 projected states, the estimate varies mainly through the projected states, and drawing more real states would not steady it.

\subsection{Contraction of the re-entry recursion}
\label[suppsec]{sec:sup-contraction}

The re-entry step of \cref{M-eq:reentry-stoch} returns an endpoint estimate to an earlier flow time along the training path, so that the projection can be applied again. The recursion applies the projection at $K$ increasing flow times $0<\tau_0<\cdots<\tau_{K-1}<1$, and these flow times form its schedule (\cref{M-sec:reentry}). The recursion converges when recursions started from different endpoint estimates approach one distribution of endpoints. Write $F_\tau=\Flow{Y}{1}{\tau}$ for the target flow map from flow time $\tau$ to the endpoint and $p_\tau=p_\tau(x)$ for the projected state of a source $x$.

\paragraph{Contraction.} The re-entry returns the estimate to the distribution on which the target velocity was trained. If $\widehat{b}_1\sim\Ps_1$, then $\overline{b}_\tau\sim\Ps_\tau$, because $\varepsilon$ is drawn afresh at each step and is independent of $\widehat{b}_1$. The step costs no network evaluation, and its fresh noise gives a different output at each draw. At fixed $\tau$ and $\eta$ the step is random, so it defines a Markov kernel $\mathcal{K}_{x,\tau}$ rather than a deterministic operator. Consider two recursions started at $\widehat{b}_1$ and $\widehat{b}_1'$ and driven by the same noise draw $\varepsilon$, a construction known as synchronous coupling. By \cref{M-eq:reentry-stoch} the noise cancels in their difference, so $\Norm{\overline{b}_\tau-\overline{b}_\tau'}=\tau\Norm{\widehat{b}_1-\widehat{b}_1'}$. By \cref{M-eq:relax} the common $p_\tau(x)$ cancels, so $\Norm{\tilde{b}_\tau-\tilde{b}_\tau'}=(1-\eta)\tau\Norm{\widehat{b}_1-\widehat{b}_1'}$. By \cref{M-eq:update}, $\Norm{\Mstep{\tau}(\tilde{b}_\tau)-\Mstep{\tau}(\tilde{b}_\tau')} \le\left[1+(1-\tau)L_{v,\tau}\right]\Norm{\tilde{b}_\tau-\tilde{b}_\tau'}$, with $L_{v,\tau}$ the Lipschitz constant of the target velocity at flow time $\tau$ (\cref{sec:sup-commute}). Composing the three gives the contraction factor
\begin{equation}
  \rho_\tau = (1-\eta)\,\tau\left[1+(1-\tau)L_{v,\tau}\right].
  \label{eq:rho-stoch}
\end{equation}
When $\rho_\tau<1$ the coupling bounds the Wasserstein-1 distance,
\begin{equation}
  \Wass\!\left(\mu \mathcal{K}_{x,\tau},\nu \mathcal{K}_{x,\tau}\right)
  \le \rho_\tau\,\Wass(\mu,\nu),
  \label{eq:sup-w1}
\end{equation}
so $\mathcal{K}_{x,\tau}$ contracts in $\Wass$, admits a unique invariant distribution $\pi_{x,\tau}$, and converges to it geometrically~\cite{ollivier2009ricci}. For example, at a target velocity with $L_{v,\tau}=1$ and the blend weight $\eta=0.5$ the factor is 0.44 at $\tau=0.65$ and 0.49 at $\tau=0.85$.

\paragraph{The limit of the recursion.} The recursion converges to a distribution of endpoints for each source $x$, not to a point. This distribution $\pi_{x,\tau}$ lies near the endpoint $F_\tau(p_\tau(x))$ that the exact target flow reaches from the projected state, without being centered on it. Two limiting cases show what $\eta$ does. At $\eta=1$ the noise cancels and the recursion collapses in one step to the value $\Mstep{\tau}(p_\tau(x))$, so every draw is the same image. As $\eta\rightarrow 0$ the projection is never blended in again and $\pi_{x,\tau}$ no longer depends on $x$, so every subject receives the same distribution. Diversity across draws therefore requires $0<\eta<1$, and $\eta$ trades subject specificity against spread.

\paragraph{Error under one-step updates.} Let $b^{\star}=\Proj{1}(a_1(x))$ be the endpoint the projection assigns to the source. Under \cref{M-eq:commutation}, $F_{\tau_k}(p_{\tau_k}(x))=b^{\star}$ at every flow time of the schedule. Write $\mathcal{S}_k$ for one step of the recursion at $\tau_k$, so that $\widehat{b}^{(k)}_1=\mathcal{S}_k(\widehat{b}^{(k-1)}_1)$. Then
\begin{equation}
  \Expect\Norm{\widehat{b}^{(k)}_1-b^{\star}} \le
  \rho_{\tau_k}\,\Expect\Norm{\widehat{b}^{(k-1)}_1-b^{\star}}
  + \beta_{\tau_k} + \epsilon_{\Pi,\tau_k},
  \label{eq:recurrence}
\end{equation}
where $\rho_{\tau_k}$ is the factor of \cref{eq:rho-stoch} with the blend weight $\eta_k$. The term $\beta_{\tau_k}=\Expect\Norm{\mathcal{S}_k(b^{\star})-b^{\star}}$ is how far one step moves the target when the projection is exact, and $\epsilon_{\Pi,\tau_k}$ collects the projection's marginal error and its cross-time inconsistency, through which $F_{\tau_k}(p_{\tau_k}(x))$ departs from $b^{\star}$. The displacement $\beta_{\tau_k}$ has two parts. The first is the error of the one-step estimate of \cref{M-eq:onestep}. For a state $b$ at flow time $t$, the exact endpoint is $\Flow{Y}{1}{t}(b)=b+\int_t^1 v_\theta(b_s,s)ds$ with $b_s=\Flow{Y}{s}{t}(b)$, and the one-step estimate replaces the integrand by its value at $s=t$, so it treats the velocity as constant over the remaining flow time. If the target velocity changes along the trajectory at a rate of at most $M_v$, so that $\Norm{\tfrac{d}{ds}v_\theta(b_s,s)}\le M_v$, then
\begin{align}
  &\Norm{\Flow{Y}{1}{t}(b)-\Mstep{t}(b)} \nonumber\\
  &\quad= \Norm{\int_t^1\!\left[v_\theta(b_s,s)-v_\theta(b,t)\right]ds} \nonumber\\
  &\quad\le \int_t^1 M_v(s-t)\,ds
    \;=\; \tfrac{1}{2}M_v(1-t)^2.
  \label{eq:onestep-error}
\end{align}
The error therefore falls quadratically as $t$ approaches 1. At the flow times PHANTESA uses, $(1-t)^2$ is 0.42 at 0.35, 0.02 at 0.85 and 0.0025 at 0.95, so the late estimates are the accurate ones. The second part comes from returning along the interpolation path rather than along the flow trajectory. It is random through $(1-\eta_k)(1-\tau_k)\varepsilon$ and also vanishes as $\tau_k\rightarrow1$.

The error at step $k$ therefore has three sources, which are the error carried over from the step before, the displacement $\beta_{\tau_k}$ of one step and the projection's error $\epsilon_{\Pi,\tau_k}$. The schedule's flow times act on all three. Advancing $\tau_k$ toward $1$ shrinks $\beta_{\tau_k}$, changes the factor $\rho_{\tau_k}$ that carries the error over, and moves the projection to where its error $\epsilon_{\Pi,\tau_k}$ is small, which is late in the flow (\cref{M-fig:flowtime}A). A schedule that ends late therefore refines the estimate from coarse to fine, and no schedule repairs a poorly trained projection.

\subsection{Inference}
\label[suppsec]{sec:sup-inference}

Inference takes one source scan and returns one synthetic target scan. It encodes the source and integrates the source flow backward from $t=1$ through the flow times of the schedule, with four steps of Heun's method between each flow time and the next, which gives the source states at every flow time of the schedule. The chain is the schedule $\tau=(0.35,0.65,0.85)$, so $K=3$, with blend weights $\eta=(1.0,0.5,0.5)$. The first weight is 1 because the first estimate is taken from the projected state alone. Each Heun step evaluates the source velocity twice, and the chain's backward integration takes 4 steps on each of its 3 intervals, from 1 to 0.85, from 0.85 to 0.65 and from 0.65 to 0.35, so it costs $3\times4\times2=24$ velocity evaluations. The encoder pass and this integration are computed once per session and shared by every draw. Each draw then costs $K=3$ evaluations of the projection, $K=3$ of the target velocity and one decoder pass, 7 network evaluations. Each re-entry draws new noise, so reading a session through the chain again gives another output. Each such output is a draw, and the prediction is the voxel-wise mean of eight draws, which differ in the noise of the re-entries alone, so one prediction of the chain costs $1+24+8\times7=81$ network evaluations. Each further flow time in the schedule adds 8 source-velocity evaluations and 2 evaluations per draw. The late reading is one projection at flow time $0.95$ followed by its one-step endpoint, so $K=1$ and no noise enters. It costs one encoder pass, 8 source-velocity evaluations for the 4 Heun steps from $t=1$ to $0.95$, one evaluation of the projection, one of the target velocity and one decoder pass, 12 network evaluations. Amyloid PET and perfusion are read by the chain and T2 by the late reading.

\section{Additional evidence for the design choices}
\label[suppsec]{sec:sup-design}

\begin{table*}[t]
\centering
\footnotesize
\setlength{\tabcolsep}{2.2pt}\setlength{\aboverulesep}{0pt}\setlength{\belowrulesep}{0pt}\renewcommand{\arraystretch}{1.18}
\caption{\textbf{Without a per-subject pseudo-target the output is not this subject's T2.} Each row is one training of the projection, scored on the 34 BraTS validation sessions on which the per-subject term was chosen; the \colOurs{terracotta} row is PHANTESA. The subject part, the share of sessions whose output correlates best with its own subject's T2 (own first) and the share whose brightest and darkest tissue classes take the order of a real T2 (T2 sign) are defined in \cref{sec:sup-selector}; the two shares count sessions and carry no spread. Each cell is the mean over the sessions with its standard deviation in small type. \sig\,The paired test separates the cell from PHANTESA's (\cref{sec:sup-scoring}).}
\label{tab:sup-persubject}
\begin{tabular}{l cc ccc c cc}
\toprule\addlinespace[2pt]
 & \multicolumn{2}{c}{Image fidelity} & \multicolumn{3}{c}{Judge's Dice} & & & \\
\cmidrule(lr){2-3}\cmidrule(lr){4-6}
Per-subject term & PSNR\,$\uparrow$ & SSIM\,$\uparrow$ & WT\,$\uparrow$ & TC\,$\uparrow$ & ET\,$\uparrow$ & \shortstack{Subject\\part\,$\uparrow$} & \shortstack{Own\\first} & \shortstack{T2\\sign} \\
\midrule[\heavyrulewidth]
Similarity term & 17.4\sd{0.4}\sig & 0.16\sd{0.03}\sig & 0.40\sd{0.27}\sig & 0.42\sd{0.37} & 0.50\sd{0.33} & $-$0.25\sd{0.12}\sig & 0\% & 0\% \\
No per-subject term & 17.2\sd{0.5}\sig & 0.17\sd{0.04}\sig & 0.44\sd{0.30}\sig & 0.43\sd{0.38} & 0.50\sd{0.34} & $-$0.08\sd{0.14}\sig & 3\% & 0\% \\
Similarity term, weight 0.1 & 17.8\sd{0.7}\sig & 0.16\sd{0.04}\sig & 0.38\sd{0.29}\sig & 0.41\sd{0.38} & 0.45\sd{0.36} & $-$0.23\sd{0.15}\sig & 0\% & 0\% \\
\midrule
Painted classes, chain & 20.4\sd{0.5}\sig & 0.33\sd{0.10}\sig & 0.57\sd{0.20} & 0.43\sd{0.35}\sig & 0.51\sd{0.29} & $+$0.56\sd{0.06}\sig & 85\% & 100\% \\
Rank map, chain & 21.0\sd{0.8}\sig & 0.48\sd{0.05}\sig & 0.56\sd{0.19}\sig & 0.42\sd{0.34}\sig & 0.50\sd{0.29} & $+$0.58\sd{0.06}\sig & 91\% & 100\% \\
\rowcolor{oursrow}\textbf{PHANTESA}, rank map, 0.95 & 21.4\sd{0.7} & 0.49\sd{0.05} & 0.59\sd{0.18} & 0.37\sd{0.32} & 0.50\sd{0.27} & $+$0.61\sd{0.07} & 91\% & 100\% \\
\midrule
\textit{Paired oracle} & 22.6\sd{0.7}\sig & 0.55\sd{0.05}\sig & 0.58\sd{0.19} & 0.48\sd{0.36}\sig & 0.57\sd{0.28}\sig & $+$0.63\sd{0.05}\sig & 100\% & 82\% \\
\bottomrule
\end{tabular}
\end{table*}

\subsection{The similarity variant of the per-subject term}
\label[suppsec]{sec:sup-selector}

On T2, a per-subject term that compares the output with the subject's own source returns the source's contrast, whereas the pseudo-target term returns the target's contrast in this subject's anatomy (\cref{tab:sup-persubject}). The variant studied here decodes the one-step estimate
\begin{equation}
  \widetilde{y} = \Dec{Y}\!\left[\Mstep{t}(p_t)\right]
  \label{eq:decoded-target}
\end{equation}
and compares $\widetilde{y}$ with $x$ by the patch-level contrastive loss of CUT~\cite{park2020contrastive}, with the MIND descriptor~\cite{heinrich2012mind} in place of CUT's learned features. MIND describes each voxel by the similarity of its patch to the patches of its six axis neighbors, divided by a local variance estimate, and has no trainable parameters, so the projection and the comparator cannot co-adapt. Here the patches are $3\times3\times3$, the local variance is bounded below by $\epsilon=10^{-5}$ and the volume's edge is replicated at the border. The six responses at each location are normalized to unit length in place of the published rescaling to a maximum of one. The descriptor $h^{\ell}$ is computed on the volume average-pooled by $2^{\ell}$, at the image grid ($\ell=0$) and the half grid ($\ell=1$). The loss $\Loss{sim}$ is averaged over the image grid and the half grid, at 256 locations sampled inside the source brain mask at each scale, with a temperature $\tau_c$. At each location the decoded estimate's descriptor is the query, the source's descriptor at the same location the positive and the source's descriptors at the other sampled locations the negatives, so the chance level is $\ln 256=5.55$ nats. A structural prior of this kind is a common way to specify what counts as shared across scan types in unpaired medical synthesis~\cite{phan2024structural}. A global criterion could not replace it, because for a deterministic $G$ that satisfies \cref{M-eq:marginal-constraint} the mutual information $I(X;G(X))$ equals $H(\Ps_Y)$ whatever the map, so it separates no two assignments of outputs to subjects. This variant is the one term that decodes during training.

\begin{table*}[t]
\centering
\footnotesize
\setlength{\tabcolsep}{3.0pt}\setlength{\aboverulesep}{0pt}\setlength{\belowrulesep}{0pt}\renewcommand{\arraystretch}{1.18}
\caption{\textbf{The class order decides which construction of the pseudo-target wins.} Scored on the validation sessions on which the construction was chosen: the projection trained on each construction, read through the chain; the rank map minus the painted class levels in PSNR with the sessions the map wins; and each pseudo-target image scored alone on the same sessions. Each cell is the mean over the sessions with its standard deviation in small type. \sig\,The paired test separates the rank map from the painted class levels (\cref{sec:sup-scoring}).}
\label{tab:sup-construction}
\begin{tabular}{l cc cc cc cc}
\toprule\addlinespace[2pt]
 & \multicolumn{2}{c}{Painted classes} & \multicolumn{2}{c}{Rank map} & \multicolumn{2}{c}{Map $-$ classes} & \multicolumn{2}{c}{Image alone, PSNR} \\
\cmidrule(lr){2-3}\cmidrule(lr){4-5}\cmidrule(lr){6-7}\cmidrule(lr){8-9}
Task, validation sessions & PSNR\,$\uparrow$ & SSIM\,$\uparrow$ & PSNR\,$\uparrow$ & SSIM\,$\uparrow$ & PSNR & Won & Classes & Map \\
\midrule[\heavyrulewidth]
Amyloid PET, ADNI (40) & 24.4\sd{1.2} & 0.58\sd{0.09} & 23.5\sd{1.1}\sig & 0.58\sd{0.10} & $-$0.83\sd{0.16}\sig & 0 & 22.8\sd{0.9} & 23.7\sd{1.1}\sig \\
T2, BraTS (34) & 20.4\sd{0.5} & 0.33\sd{0.10} & 21.0\sd{0.8}\sig & 0.48\sd{0.05}\sig & $+$0.57\sd{0.41}\sig & 31 & 19.7\sd{1.8} & 21.6\sd{1.0}\sig \\
Perfusion, UPENN-GBM (27) & 18.6\sd{0.7} & 0.24\sd{0.05} & 17.7\sd{0.8}\sig & 0.25\sd{0.06}\sig & $-$0.87\sd{0.28}\sig & 0 & 17.6\sd{2.1} & 18.2\sd{0.6} \\
\bottomrule
\end{tabular}
\end{table*}

The variant fails for two reasons, both measured on T2. First, its minimizer is a copy of the source. MIND compares squared differences of neighboring patches, so it is blind to the sign of contrast, and no output matches the source's descriptors better than the source itself. On the T2 validation sessions, where a lower loss is preferred, a copy of the source scores 3.9 nats, the subject's own real T2 scores 4.2 to 4.5 and another subject's T2 scores 5.1 to 5.8, around the chance level. The loss therefore prefers the copy to the subject's own T2. Where the target keeps the source's tissue order the copy has the right contrast, whereas where the target inverts it, as T2 does, the output keeps the source's contrast. Second, switching the term off leaves the population term alone, and the output is then the target population's average brain.

Every variant is scored on T2's 34 BraTS validation sessions, on which the per-subject term was chosen (\cref{sec:sup-pseudo-build}). Three scores check whether an output shows its own subject's T2 (\cref{tab:sup-persubject}). The first is the correlation of the output with its own subject's real T2 minus its mean correlation with the other subjects' real T2s, taken on the band of the image between Gaussian blurs of 2 and 4 voxels. It is positive when the output follows its own subject, and we call it the subject part. The second is the share of the 34 sessions in which the output correlates best with its own subject's T2. The third is T2's sign, the share of sessions whose output is brighter in the darkest tissue class of its source scan than in the brightest one, as T2's inverted class order requires. Each variant replaces the per-subject term of the projection that is trained on painted class levels, and every projection row of the table is read by the chain except PHANTESA's, which reads T2 once at flow time 0.95. With the similarity term the subject part is $-0.25$, so the output follows the source's contrast, which is opposite to the target's. Without a per-subject term it is $-0.08$, near zero, and the population term ends training at $-0.005$ against 0.121 for PHANTESA, so the projected states match the target population while carrying no subject. Both variants score 3.0 to 3.2\,dB lower in PSNR than the painted class levels and fall below them in every one of the 34 sessions. Neither shows T2's sign in any session, and the output correlates best with its own subject in at most 1 of 34. With either pseudo-target the output shows T2's sign in all 34 sessions and correlates best with its own subject in 29 to 31 of them. A real T2 shows T2's sign in 31 of the 34 sessions and the paired oracle (\cref{sec:sup-oracle}) in 28, whereas the pseudo-target imposes it in every session. The similarity term is trained for 20,000 steps, ten times the others, so its failure does not come from short training, and at a tenth of its weight it fails in the same way.

Both variants also lower the segmentation judge's whole-tumor Dice, by 0.14 to 0.17 against the painted class levels. The judge's score is the Dice of its segmentation against the expert label, which is twice the shared volume of two regions over the sum of their volumes. With all four sequences real, it reaches 0.65 for the whole tumor on the 34 sessions. On the 20 sessions whose expert label contains both a tumor core and an enhancing tumor, it reaches 0.52 for the tumor core and 0.63 for the enhancing tumor. On tumor core both variants stay within 0.01 of the painted class levels, at 0.42 and 0.43 against 0.43, because the judge reads the inner regions mostly from the three sequences that stay real, and a T2 without structure does not mislead it there. PHANTESA's tumor-core Dice, 0.37, is the lowest of the table, and the chain readings of either pseudo-target score 0.04 to 0.06 higher there; \cref{sec:sup-reading} weighs this against image fidelity. The per-subject pseudo-target is therefore what makes the output this subject's T2, in contrast and in anatomy.

On the two tasks whose target keeps the source's class order, removing the per-subject term gives a different result than on T2 (\cref{tab:sup-ablation}). There the output still shows its own subject and loses fidelity. On amyloid PET and perfusion this ablation sets the weight of the pseudo-target term to zero and is read through the chain. Its output correlates best with its own subject's real target in 34 of the 40 amyloid PET validation sessions and in 25 of the 27 perfusion sessions, against 36 and 23 for PHANTESA. Its PSNR is lower in every session, by 1.29\,dB on amyloid PET and by 2.98\,dB on perfusion, against 4.21\,dB on T2. The level of the biomarker moves too. On the 22 perfusion validation sessions whose ratio is read, the mean ratio of the enhancing tumor is 5.06, against 1.80 for PHANTESA and 1.99 for the real maps. On the 168 ADNI validation sessions that pass the registration checks, the mean predicted Centiloid is 19.5, against 3.0 for PHANTESA and 43.0 for the real scans. Its absolute error is 2.1 Centiloid below PHANTESA's and ties with it, because the 95\% interval of the difference runs from $-5.5$ to $1.2$. On the three tasks studied, the per-subject term is thus needed for the subject only where the class order inverts, and it holds the output's PSNR on all three.

\subsection{The construction of the pseudo-target}
\label[suppsec]{sec:sup-pseudo-info}

PHANTESA uses the rank map on T2 and the painted class levels on amyloid PET and perfusion, because on each task the projection trained on that construction scores the higher PSNR, and the order of the task's tissue classes predicts which (\cref{tab:sup-construction}, \cref{M-fig:pseudo}B). On each task the projection is trained twice with identical settings, on the painted class levels of \cref{M-eq:pseudo-target} and on the rank map of \cref{M-eq:pseudo-map}, and both are scored on the validation sessions on which the construction was chosen, 40 on amyloid PET, 34 on T2 and 27 on perfusion (\cref{sec:sup-pseudo-build}). The rank map scores higher than the painted class levels only on T2, where the target inverts the tissue order, by 0.57\,dB of PSNR through the chain (95\% interval 0.45 to 0.69) and by 0.69\,dB when both are read once at flow time 0.95. Where the order is kept it scores lower in every session, by 0.83\,dB on amyloid PET and 0.87\,dB on perfusion. In SSIM the rank map ties with the painted class levels on amyloid PET and is ahead by 0.01 on perfusion and 0.15 on T2, and the construction is chosen on PSNR, as the reading is.

The fidelity of a pseudo-target image does not predict the fidelity of the projection trained on it. Each pseudo-target image is also scored on its own, without any projection, on the same sessions. On amyloid PET the rank-map image scores 0.89\,dB higher than the painted image and on perfusion 0.53\,dB higher, a difference the test separates on amyloid PET and not on perfusion, yet on both tasks the projection trained on it scores lower. The reason is how the projection uses each image. A painted image leaves the detail inside each class unspecified, so the projection learns that detail from the population term and the frozen target flow. Accordingly, the projection trained on painted class levels scores 1.0 to 1.6\,dB above its own pseudo-target image where the order is kept, and 0.7\,dB above it on T2. A rank-map image instead fills each class with the source's own variation of intensity inside that class, rescaled to the target population's histogram. We call this variation the gradation. Its shape is right but its amplitude too wide. After each class's mean is removed, the gradation's correlation with the real target inside the classes is 0.53 on amyloid PET, 0.47 on T2 and 0.12 on perfusion, against 0.42, 0.43 and 0.02 for the painted image, whose only variation inside a class is its blurred border, and the paired test separates the two on every task. The right shape raises SSIM, and the wide amplitude lowers PSNR. Because the pseudo-target term regresses the projection onto the image, the projection copies both effects, and in PSNR it at most ties with the rank-map image. Read through the chain it ties with the image on amyloid PET and stays 0.44 and 0.58\,dB below it on perfusion and T2. On T2, read once at flow time 0.95 as PHANTESA is, it stays 0.17\,dB below the image (95\% interval 0.03 to 0.30) and 0.046 below it in SSIM, so on T2 the projection adds no PSNR to its pseudo-target image. Where the order inverts, the painted image scores 1.9\,dB below the rank-map image, and the rank map, which carries the inverted order voxel by voxel, gives the better projection under both readings.

On the three tasks studied, the class order alone therefore predicts the construction. Where the order is kept, the population term and the frozen target flow supply the detail inside the classes better than the rank map's too-wide gradation does, and where the order inverts, only the rank map carries the inverted order voxel by voxel.

\subsection{The components of the projection's training}
\label[suppsec]{sec:sup-ablation}

\begin{table*}[t]
\centering
\footnotesize
\setlength{\tabcolsep}{1.7pt}\setlength{\aboverulesep}{0pt}\setlength{\belowrulesep}{0pt}\renewcommand{\arraystretch}{1.18}
\caption{\textbf{Each part of the projection's training, removed in turn on the three tasks.} Each ablation minus PHANTESA per session, with PSNR in dB, scored on the validation sessions on which the design was chosen, with the sessions the ablation wins. Each difference is the mean over the sessions with its standard deviation in small type. \sig\,The paired test separates the difference from zero (\cref{sec:sup-scoring}).}
\label{tab:sup-ablation}
\begin{tabular}{l cccc cccc cccc}
\toprule\addlinespace[2pt]
 & \multicolumn{4}{c}{Amyloid PET, ADNI (40)} & \multicolumn{4}{c}{T2, BraTS (34)} & \multicolumn{4}{c}{Perfusion, UPENN-GBM (27)} \\
\cmidrule(lr){2-5}\cmidrule(lr){6-9}\cmidrule(lr){10-13}
Ablation $-$ PHANTESA & PSNR & Won & SSIM & Won & PSNR & Won & SSIM & Won & PSNR & Won & SSIM & Won \\
\midrule[\heavyrulewidth]
No residual form & $-$0.05\sd{0.26} & 19 & $-$0.009\sd{0.022}\sig & 16 & $-$0.25\sd{0.29}\sig & 8 & $-$0.062\sd{0.035}\sig & 1 & $-$0.48\sd{0.35}\sig & 1 & $-$0.018\sd{0.028}\sig & 6 \\
Flow time 0.95 only & $-$0.92\sd{0.41}\sig & 0 & $-$0.041\sd{0.022}\sig & 1 & $-$0.79\sd{0.38}\sig & 0 & $-$0.030\sd{0.026}\sig & 1 & $-$0.24\sd{0.57} & 12 & $-$0.032\sd{0.022}\sig & 1 \\
No consistency term & $-$1.88\sd{1.17}\sig & 0 & $-$0.033\sd{0.031}\sig & 1 & $-$0.52\sd{0.37}\sig & 0 & $-$0.049\sd{0.047}\sig & 1 & 0.00\sd{0.21} & 16 & $-$0.007\sd{0.013}\sig & 8 \\
No population term & $-$0.60\sd{0.63}\sig & 5 & $-$0.063\sd{0.028}\sig & 0 & $+$0.67\sd{0.40}\sig & 33 & $+$0.011\sd{0.022}\sig & 25 & $-$0.21\sd{0.49} & 10 & $-$0.046\sd{0.028}\sig & 1 \\
No per-subject term & $-$1.29\sd{0.38}\sig & 0 & $-$0.032\sd{0.017}\sig & 1 & $-$4.21\sd{1.09}\sig & 0 & $-$0.315\sd{0.072}\sig & 0 & $-$2.98\sd{1.01}\sig & 0 & $-$0.002\sd{0.034} & 12 \\
\bottomrule
\end{tabular}
\end{table*}

\begin{table*}[t]
\centering
\footnotesize
\setlength{\tabcolsep}{3.0pt}\setlength{\aboverulesep}{0pt}\setlength{\belowrulesep}{0pt}\renewcommand{\arraystretch}{1.18}
\caption{\textbf{On the latent the chain's draws lie as near the real target scans as one projection, or nearer, and without the population term the outputs lie farther.} Unbiased MMD$^2\times10^{3}$ to the real target scans of the target subjects' validation sessions, 35 of OASIS-3, 64 of BraTS and 26 of UPENN-GBM, on the whole latent as the population term takes it; the unbiased estimate can be negative. Each row draws from the validation sessions of one group of subjects. The split-half floor is the 95th percentile of the same distance between two random halves of those real scans, over 200 splits. One projection is read once at flow time 0.95, and \cref{prop:marginal}'s map integrates the target flow from it. Step 1 is the chain's first estimate, at flow time 0.35, step 2 its estimate after the re-entry at 0.65, and a draw its output after the re-entry at 0.85, one per session; the mean of draws averages the eight draws of each session, which differ only in the noise of their re-entries. T2 is read once and has no chain. The last column is the prediction of the projection trained without the population term, read as PHANTESA is read. A distance between two sets of sessions has no per-session values, so the table carries no standard deviation. \sig\,The permutation test separates the set from the real target scans; for the draw minus one projection and for the last column against PHANTESA's prediction, the bootstrap interval of the difference excludes zero (\cref{sec:sup-scoring}).}
\label{tab:sup-marginal}
\begin{tabular}{l c cc cccc c c c}
\toprule\addlinespace[2pt]
 & & \multicolumn{2}{c}{One projection} & \multicolumn{4}{c}{The chain} & & & \\
\cmidrule(lr){3-4}\cmidrule(lr){5-8}
Task, validation sessions & \shortstack{Split-half\\floor} & Read once & Prop.~\ref{prop:marginal} & Step 1 & Step 2 & Draw & \shortstack{Mean of\\draws} & \shortstack{Draw $-$\\read once} & \shortstack{Pseudo-\\targets} & \shortstack{No popu-\\lation term} \\
\midrule[\heavyrulewidth]
\multicolumn{11}{l}{\textbf{Amyloid PET}} \\
\quad Source subjects, ADNI (168) & 26.3 & 30.0\sig & 30.2\sig & 39.6\sig & 26.6\sig & 23.1\sig & 23.0\sig & $-$6.9\sig & 255.3\sig & 146.3\sig \\
\quad Target subjects, OASIS-3 (35) & 26.4 & $-$17.7 & $-$17.4 & $-$10.3 & $-$25.0 & $-$27.4 & $-$27.5 & $-$9.7\sig & 179.2\sig & 87.4\sig \\
\multicolumn{11}{l}{\textbf{T2, BraTS}} \\
\quad Source subjects (34) & 14.7 & 85.8\sig & 85.6\sig & -- & -- & -- & -- & -- & 52.7\sig & 146.9\sig \\
\multicolumn{11}{l}{\textbf{Perfusion, UPENN-GBM}} \\
\quad Source subjects (27) & 27.3 & 21.0 & 21.3 & 41.3\sig & 19.0 & 16.5 & 17.2 & $-$4.5\sig & 66.3\sig & 72.1\sig \\
\quad Target subjects (26) & 27.4 & $-$23.0 & $-$22.6 & 2.1 & $-$27.3 & $-$31.2 & $-$30.5 & $-$8.2\sig & 4.4 & 31.8\sig \\
\bottomrule
\end{tabular}
\end{table*}

To show what each part of the projection's training contributes, each ablation trains PHANTESA's projection without one part, on each of the three tasks, and is scored on the validation sessions on which the design was chosen (\cref{M-fig:ablations}; \cref{tab:sup-ablation}). No residual form trains $\Proj{t}(a)=f_\phi(a,t)$, flow time 0.95 only narrows the training range from all flow times to that flow time alone and reads one projection there, no consistency term sets $\lambda_{\mathrm{comm}}=0$ and no population term sets $\lambda_{\mathrm{marg}}=0$. No per-subject term sets the weight of the pseudo-target term to zero, and \cref{sec:sup-selector} discusses it. Apart from flow time 0.95 only, every ablation is read as PHANTESA is read on its task, through the chain on amyloid PET and perfusion and once at flow time 0.95 on T2. The Wilcoxon test is corrected over the five ablations of each column of \cref{tab:sup-ablation}.

Removing the residual form, the consistency term or the training range lowers SSIM on every task. The nine differences are 0.007 to 0.062, and the test separates each of them from zero. The loss of PSNR depends on the task. Flow time 0.95 only and no consistency term each lower PSNR in every session of amyloid PET and T2, by 0.5 to 1.9\,dB on average. On perfusion the test separates neither loss from zero, at 0.24\,dB for flow time 0.95 only and 0.00\,dB for no consistency term. No residual form lowers PSNR by 0.48\,dB on perfusion, in 26 of the 27 sessions, and by 0.25\,dB on T2. On amyloid PET its PSNR ties, because the difference of $-0.05$\,dB has a 95\% interval from $-0.17$ to $0.05$, and 16 of the 40 sessions have the higher SSIM without it.

Flow time 0.95 only changes the reading as well as the training. On T2 PHANTESA already reads one projection at flow time 0.95, so the 0.79\,dB come from the training range alone. On amyloid PET PHANTESA reads the chain, so the ablation is also compared with PHANTESA read once at flow time 0.95. Against that reading it loses 1.05\,dB (95\% interval 0.91 to 1.21). This is no less than the 0.92\,dB that it loses against PHANTESA read through the chain (\cref{tab:sup-ablation}), so the training range accounts for the whole loss. The chain applies the projection at three flow times at which this ablation was not trained, and read through the chain the ablation loses 2.61\,dB (2.20 to 3.11). On perfusion PHANTESA reads the chain as well. Against PHANTESA read once at flow time 0.95 the ablation ties in PSNR, with a difference of 0.07\,dB (95\% interval $-0.14$ to $0.27$), and it loses 0.014 of SSIM (0.008 to 0.022). Read through the chain, the ablation scores 0.32\,dB above PHANTESA (0.16 to 0.48) and 0.017 below it in SSIM. On perfusion the training range therefore holds SSIM and not PSNR.

Removing the population term moves the outputs away from the target population on every task, and what it does to fidelity depends on the task (\cref{tab:sup-marginal}). The distance to the target population is the one of \cref{sec:sup-chain}, taken on the prediction, in units of $10^{-3}$. Its floor is the distance that two random halves of the real scans reach. On amyloid PET it rises from 23.0 to 146.3 on the ADNI sessions and from below the floor to 87.4 on the OASIS-3 sessions. On perfusion it rises from 17.2 to 72.1 for the source subjects, where the pseudo-targets themselves lie at 66.3, and from below the floor to 31.8 for the target subjects. On T2 it rises from 85.8 to 146.9. In each of these five comparisons the 95\% interval of the difference excludes zero (\cref{sec:sup-scoring}). Fidelity falls as well on the two tasks whose pseudo-target is the painted class levels. On amyloid PET the ablation loses 0.60\,dB of PSNR and 0.063 of SSIM. On perfusion it loses 0.046 of SSIM, and the test does not separate its loss of 0.21\,dB from zero. On T2 it gains 0.67\,dB of PSNR, in 33 of the 34 sessions, and 0.011 of SSIM. On the 168 ADNI validation sessions that pass the registration checks, the mean predicted Centiloid falls from 3.0 to $-22.2$ without the term, against 43.0 for the real scans, and its absolute error rises by 21.0 (95\% interval 18.9 to 22.9). On the 22 perfusion validation sessions whose ratio is read, the overlap of the top tenth of rCBV (\cref{sec:sup-verification}) falls by 0.053 (0.030 to 0.077). The correlation of the whole tumor's ratio falls from 0.63 to 0.54 and that of the enhancing tumor's from 0.22 to 0.17, and the intervals of both differences include zero. On T2 the judge's Dice changes on no region by an amount whose interval excludes zero.

On T2 alone, removing the population term raises PSNR, and two facts set T2 apart. First, its pseudo-target is the rank map, which already carries the source's detail inside each class at the target population's histogram, whereas the painted class levels leave that detail to the population term and the frozen target flow (\cref{sec:sup-pseudo-info}). Second, T2's target subjects were scanned at the Siemens sites and its source subjects at the GE sites (\cref{tab:sup-splits}), and the real scans of the two groups differ as populations. The real T2 scans of the 34 validation sessions lie 54.5 units from the 64 real scans of the target subjects' validation sessions, against a split-half floor of 14.7. The population term therefore draws each T2 output toward the scans of the Siemens sites, whereas PSNR compares it with its own session's real scan from the GE sites. The gain of 0.67\,dB has a cost in contrast. Without the term the three tissue classes leave T2's inverted order in 4 of the 34 sessions, against none for PHANTESA. The brightest and the darkest class keep it in all 34 sessions.

\begin{table*}[t]
\centering
\footnotesize
\setlength{\tabcolsep}{4pt}\setlength{\aboverulesep}{0pt}\setlength{\belowrulesep}{0pt}\renewcommand{\arraystretch}{1.18}
\caption{\textbf{The paired oracle exceeds PHANTESA on every test set.} PHANTESA and the paired oracle on both test sets, beside the PSNR of the better of ResViT-2.5D and Pix2Pix-3D. Each cell is the mean over the sessions with its standard deviation in small type. \sig\,The paired test separates the difference from zero (\cref{sec:sup-scoring}).}
\label{tab:sup-oracle}
\begin{tabular}{l cc cc cc}
\toprule\addlinespace[2pt]
 & \multicolumn{2}{c}{PHANTESA} & \multicolumn{2}{c}{Paired oracle} & \shortstack{Oracle $-$\\PHANTESA} & \shortstack{Best\\paired} \\
\cmidrule(lr){2-3}\cmidrule(lr){4-5}
Task, test set & PSNR\,$\uparrow$ & SSIM\,$\uparrow$ & PSNR\,$\uparrow$ & SSIM\,$\uparrow$ & PSNR & PSNR \\
\midrule[\heavyrulewidth]
\multicolumn{7}{l}{\textbf{Amyloid PET}} \\
\quad Test set, ADNI (177) & 24.7\sd{1.2} & 0.59\sd{0.09} & 25.9\sd{1.2} & 0.64\sd{0.08} & $+$1.14\sd{0.34}\sig & 26.9\sd{1.3} \\
\quad Independent, OASIS-3 (76) & 25.8\sd{1.5} & 0.65\sd{0.10} & 27.2\sd{1.6} & 0.72\sd{0.10} & $+$1.34\sd{0.33}\sig & 27.1\sd{1.7} \\
\multicolumn{7}{l}{\textbf{T2, BraTS}} \\
\quad Test set (60) & 20.9\sd{0.8} & 0.45\sd{0.07} & 23.3\sd{0.7} & 0.57\sd{0.06} & $+$2.49\sd{0.90}\sig & 27.6\sd{1.6} \\
\quad Independent (60) & 21.0\sd{0.8} & 0.40\sd{0.07} & 22.1\sd{0.7} & 0.50\sd{0.06} & $+$1.16\sd{0.54}\sig & 26.3\sd{1.5} \\
\multicolumn{7}{l}{\textbf{Perfusion, UPENN-GBM}} \\
\quad Test set (49) & 18.2\sd{1.0} & 0.23\sd{0.06} & 20.2\sd{0.5} & 0.45\sd{0.04} & $+$2.02\sd{0.82}\sig & 22.0\sd{0.7} \\
\quad Independent (48) & 18.7\sd{0.9} & 0.26\sd{0.06} & 20.3\sd{0.5} & 0.45\sd{0.05} & $+$1.55\sd{0.81}\sig & 22.0\sd{0.9} \\
\bottomrule
\end{tabular}
\end{table*}

Without the consistency term the projections at two flow times disagree more. \Cref{M-eq:loss-comm} measures the mismatch between projecting and then advancing along the flow and advancing and then projecting. This mismatch ends training at 0.168 against PHANTESA's 0.107 on amyloid PET, at 0.173 against 0.119 on T2 and at 0.108 against 0.075 on perfusion. The residual form keeps structure, because the projection without it loses SSIM on every task. None of the four ablations of \cref{M-fig:ablations} changes whose subject the output shows. On T2 each of the four outputs correlates best with its own subject's real target in 29 or 30 of the 34 sessions, against 31 for PHANTESA. On amyloid PET it does so in 35 to 37 of the 40, against 36, and on perfusion in 21 to 23 of the 27, against 23. Nor does the segmentation judge read the ablations' T2 scans much differently. Removing the residual form lowers its whole-tumor Dice by 0.03 (95\% interval 0.01 to 0.08), and no other ablation changes a Dice by an amount whose interval excludes zero. On perfusion, removing the residual form, the consistency term or the training range lowers the overlap of the top tenth of rCBV by 0.010 to 0.031, with intervals that exclude zero, and changes neither ratio's correlation by an amount whose interval excludes zero. These four parts therefore shape the output's fidelity, and the population term also sets the output's distance to the target population. The pseudo-target term ties the output to its subject where the class order inverts (\cref{sec:sup-selector}).

\subsection{The reading of each task}
\label[suppsec]{sec:sup-reading}

Each task's reading is chosen on its validation sessions (\cref{M-fig:flowtime}C). On T2, reading once at flow time 0.95 scores 0.41\,dB higher in PSNR than the chain (95\% interval 0.34 to 0.48), in all 34 sessions, so T2 uses the late reading. On perfusion the chain scores 0.30\,dB higher (0.24 to 0.37), in 26 of the 27 sessions. On amyloid PET the two readings tie in PSNR, the late reading scoring 0.14\,dB higher (interval $-0.02$ to 0.34) in 20 of the 40 sessions, whereas the chain has the higher SSIM, by 0.014 in 37 of the 40 sessions, so amyloid PET and perfusion keep the chain. The chain has the higher SSIM in 5 of the 6 comparisons over tasks and constructions, by 0.014 to 0.022, because the mean of eight draws is smoother than one deterministic estimate; the exception is the rank map on T2, where the late reading is higher by 0.008. On T2 the late reading costs tumor-core Dice. On the 20 sessions with a tumor core, the rank map read through the chain scores 0.045 higher there (95\% interval 0.006 to 0.108), whereas the late reading scores 0.029 higher in whole-tumor Dice (0.003 to 0.058; \cref{tab:sup-persubject}). The reading is chosen on PSNR and, where the PSNR ties as on amyloid PET, on SSIM.

\subsection{The chain's draws and the target population}
\label[suppsec]{sec:sup-chain}

\begin{figure*}[t]
  \centering
  \includegraphics[width=\linewidth]{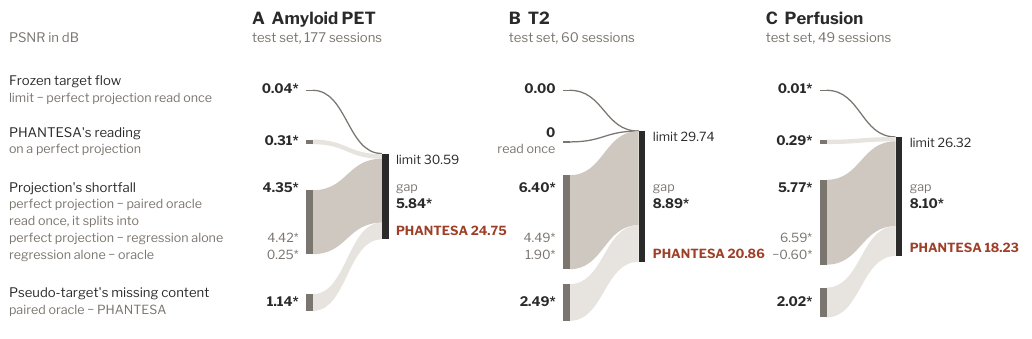}
  \caption{\textbf{The gap from PHANTESA to the autoencoder's limit, test set.} Most of it opens in the projection. Each panel reads from left to right. The four numbers on the left are the parts of the gap in dB, and the label beside each names the two measurements whose difference it is. Their bands merge into the bar on the right, which runs from the autoencoder's limit at its top to PHANTESA's PSNR at its bottom, so its height is the whole gap, drawn at one scale for the three tasks. The two smaller numbers under the projection's shortfall divide it at the regression alone, with both models read once. An asterisk marks a part that the paired test separates from zero (\cref{sec:sup-scoring}).}
  \label{fig:sup-ledger}
\end{figure*}

On the latent, which the population term reads, the chain's draws lie as near the target population as one projection, or nearer. The population needs its own measurement, because re-entry can raise a session's PSNR against its own real target (\cref{M-fig:flowtime}C) and still move the draws away from the target population. \Cref{prop:marginal} covers one projection read through the target flow, whereas the chain re-enters the flow and draws new noise at each re-entry, so the proposition does not cover it. The population is therefore measured on one draw of each session and on the mean of each session's eight draws, which is PHANTESA's prediction. The chain belongs to the reading, which is chosen on the validation sessions, so the population is measured on validation sessions too. The reference is the target population. It is represented by the real target scans of the validation sessions held out from the target subjects, which are 35 OASIS-3 sessions on amyloid PET and 26 UPENN-GBM sessions on perfusion. The draws come from the validation sessions of the source subjects and of the target subjects of the two tasks, four cohorts in all, and on ADNI from the 168 sessions that pass the registration checks. The distance is the population term's unbiased MMD$^2$ on the whole latent as the population term takes it, and its values are given in units of $10^{-3}$ (\cref{tab:sup-marginal}). Even two halves drawn from the same real scans do not lie at distance zero, because each half is small and random. Each half holds 17 or 18 OASIS-3 sessions or 13 UPENN-GBM sessions. The real scans are therefore split into two random halves 200 times, and the 95th percentile of the distance between the halves is the split-half floor. The floor is how large that distance gets by chance alone. Each cohort's measurement draws its own 200 splits, so the two floors of one set of real scans differ by 0.1 unit. Synthetic scans that lie below the floor are no farther from the real scans than one half of the real scans is from the other. The map of \cref{prop:marginal} integrates the target flow from the projected state to its endpoint. One projection read once at flow time 0.95 lies within 0.4 units of it in every cohort, so the one-step estimate stands in for that map.

Compared with one projection, the draws lie nearer to the real scans in all four cohorts, and the 95\% interval of each difference excludes zero. The distance falls at each re-entry. The draws lie below the floor in all four cohorts. One projection lies below it in three and above it on the ADNI sessions, at 30.0 against 26.3. ADNI's own real amyloid PET scans also lie above the floor, at 31.4, and the permutation test separates them from the OASIS-3 scans, so on ADNI the floor and the permutation test reflect the gap between the two cohorts. The pseudo-targets of the same sessions lie far above the floor in three of the four cohorts, at 66 to 255 units against floors of 26.3 to 27.4, so the measure separates a population that is not the target's. On the perfusion sessions of the target subjects they lie below the floor, at 4.4 against 27.4, so there the latent distance cannot tell the pseudo-targets from the real scans. Matching the latent population does not carry the spread of amyloid burden. On the OASIS-3 validation sessions the chain's predicted Centiloid has a standard deviation of 12.0 across sessions and one projection's 12.7, against 44.0 for the real scans, and on the ADNI validation sessions they have 8.8 and 8.4 against 51.9. Because one projection falls short there too, the shortfall belongs to the projection and not to the recursion. The mean of eight draws lies within 1 unit of one draw's distance, so averaging keeps the draws' population.

The table's row for T2 and its last column belong to the ablation of the population term (\cref{sec:sup-ablation}). T2 is read once and has no chain, so T2's row compares that one prediction with the 64 real T2 scans of the target subjects' validation sessions.

\subsection{The projection trained on real targets}
\label[suppsec]{sec:sup-oracle}

\begin{figure}[t]
  \centering
  \includegraphics[width=\linewidth]{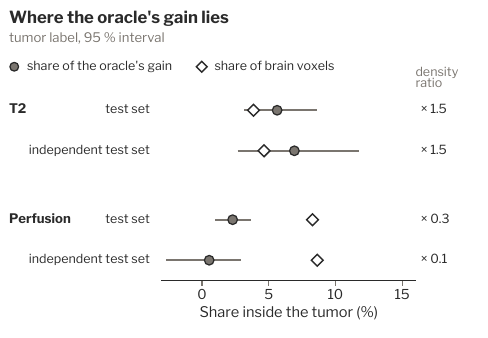}
  \caption{\textbf{Where the oracle's gain over PHANTESA lies.} It lies mostly in normal-appearing tissue. Discs give the tumor's share of the gain, diamonds the tumor's share of brain voxels.}
  \label{fig:sup-lesion}
\end{figure}

More than 70\% of the distance between PHANTESA's PSNR and the autoencoder's limit comes from the projection, and the rest from the pseudo-target and the reading. The limit is the PSNR of a real target scan after the target autoencoder encodes and decodes it. Three references separate the causes. The paired oracle trains the projection as PHANTESA does but replaces each subject's pseudo-target by that subject's own real target scan. The perfect projection involves no training. For each test subject it encodes the real target scan, carries the latent back along the target flow to the flow time of the reading and gives that state to the target flow, so no projection can return a state closer to the real scan. The regression alone trains the projection's network at one flow time with a single term, the relative squared distance of \cref{M-eq:loss-pt} to the state of the subject's real target.

Read as PHANTESA is read, the oracle exceeds PHANTESA by 1.1 to 2.5\,dB on the test set and by 1.2 to 1.6\,dB on the independent test set, so the content that the pseudo-target misses is worth that much (\cref{tab:sup-oracle}). Training the oracle only at flow times $0.6$ to $0.95$ changes T2 by $-0.07$\,dB (95\% interval $-0.23$ to $+0.09$), so the early flow times add nothing to it.

The gap ledger splits each task's distance to the limit into paired differences on the same test sessions, 177 for amyloid PET, 60 for T2 and 49 for perfusion (\cref{fig:sup-ledger}). Its entries come from our own implementation of the readings and the scoring, which reproduces the benchmark's PSNR to the third decimal. The generative models lose almost nothing. From a perfect projection, the frozen target flow read once returns the autoencoder's limit within 0.05\,dB on every task. The projection's shortfall is the perfect projection minus the oracle, both read as PHANTESA is. It is 4.3 to 6.4\,dB, more than 70\% of the distance on every task (74\% on amyloid PET, 72\% on T2, 71\% on perfusion). The two numbers under the projection's band split the same shortfall with both models read once, so on the two tasks that read the chain they do not add up to it. The oracle minus PHANTESA is the pseudo-target's missing content, 1.1 to 2.5\,dB, and the chain costs a perfect projection a further 0.3\,dB on amyloid PET and perfusion.

The shortfall lies in the projection's objective and in its application, which is the flow time at which the projection moves a state between the flows and the way its output is read. The regression alone tests the objective. It trains for 3,000 steps of 8 sessions on the oracle's latents, at flow time 0.95 on T2 and 0.85 on the other tasks, and is read once at that time. A perfect projection exceeds it by 4.5\,dB on T2, 4.4 on amyloid PET and 6.6 on perfusion, so a direct regression onto the real target does not close the shortfall. The oracle's population and consistency terms constrain its prediction beyond the regression onto the real target. Read once, the oracle lies 1.9\,dB below the regression alone on T2 and 0.3\,dB below it on amyloid PET, and 0.6\,dB above it on perfusion (\cref{fig:sup-ledger}). The constraint therefore costs PSNR on T2 and amyloid PET and raises it on perfusion.

The application decides much of the shortfall, because the projection is accurate only late in the flow (\cref{M-fig:flowtime}A). The error of a projected state is its squared distance to the real target state divided by that state's squared norm, and lower is better. A projection that always returns the zero state carries nothing of the target and has an error of exactly 1, so this reference is called predicting zero, and an error near 1 means that the projection has recovered nothing of the target state. On 12 T2 validation sessions the oracle's error is 1.03 at flow time 0.05, 0.98 at 0.2, 0.82 at 0.35 and 0.33 at 0.95, and PHANTESA's is 1.07 at 0.05 and 0.41 at 0.95. Up to flow time 0.2 a projection is therefore no better than predicting zero, even when trained on real targets, because near the Gaussian end the two flows' states do not determine each other. The training times below 0.35 thus teach the projection little, although they fill a third of the range. The share of fine detail in the remaining error grows along the flow. With the decoded error split into its Gaussian blur at 3 voxels and the fine band that the blur removes, the fine band holds 68\% of the two bands' squared error at flow time 0.05 and 80\% at 0.95 for PHANTESA, on the same 12 sessions. \Cref{M-fig:flowtime}B shows one of them, whose shares are 60\% and 79\%.

The reading follows from the same fact. The chain costs a perfect projection 0.3\,dB on amyloid PET and perfusion, yet it gains PHANTESA 0.1 to 0.3\,dB there against one projection read at 0.85. This comparison keeps the flow time of the chain's last projection, whereas \cref{sec:sup-reading} compares the chain with the late reading at 0.95. Re-entry therefore helps only a projection that is still far from the real target state. On T2 the chain costs a perfect projection $0.81$\,dB and the oracle $0.65$\,dB, each more than it costs PHANTESA, so the late reading is the better one on T2 whatever the projection.

The content that the pseudo-target misses lies mostly outside the lesion (\cref{fig:sup-lesion}). The oracle's gain is the squared error it removes relative to PHANTESA. On T2, 5.6\% of this gain (95\% interval 3.2 to 8.6) lies inside the expert tumor label on the test set, and 6.9\% on the independent test set, while the tumor fills about 4\% of the brain. The gain is thus about 1.5 times denser inside the tumor, and about 94\% of it still lies in normal-appearing tissue. On perfusion the tumor fills 8\% of the voxels and holds 2.3\% of the gain (1.0 to 3.6), so what the oracle adds there is perfusion contrast in normal-appearing tissue.

\section{Experiment settings}
\label[suppsec]{sec:sup-settings}

A session is one imaging visit of one subject, and every score compares a synthetic scan with the real scan of the same session. Image fidelity measures how close the voxel values of the two scans are. The analytical verification reads one biomarker from each scan and measures how well the values from the synthetic scans follow those from the real scans. On amyloid PET the biomarker is the amyloid burden in Centiloid, on T2 the segmentation of the whole tumor, and on perfusion the blood volume in the tumor against its mirror region. On amyloid PET and perfusion the synthetic values are correlated with the real ones over the sessions, and on T2 the segmentation obtained with the synthetic scan is scored by its Dice against the expert label.

\subsection{Data and test sets}
\label[suppsec]{sec:sup-data}

\paragraph{Preprocessing.} Every scan is restricted to its brain mask, and its intensities inside the mask are clamped to their 1st and 99th percentiles. It is cropped to the mask's bounding box, resized so that its longest axis spans 128 voxels and padded to a $128^3$ volume. The intensities inside the mask are then standardized to zero mean and unit standard deviation per scan and divided by five, so that a value five standard deviations from the mean reaches $1$ or $-1$, and the result is clipped to $[-1,1]$. Voxels outside the mask are set to $-1$. The perfusion scan covers only a slab of the brain, so on UPENN-GBM the mask is the brain mask restricted to that slab.

\begin{table*}[t]
\centering
\footnotesize
\setlength{\tabcolsep}{6pt}\setlength{\aboverulesep}{0pt}\setlength{\belowrulesep}{0pt}\renewcommand{\arraystretch}{1.18}
\caption{\textbf{Every task trains on two disjoint groups of subjects and tests on sessions that training never sees.} Each cell gives sessions, with the number of subjects in parentheses where it differs; a test set holds one session per subject. On amyloid PET a session whose registration fails is not scored, and on perfusion the biomarker is read only on the sessions whose regions pass their checks (\cref{sec:sup-verification}).}
\label{tab:sup-splits}
\begin{tabular}{l ccc}
\toprule\addlinespace[2pt]
 & Amyloid PET & T2 & Perfusion \\
\midrule[\heavyrulewidth]
\multicolumn{4}{l}{\textbf{Subjects}} \\
\quad Source subjects, source scan type & ADNI, T1 & BraTS, GE sites, T1Gd & UPENN-GBM, one half, T1Gd \\
\quad Target subjects, target scan type & OASIS-3, amyloid PET & BraTS, Siemens sites, T2 & UPENN-GBM, other half, rCBV \\
\multicolumn{4}{l}{\textbf{Training}} \\
\quad Unpaired models, source scans & 635\,(345) & 656\,(376) & 189\,(175) \\
\quad Unpaired models, target scans & 422\,(316) & 587\,(195) & 189\,(175) \\
\quad Paired models, both scans & 635\,(345) & 1,243\,(571) & 378\,(350) \\
\multicolumn{4}{l}{\textbf{Validation sessions}} \\
\quad Held out from the source subjects & 170\,(94) & 34\,(20) & 27\,(25) \\
\quad Used for the design choices & 40\,(22) & 34\,(20) & 27\,(25) \\
\multicolumn{4}{l}{\textbf{Test set, source subjects}} \\
\quad Held out & 185 & 60 & 49 \\
\quad Scored & 177 & 60 & 49 \\
\quad Biomarker read & 177 & 60 & 45 \\
\multicolumn{4}{l}{\textbf{Independent test set, target subjects}} \\
\quad Held out & 79 & 60 & 48 \\
\quad Scored & 76 & 60 & 48 \\
\quad Biomarker read & 76 & 60 & 46 \\
\bottomrule
\end{tabular}
\end{table*}

\paragraph{Training data and test sets.} Each task's subjects form a group of source subjects and a group of target subjects, with no subject in both. Validation and test sessions come from subjects that give no training session, so no model is scored on a subject it trained on (\cref{tab:sup-splits}). The unpaired models train on the source scans of the source subjects and the target scans of the target subjects, so no training session gives them both scans of one subject. The paired models train on the sessions that hold both scans, from the ADNI subjects on amyloid PET and from both groups on T2 and perfusion. A subject may give several training or validation sessions. The validation sessions come from source subjects held out from training. The design choices of \cref{sec:sup-design} are made on the same sessions. The study of the chain's draws reads the 168 of the 170 ADNI validation sessions that pass the registration checks and also the validation sessions held out from the target subjects, 35 sessions of 28 OASIS-3 subjects and 26 sessions of 25 UPENN-GBM subjects (\cref{sec:sup-chain}). The test set holds one session of each source subject held out for testing, and the main paper reports it. The independent test set holds one session of each target subject held out for testing, so it repeats every result on subjects that the test set does not contain. The amyloid PET task is defined across datasets, so its paired models train on the ADNI pairs alone, whereas on T2 and perfusion they train on both groups. No model therefore trains on an MRI of OASIS-3, and there the independent test set measures how each model transfers to a cohort whose MRI it has never seen. Each amyloid PET test set also lies in the training cohort of one kind of model. The paired models train on ADNI's MRI and PET and on no scan of OASIS-3, and PHANTESA's target model trains on the PET of OASIS-3 and on no PET of ADNI. Every amyloid PET scan of both cohorts is a florbetapir scan, and the OASIS-3 scans of other tracers are not used. A failed check removes a session only from the scores it affects. On amyloid PET a failed registration removes it from every score, whereas on perfusion the region checks remove it from the biomarker alone, so the last two rows of each test set in \cref{tab:sup-splits} differ there (\cref{sec:sup-verification}). Every scan comes from a public cohort that releases de-identified data, and this study is a secondary analysis under each provider's terms of use.

\paragraph{The paired models train on more scans.} On T2 and perfusion the paired models train on every training session of both groups, 1,243 and 378 sessions, and use both scans of each, 2,486 and 756 scans. The unpaired models use one scan of each of these sessions. From a source subject's session they take the source scan, and from a target subject's session the target scan, 1,243 and 378 scans in all. On amyloid PET the paired models use the 635 ADNI training sessions with both scans, 1,270 scans, whereas the unpaired models use 635 ADNI MRIs and 422 OASIS-3 PET scans, 1,057 in all. \Cref{tab:sup-splits} counts sessions, so each of its paired sessions holds two scans. Their pairs come from ADNI alone, so that the independent test set from OASIS-3 shows clearly how they generalize to another cohort.

\paragraph{Human-subject data.} ADNI acquired its data with the approval of the institutional review boards of all its participating institutions, and OASIS-3 under procedures approved by the Institutional Review Board of Washington University School of Medicine~\cite{lamontagne2019oasis3}. Both obtained written informed consent from all participants. We applied to both data providers and were granted access for academic use under their data use agreements. UPENN-GBM was collected under a protocol approved by the Institutional Review Board of the University of Pennsylvania Health System, with informed consent from all participants, and is released through The Cancer Imaging Archive under a Creative Commons Attribution license~\cite{bakas2022upenn}. The BraTS 2024 post-treatment glioma data were contributed retrospectively by seven academic medical centers and are released by the challenge's organizers~\cite{deverdier2024brats}. This work analyzes the de-identified data of these four cohorts and acquires no new data from human subjects, so it underwent no separate ethics review.

\subsection{Baselines}
\label[suppsec]{sec:sup-baselines}

Every baseline reads the same preprocessed volumes as PHANTESA and is scored by the same procedure. The paired models train on the paired training sessions of \cref{tab:sup-splits}, and the unpaired models on the same unpaired scans as PHANTESA.

\paragraph{Paired models.} ResViT-2.5D~\cite{dalmaz2022resvit} synthesizes one slice at a time. Its generator joins a residual convolutional encoder and decoder through vision-transformer blocks, and it trains with a voxel-wise L1 term and an adversarial term against a patch discriminator. Pix2Pix-3D~\cite{isola2017image} is the conditional adversarial network of pix2pix with its U-Net generator and patch discriminator made three-dimensional. It trains on whole volumes with the same L1 and adversarial terms. CoCoLIT~\cite{sargood2025cocolit} runs its published code in four stages. It trains an autoencoder for the source scan type and one for the target scan type, then a latent diffusion model of the target latents, and then a ControlNet that conditions that model on the source latent. CoCoLIT is trained on all three tasks.

CoCoLIT's training failed on T2 and on perfusion. On perfusion its two autoencoders output the mean image throughout training. Its MRI autoencoder reconstructs four held-out source scans at 13.6\,dB, against 30\,dB for its MRI autoencoder on the amyloid PET task, so its latents carry no anatomy. On T2 its source autoencoder reconstructs four held-out scans at 28.8\,dB, whereas its target autoencoder reconstructs the real T2 of the same sessions at 11.8\,dB. Its diffusion model is trained on the latents of the target autoencoder, so every later stage builds on an autoencoder that does not reconstruct T2. Accordingly CoCoLIT's output is not a T2. For example, the ventricles in its output are dark although they are bright in a T2 (\cref{fig:sup-qual-t2}). Both runs stand in the benchmark table as failed runs, and their scores there measure no synthesis.

\paragraph{Unpaired models.} CUT-3D~\cite{park2020contrastive} synthesizes whole volumes with an adversarial term and a patch-wise contrastive term. The contrastive term draws the features of each output patch toward those of the source patch at the same place and away from those of other source patches. The published two-dimensional code is made three-dimensional, with its ResNet generator and patch discriminator. Retrieval-RF is a baseline built for this benchmark from a Rectified Flow (RF) and retrieved pseudo-pairs. It builds the pseudo-pairs before training. Each source session is matched to the target session nearest to it in age, sex, intracranial volume and the overlap of the two brain masks, never by diagnosis or tumor label. A conditional Rectified Flow is then trained in the latent space of PHANTESA's autoencoders with the source latent given to its network as a second input beside the state. It generates a target latent from noise, which PHANTESA's target decoder decodes. DDIB~\cite{su2023dual} trains nothing. It encodes the source, integrates PHANTESA's source flow from the source latent back to noise and its target flow from that noise forward to a target latent, and decodes the result, so it shows what the two flows give without a projection.

\paragraph{Two further unpaired models.} The Supplement adds two unpaired models to the three of the main paper. CycleGAN-3D~\cite{zhu2017unpaired} trains a generator from the source scan type to the target scan type and a second generator back, each with an adversarial term against a patch discriminator. Its cycle term requires that a scan carried to the other scan type and back returns to itself, which ties each output to its source. At inference the generator from the source to the target scan type runs once. UNSB-3D~\cite{kim2024unpaired} learns a Schr\"odinger bridge from the source population to the target population. A Schr\"odinger bridge is a stochastic path between two populations, and UNSB-3D follows it in a sequence of 5 steps. One generator serves every step and predicts the target scan from the current state. UNSB-3D trains with an adversarial term, a bridge term that keeps each prediction near its state, and the patch-wise contrastive term of CUT. At inference the generator runs once per step. Both models are made three-dimensional as CUT-3D is, and both train on PHANTESA's unpaired data.

UNSB-3D's training failed on amyloid PET. Its generator collapsed early in training, and its output inside the brain stayed at the background value from then on (\cref{fig:sup-qual-av45}). UNSB-3D's PSNR on the validation sessions fell from 21.9\,dB before the collapse to about 6\,dB after it. Its output gives no defined uptake ratio on any session, because its mean uptake in the whole cerebellum is not positive, so UNSB-3D has no Centiloid and no staging on this task. The run stands in the benchmark table as a failed run.

\subsection{Image fidelity and statistical tests}
\label[suppsec]{sec:sup-scoring}

One procedure scores every model, PHANTESA included, from its synthetic scans of the same sessions, and the scores are averaged over the sessions. PSNR is $10\log_{10}(4/\mathrm{MSE})$ in dB, with the mean squared error taken inside the brain mask of the preprocessed volume and 4 the square of the intensity span of $[-1,1]$. The span is fixed at 2 for every volume rather than taken from the volume's own intensities, because a fixed span exposes an offset. A model may add the same constant to every voxel of its synthetic scan. This offset raises the mean squared error against the real scan, so the model loses PSNR. A peak taken from each volume's own maximum would rise with the offset and partly hide that loss. A higher PSNR means that the synthetic scan lies closer to the real target scan voxel by voxel. A real target scan that the target autoencoder encodes and decodes scores 26.3 to 30.6\,dB on the test set. This PSNR is the autoencoder's limit, and the gap ledger of \cref{sec:sup-oracle} splits PHANTESA's distance from it into parts (\cref{fig:sup-ledger}). SSIM~\cite{wang2004image} compares local windows of the two volumes in brightness, contrast and structure and equals 1 for two identical windows. It is computed in 3D with a cubic window of 7 voxels per side and the same span of 2, and the score is its mean over the voxels inside the brain mask. The benchmark table (\cref{tab:benchmark}) gives each score as its mean with its standard deviation over the sessions, and each correlation with the standard deviation of its value over 10,000 bootstrap resamples of the sessions.

\begin{table*}[t]
\centering
\footnotesize
\setlength{\tabcolsep}{3.4pt}\setlength{\aboverulesep}{0pt}\setlength{\belowrulesep}{0pt}\renewcommand{\arraystretch}{1.18}
\caption{\textbf{PHANTESA leads the compared unpaired models in PSNR on every task and both test sets, but stays below the best paired models.} Rows tinted \colPaired{blue} are the paired models and the \colOurs{terracotta} row is PHANTESA. Each cell is the mean over the sessions with its standard deviation in small type, or a correlation $r$ with its bootstrap standard deviation, and $\Delta$ is the whole-volume score minus the brain-mask score; $\Delta$PSNR is positive for a model that leaves the background empty and negative for one that writes signal there. \sig\,The paired test separates the cell from PHANTESA's (\cref{sec:sup-scoring}). Bold marks the best model in a block's column and underline the second. $^{\ddagger}$A failed run, with its scores reported as they are. CoCoLIT's training failed on T2 and on perfusion and UNSB-3D's on amyloid PET (\cref{sec:sup-baselines}), and on perfusion DDIB, which trains nothing, outputs maps far below the real ones. UNSB-3D's failed run yields no defined uptake ratio, so it has no Centiloid. On perfusion the ratios of CoCoLIT and DDIB are defined on 4 and 14 of the 45 read sessions of the test set and on 3 and 6 of the 46 of the independent test set, below the half that scoring requires. A dash is a cell that is not scored.}
\label{tab:benchmark}
\begin{tabular}{l ccccc ccccc}
\toprule\addlinespace[2pt]
 & \multicolumn{5}{c}{Test set} & \multicolumn{5}{c}{Independent test set} \\
\cmidrule(lr){2-6}\cmidrule(lr){7-11}
Model & SSIM\,$\uparrow$ & PSNR\,$\uparrow$ & Biomarker\,$\uparrow$ & $\Delta$SSIM & $\Delta$PSNR & SSIM\,$\uparrow$ & PSNR\,$\uparrow$ & Biomarker\,$\uparrow$ & $\Delta$SSIM & $\Delta$PSNR \\
\midrule[\heavyrulewidth]
\multicolumn{1}{l}{\textbf{Amyloid PET}} & \multicolumn{5}{c}{\textcolor{markgrey}{ADNI, 177 sessions; Centiloid $r$}} & \multicolumn{5}{c}{\textcolor{markgrey}{OASIS-3, 76 sessions; Centiloid $r$}} \\
\midrule
\rowcolor{pairedrow}ResViT-2.5D & \textbf{0.65}\sd{0.07}\sig & \textbf{26.9}\sd{1.3}\sig & \textbf{0.29}\sd{0.07} & $+$0.25 & $+$5.2 & 0.64\sd{0.08} & \underline{26.9}\sd{1.6}\sig & 0.22\sd{0.09} & $+$0.25 & $+$5.3 \\
\rowcolor{pairedrow}Pix2Pix-3D & \underline{0.64}\sd{0.08}\sig & \underline{26.7}\sd{1.4}\sig & 0.19\sd{0.07} & $+$0.25 & $+$5.1 & \textbf{0.65}\sd{0.09} & \textbf{27.1}\sd{1.7}\sig & 0.18\sd{0.10} & $+$0.24 & $+$5.1 \\
\rowcolor{pairedrow}CoCoLIT & 0.45\sd{0.05}\sig & 24.5\sd{1.0}\sig & $-$0.10\sd{0.08} & $+$0.37 & $+$2.7 & 0.47\sd{0.06}\sig & 24.9\sd{1.1}\sig & $-$0.08\sd{0.10} & $+$0.36 & $+$2.6 \\
CUT-3D & 0.55\sd{0.09}\sig & 21.5\sd{0.7}\sig & 0.14\sd{0.07} & $+$0.27 & $+$4.0 & 0.59\sd{0.10}\sig & 21.9\sd{0.9}\sig & 0.21\sd{0.16} & $+$0.23 & $+$4.0 \\
Retrieval-RF & 0.35\sd{0.06}\sig & 18.8\sd{1.2}\sig & $-$0.04\sd{0.06} & $+$0.41 & $+$4.1 & 0.41\sd{0.06}\sig & 19.5\sd{1.3}\sig & \textbf{0.26}\sd{0.12} & $+$0.37 & $+$3.8 \\
DDIB & 0.27\sd{0.06}\sig & 18.2\sd{1.7}\sig & 0.01\sd{0.08} & $+$0.45 & $+$3.6 & 0.30\sd{0.06}\sig & 18.9\sd{1.7}\sig & \underline{0.26}\sd{0.13} & $+$0.44 & $+$3.3 \\
CycleGAN-3D & 0.53\sd{0.11}\sig & 21.0\sd{1.0}\sig & \underline{0.24}\sd{0.07} & $+$0.30 & $+$4.9 & 0.60\sd{0.10}\sig & 22.2\sd{1.2}\sig & 0.13\sd{0.15} & $+$0.26 & $+$4.8 \\
UNSB-3D$^{\ddagger}$ & $-$0.02\sd{0.01}\sig & 6.1\sd{0.1}\sig & -- & $+$0.64 & $+$5.2 & $-$0.02\sd{0.02}\sig & 6.1\sd{0.1}\sig & -- & $+$0.64 & $+$5.3 \\
\rowcolor{oursrow}\textbf{PHANTESA} & 0.59\sd{0.09} & 24.7\sd{1.2} & 0.10\sd{0.07} & $+$0.28 & $+$4.4 & \underline{0.65}\sd{0.10} & 25.8\sd{1.5} & 0.08\sd{0.11} & $+$0.24 & $+$4.1 \\
\midrule
\multicolumn{1}{l}{\textbf{T2}} & \multicolumn{5}{c}{\textcolor{markgrey}{BraTS, 60 sessions; whole-tumor Dice}} & \multicolumn{5}{c}{\textcolor{markgrey}{BraTS, 60 sessions; whole-tumor Dice}} \\
\midrule
\rowcolor{pairedrow}ResViT-2.5D & \textbf{0.76}\sd{0.07}\sig & \textbf{27.6}\sd{1.6}\sig & \textbf{0.66}\sd{0.24}\sig & $+$0.17 & $+$5.4 & \textbf{0.71}\sd{0.08}\sig & \textbf{26.3}\sd{1.5}\sig & \textbf{0.71}\sd{0.24} & $+$0.20 & $+$5.4 \\
\rowcolor{pairedrow}Pix2Pix-3D & \underline{0.72}\sd{0.08}\sig & \underline{26.0}\sd{1.3}\sig & \underline{0.64}\sd{0.25} & $+$0.20 & $+$5.2 & \underline{0.68}\sd{0.09}\sig & \underline{25.3}\sd{1.4}\sig & \underline{0.70}\sd{0.24} & $+$0.22 & $+$5.3 \\
\rowcolor{pairedrow}CoCoLIT$^{\ddagger}$ & 0.08\sd{0.02}\sig & 12.3\sd{0.9}\sig & 0.55\sd{0.26}\sig & $+$0.32 & $+$2.5 & 0.06\sd{0.02}\sig & 12.1\sd{0.6}\sig & 0.55\sd{0.30}\sig & $+$0.33 & $+$2.5 \\
CUT-3D & 0.15\sd{0.02}\sig & 15.5\sd{0.6}\sig & 0.30\sd{0.25}\sig & $+$0.55 & $+$4.4 & 0.17\sd{0.03}\sig & 15.4\sd{0.7}\sig & 0.44\sd{0.31}\sig & $+$0.55 & $+$4.8 \\
Retrieval-RF & 0.16\sd{0.03}\sig & 15.6\sd{1.3}\sig & 0.48\sd{0.28}\sig & $+$0.52 & $+$3.3 & 0.18\sd{0.04}\sig & 15.7\sd{1.1}\sig & 0.57\sd{0.30}\sig & $+$0.51 & $+$3.4 \\
DDIB & 0.09\sd{0.03}\sig & 12.9\sd{1.9}\sig & 0.49\sd{0.28}\sig & $+$0.58 & $+$4.4 & 0.10\sd{0.03}\sig & 12.8\sd{1.0}\sig & 0.51\sd{0.31}\sig & $+$0.58 & $+$4.9 \\
CycleGAN-3D & 0.15\sd{0.03}\sig & 13.8\sd{1.2}\sig & 0.47\sd{0.26}\sig & $+$0.55 & $+$5.2 & 0.16\sd{0.03}\sig & 13.6\sd{1.3}\sig & 0.58\sd{0.27}\sig & $+$0.55 & $+$5.3 \\
UNSB-3D & 0.16\sd{0.02}\sig & 15.4\sd{0.7}\sig & 0.50\sd{0.27}\sig & $+$0.55 & $+$4.5 & 0.17\sd{0.02}\sig & 15.3\sd{0.8}\sig & 0.50\sd{0.32}\sig & $+$0.54 & $+$4.6 \\
\rowcolor{oursrow}\textbf{PHANTESA} & 0.45\sd{0.07} & 20.9\sd{0.8} & 0.63\sd{0.23} & $+$0.37 & $+$4.3 & 0.40\sd{0.07} & 21.0\sd{0.8} & 0.69\sd{0.24} & $+$0.41 & $+$4.2 \\
\textit{Real scan} & -- & -- & 0.67\sd{0.24}\sig &  &  & -- & -- & 0.73\sd{0.24}\sig &  &  \\
\midrule
\multicolumn{1}{l}{\textbf{Perfusion}} & \multicolumn{5}{c}{\textcolor{markgrey}{UPENN-GBM, 49 sessions; ratio $r$ on 45}} & \multicolumn{5}{c}{\textcolor{markgrey}{UPENN-GBM, 48 sessions; ratio $r$ on 46}} \\
\midrule
\rowcolor{pairedrow}ResViT-2.5D & \textbf{0.55}\sd{0.06}\sig & \textbf{22.0}\sd{0.7}\sig & 0.79\sd{0.07} & $+$0.37 & $+$7.7 & \textbf{0.54}\sd{0.07}\sig & \textbf{22.0}\sd{0.9}\sig & \underline{0.82}\sd{0.05} & $+$0.38 & $+$7.8 \\
\rowcolor{pairedrow}Pix2Pix-3D & \underline{0.47}\sd{0.05}\sig & \underline{20.4}\sd{0.5}\sig & 0.76\sd{0.08} & $+$0.44 & $+$7.6 & \underline{0.46}\sd{0.05}\sig & \underline{20.4}\sd{0.6}\sig & 0.81\sd{0.06} & $+$0.44 & $+$7.6 \\
\rowcolor{pairedrow}CoCoLIT$^{\ddagger}$ & 0.12\sd{0.04}\sig & 8.9\sd{2.1}\sig & -- & $+$0.30 & $+$2.5 & 0.12\sd{0.04}\sig & 8.8\sd{2.1}\sig & -- & $+$0.31 & $+$2.6 \\
CUT-3D & 0.28\sd{0.04}\sig & 17.1\sd{0.7}\sig & \textbf{0.85}\sd{0.07} & $+$0.58 & $+$7.6 & 0.29\sd{0.04}\sig & 17.3\sd{0.7}\sig & \textbf{0.82}\sd{0.07} & $+$0.57 & $+$7.6 \\
Retrieval-RF & 0.13\sd{0.03}\sig & 13.7\sd{1.2}\sig & $-$0.01\sd{0.15}\sig & $+$0.67 & $+$6.4 & 0.15\sd{0.03}\sig & 14.1\sd{1.2}\sig & 0.02\sd{0.12}\sig & $+$0.66 & $+$6.5 \\
DDIB$^{\ddagger}$ & 0.05\sd{0.05}\sig & 8.7\sd{3.7}\sig & -- & $+$0.54 & $+$4.2 & 0.03\sd{0.04}\sig & 7.3\sd{2.6}\sig & -- & $+$0.52 & $+$4.3 \\
CycleGAN-3D & 0.25\sd{0.03}\sig & 17.2\sd{0.6}\sig & \underline{0.81}\sd{0.08} & $+$0.61 & $+$7.5 & 0.26\sd{0.04} & 17.5\sd{0.6}\sig & 0.81\sd{0.05} & $+$0.61 & $+$7.6 \\
UNSB-3D & 0.25\sd{0.03}\sig & 17.4\sd{0.7}\sig & 0.77\sd{0.08} & $+$0.60 & $+$7.4 & 0.25\sd{0.03} & 17.6\sd{0.7}\sig & 0.63\sd{0.08} & $+$0.61 & $+$7.4 \\
\rowcolor{oursrow}\textbf{PHANTESA} & 0.23\sd{0.06} & 18.2\sd{1.0} & 0.75\sd{0.06} & $+$0.63 & $+$7.6 & 0.26\sd{0.06} & 18.7\sd{0.9} & 0.75\sd{0.07} & $+$0.61 & $+$7.7 \\
\bottomrule
\end{tabular}
\end{table*}

\begin{table*}[t]
\centering
\footnotesize
\setlength{\tabcolsep}{2.6pt}\setlength{\aboverulesep}{0pt}\setlength{\belowrulesep}{0pt}\renewcommand{\arraystretch}{1.18}
\caption{\textbf{Further readings of each biomarker on both test sets.} The staging stays near its chance level of 0.25 except for DDIB on OASIS-3, and on perfusion ResViT-2.5D and Pix2Pix-3D have the highest values of both further readings. Rows tinted \colPaired{blue} are the paired models and the \colOurs{terracotta} row is PHANTESA. Each reading has its test set on the left and its independent test set on the right. Centiloid and the Dice are the mean over the sessions with the standard deviation in small type. The staging's score is its balanced accuracy (\cref{sec:sup-verification}). The judge's Dice on the tumor core and on the enhancing tumor is taken on the 36 and 49 sessions whose expert label holds the region, and the ratio's $r$ inside the enhancing tumor on at most 41 and 45 sessions. The top tenth is the Dice between the tenth of the slab's voxels with the highest blood volume in the synthetic and in the real map. $^{\ddagger}$A failed run, which is CoCoLIT on T2 and on perfusion (\cref{sec:sup-baselines}) or DDIB on perfusion (\cref{sec:sup-verification}). UNSB-3D's failed run on amyloid PET yields no defined uptake ratio and is not scored (\cref{sec:sup-baselines}). A dash is a cell that is not scored.}
\label{tab:sup-secondary}
\begin{tabular}{l cc cc cc cc cc cc}
\toprule\addlinespace[2pt]
 & \multicolumn{4}{c}{Amyloid PET} & \multicolumn{4}{c}{T2, the judge's Dice} & \multicolumn{4}{c}{Perfusion} \\
\cmidrule(lr){2-5}\cmidrule(lr){6-9}\cmidrule(lr){10-13}
 & \multicolumn{2}{c}{Centiloid} & \multicolumn{2}{c}{Staging accuracy\,$\uparrow$} & \multicolumn{2}{c}{Tumor core\,$\uparrow$} & \multicolumn{2}{c}{Enhancing tumor\,$\uparrow$} & \multicolumn{2}{c}{Ratio $r$, enh.\,$\uparrow$} & \multicolumn{2}{c}{Top tenth\,$\uparrow$} \\
\cmidrule(lr){2-3}\cmidrule(lr){4-5}\cmidrule(lr){6-7}\cmidrule(lr){8-9}\cmidrule(lr){10-11}\cmidrule(lr){12-13}
Model & Test & Indep. & Test & Indep. & Test & Indep. & Test & Indep. & Test & Indep. & Test & Indep. \\
\midrule[\heavyrulewidth]
\rowcolor{pairedrow}ResViT-2.5D & 40.3\sd{18.6} & 35.1\sd{17.7} & 0.33 & 0.24 & 0.51\sd{0.39} & 0.47\sd{0.38} & 0.62\sd{0.31} & 0.61\sd{0.31} & 0.63 & 0.58 & 0.49\sd{0.06} & 0.49\sd{0.08} \\
\rowcolor{pairedrow}Pix2Pix-3D & 35.7\sd{26.6} & 27.5\sd{22.4} & 0.35 & 0.27 & 0.51\sd{0.39} & 0.48\sd{0.38} & 0.61\sd{0.31} & 0.61\sd{0.31} & 0.56 & 0.57 & 0.43\sd{0.05} & 0.42\sd{0.06} \\
\rowcolor{pairedrow}CoCoLIT & 7.3\sd{15.7} & 0.7\sd{14.0} & 0.22 & 0.19 & 0.50\sd{0.37}\rlap{$^{\ddagger}$} & 0.44\sd{0.38}\rlap{$^{\ddagger}$} & 0.61\sd{0.30}\rlap{$^{\ddagger}$} & 0.55\sd{0.35}\rlap{$^{\ddagger}$} & -- & -- & 0.14\sd{0.04}\rlap{$^{\ddagger}$} & 0.15\sd{0.04}\rlap{$^{\ddagger}$} \\
CUT-3D & 13.1\sd{13.0} & $-$3.0\sd{13.9} & 0.27 & 0.31 & 0.40\sd{0.39} & 0.40\sd{0.40} & 0.52\sd{0.38} & 0.49\sd{0.39} & 0.37 & 0.53 & 0.31\sd{0.06} & 0.33\sd{0.06} \\
Retrieval-RF & 25.5\sd{40.5} & 29.4\sd{34.2} & 0.22 & 0.35 & 0.46\sd{0.38} & 0.44\sd{0.39} & 0.57\sd{0.34} & 0.53\sd{0.37} & $-$0.17 & 0.04 & 0.17\sd{0.03} & 0.17\sd{0.03} \\
DDIB & 69.2\sd{52.0} & 49.4\sd{52.1} & 0.19 & 0.56 & 0.51\sd{0.38} & 0.42\sd{0.39} & 0.58\sd{0.33} & 0.49\sd{0.39} & -- & -- & 0.13\sd{0.05}\rlap{$^{\ddagger}$} & 0.14\sd{0.06}\rlap{$^{\ddagger}$} \\
CycleGAN-3D & 5.5\sd{11.0} & $-$8.4\sd{12.0} & 0.30 & 0.28 & 0.46\sd{0.39} & 0.42\sd{0.38} & 0.53\sd{0.36} & 0.52\sd{0.35} & 0.24 & 0.47 & 0.26\sd{0.06} & 0.28\sd{0.06} \\
UNSB-3D & -- & -- & -- & -- & 0.51\sd{0.38} & 0.42\sd{0.40} & 0.60\sd{0.33} & 0.49\sd{0.40} & 0.21 & 0.27 & 0.30\sd{0.06} & 0.31\sd{0.06} \\
\rowcolor{oursrow}\textbf{PHANTESA} & 4.4\sd{7.9} & $-$2.9\sd{9.8} & 0.20 & 0.25 & 0.46\sd{0.38} & 0.45\sd{0.37} & 0.56\sd{0.31} & 0.59\sd{0.32} & 0.36 & 0.47 & 0.25\sd{0.09} & 0.29\sd{0.09} \\
\textit{Real scan} & 44.4\sd{53.8} & 27.4\sd{47.3} & -- & -- & 0.52\sd{0.39} & 0.48\sd{0.37} & 0.63\sd{0.30} & 0.61\sd{0.32} & -- & -- & -- & -- \\
\bottomrule
\end{tabular}
\end{table*}

Every score is taken inside the brain mask because the rest of the volume is empty by construction. Preprocessing sets every voxel outside the mask to $-1$, and the mask covers about 30\% of the $128^3$ volume on amyloid PET and T2 and 17\% on perfusion, whose slab is thin. A model that leaves this background empty matches the real scan on most of the volume without reproducing anything of the brain, so a score over the whole volume rises with the background's share and not with the synthetic brain. The benchmark table also gives, for SSIM and PSNR, the mean change from the score inside the brain mask to the score on the whole volume. A model that leaves the background empty gains from it, so its change is positive, whereas a model that writes signal where the real scan is empty loses, so its change is negative.

Because every model is scored on the same sessions, each comparison is paired. Within one task and one test set, two models are compared on an image metric by a Wilcoxon signed-rank test~\cite{wilcoxon1945individual} on their per-session differences. They are compared on a biomarker value by a paired bootstrap test, which resamples the subjects 10,000 times and recomputes the difference between the two models' values each time. Its $p$-value is twice the smaller share of resampled differences on one side of zero. Many pairs are tested at once, so the $p$-values of one metric are corrected together by Holm's procedure~\cite{holm1979simple}. The tests separate a pair when its corrected $p$-value is below 0.05. The benchmark table compares every model with PHANTESA and corrects the $p$-values of one column together. An asterisk marks each cell that the test separates from PHANTESA's. A comparison of two other models in the text is corrected together with all pairs of its metric. A difference that a study reports against zero, such as an ablation minus PHANTESA or a part of the gap ledger, is tested by the same Wilcoxon test on its per-session values and corrected within its table's column or its task. A pair that the tests do not separate is called level. The 95\% intervals that the text gives with differences are bootstrap intervals over the sessions. The interval of a difference between two distances to the target population comes from 1,000 resamples of the sessions. Each resample takes the same sessions for both sets of predictions and keeps the real scans fixed. Within a set, the pairs of a resampled session with its own copy are left out, since a session drawn twice would otherwise count as two distinct sessions. The interval runs from the 2.5th to the 97.5th percentile of the 1,000 differences. The distance of a set of predictions to the real target scans is tested by a permutation test. It pools the two sets and splits the pool at random 1,000 times into sets of the original sizes. Its $p$-value is the share of splits whose distance is at least the observed one. The $p$-values of the sets compared with one set of real scans are corrected together by Holm's procedure.

\paragraph{The session shown per task.} For each task the image grid of the main paper shows one test session, chosen to be typical of every model at once (\cref{M-fig:qualitative}). The test sessions are ranked by PSNR separately for each model. For each session, the distance of its rank from the median rank is taken for every model, and the largest of these distances is kept. The shown session is the one whose largest distance is smallest. On T2 only sessions whose real scan the judge segments with a whole-tumor Dice of at least 0.5 are eligible, so that the shown session has a tumor for the judge to segment. On perfusion only sessions whose ratio is read and whose tumor lies on one side of the mid-sagittal plane are eligible, so that the tumor and its mirror region do not overlap in the image. A tumor lies on one side when at most 5\% of its voxels have a mirror voxel inside the tumor, which holds for 33 of the 45 sessions whose ratio is read. PHANTESA stands beside the unpaired baseline with the highest PSNR.

\subsection{Analytical verification}
\label[suppsec]{sec:sup-verification}

\paragraph{Amyloid PET.} The biomarker is the amyloid burden in Centiloid~\cite{klunk2015centiloid}. On this scale 0 is the level of young controls and 100 that of typical Alzheimer's disease patients. It is read in the same four steps from the preprocessed volume of a PET scan, real or synthetic. First, the scan is returned from the preprocessed volume to the session's own grid by inverting the crop and the resizing. Second, it is carried to a common template space by the registration of the session's MRI to the template, rigid first and then affine. Third, the uptake ratio is the mean uptake in the standard Centiloid cortical composite divided by the mean uptake in the whole cerebellum, each region restricted to the subject's brain mask. A ratio of uptakes needs the zero of the uptake scale, yet standardization subtracted each scan's mean and so removed that zero. One constant is therefore added to every voxel of every scan, real or synthetic, before the ratio is taken. It is 0.446 in the units of the preprocessed volume, the mean over the real amyloid PET scans of the amount that standardization subtracted, so no model receives a correction of its own. Fourth, Centiloid is $175.43$ times the ratio minus $182.26$. An earlier study applies this formula to the tracer florbetapir as the Centiloid project defines it~\cite{risacher2017olfactory}. The published calibration for that tracer prints the coefficients as 175 and 182~\cite{navitsky2018standardization}, and the two forms differ by 0.4 Centiloid at a ratio of 1.5. Each scan is read from its own voxels and no synthetic scan is rescaled to the real one, so a model's intensity bias stays in its Centiloid values, and their mean and spread can be set beside those of the real scans.

A session is scored only when its registration passes four checks that guard against a failed alignment. The brain mask overlaps the template's with a Dice of at least 0.90. The mask holds at least 0.85 of the template mask's volume, its center lies within 6\,mm of the template's, and at least 90\% of each region lies inside the mask. The 177 and 76 sessions of the two test sets are those that pass. The analytical verification is the Pearson correlation $r$ between the synthetic and the real Centiloid over the sessions. It is 1 when a model orders and spaces the subjects as the real scans do and 0 when its values are unrelated to theirs. The real scans spread widely, with a standard deviation of about 50 Centiloid in both cohorts. This spread comes from the amyloid burden, which the MRI does not show, so a synthetic scan that carries only the anatomy cannot order the subjects by it, and its $r$ stays low.

Staging converts each Centiloid value into one of four stages, so that a model is judged on how it orders the subjects and not on the level of its values, and \cref{sec:sup-benchmark} reports it. The real scans are staged by the published cut points at 10, 30 and 60 Centiloid~\cite{collij2024centiloid}, which separate no amyloid pathology, an intermediate range, established pathology and high burden. A model's values may be shifted or compressed, so each model, PHANTESA included, is staged by cut points of its own. They are placed on the model's predictions for the ADNI validation sessions so that each stage receives the same share of sessions as it has among the real scans of those sessions, and the same cut points stage the model on both test sets. For example, if 40\% of the real scans are in the lowest stage, the model's lowest 40\% of values are assigned to it. The agreement with the real stages then measures only whether the model orders the subjects. On OASIS-3 the stages hold other shares of the sessions than on ADNI, so there the agreement also falls with that difference. It is reported as the balanced accuracy, which is the mean over the four stages of the share of a stage's sessions that the model assigns to that stage, and its chance level is 0.25.

\paragraph{T2.} The biomarker is the segmentation of the whole tumor. The segmentation judge is a fixed tumor segmentation network, the brain-tumor SegResNet~\cite{myronenko2019segresnet} that MONAI distributes pretrained on the pre-operative scans and expert labels of BraTS 2018. It takes the session's four sequences, three of them real, and outputs three nested regions, the enhancing tumor, the tumor core and the whole tumor. For each model the judge receives the model's synthetic T2 in place of the real T2 and the three other sequences real. The judge was trained on scans in their own intensity scale, so the synthetic T2 is first returned to the real scan's grid and intensity scale by inverting the preprocessing with the real T2's mean and standard deviation, and every model is restored in this same way. The rescaling is an affine map of each scan's intensities, so it keeps the synthetic scan's own pattern of contrast, and the judge standardizes each sequence again before it segments. Each region is scored by its Dice against the expert label on the sessions whose label contains that region. The expert label's whole tumor is the union of the enhancing tumor, the non-enhancing tumor core and the surrounding non-enhancing hyperintensity, and the resection cavity left by surgery is not part of it. The edema is not scored apart from the whole tumor. The judge never sees a synthetic scan in training, so a change in its Dice reflects what the synthetic T2 shows or omits. The upper reference is the Dice the judge reaches with all four sequences real, 0.67 for the whole tumor on the test set and 0.73 on the independent test set, and a model whose Dice equals it loses nothing by replacing the real T2 for this purpose. A reference without subject information is the projection trained without a per-subject term. Its output is not tied to its subject, yet on the 34 validation sessions of \cref{tab:sup-persubject} it reaches 0.44, against 0.65 with the real T2, because the judge also reads the three real sequences. The judge still responds to the T2 it is given. With CUT-3D's T2 its whole-tumor Dice is 0.30 on the test set, against 0.67 with the real T2 (\cref{tab:benchmark}). The Dice therefore measures whether a synthetic T2 supports or misleads the segmentation beside three real sequences.

\paragraph{Perfusion.} The biomarker compares the blood volume inside the tumor with the blood volume in the same place on the opposite side of the brain. The tumor region is the automated tumor segmentation that UPENN-GBM provides for each session. It holds the enhancing tumor, its necrotic core and the surrounding edema, which together make the whole tumor. Edema holds less blood volume than healthy tissue, so where the edema dominates, the tumor region holds less blood volume than its mirror region, and the ratio defined below lies under 1. The mirror region is the tumor region reflected across the mid-sagittal plane, which divides the brain into its left and right halves, restricted to the brain and the perfusion slab and with any tumor voxels removed. The scans of UPENN-GBM lie on one atlas grid, so the mirror image is the left-right flip of that grid. The biomarker value is the mean rCBV in the tumor divided by the mean rCBV in the mirror region, so a value above 1 means more blood volume in the tumor than in the opposite tissue. Every session of both test sets is a pre-operative baseline scan. On the real maps of the test set this ratio runs from 0.5 to 2.8. Standardization subtracts each map's mean and so moves the zero of the rCBV scale, which a ratio of means needs. One constant is therefore added back to every volume, real or synthetic. It is the median, over the cohort's real maps, of the shift that standardization applies, so no model receives a correction of its own. A session's ratio is read only when two checks pass. Its brain mask overlaps its own mirror image with a Dice of at least 0.90, which guards against a head that lies off the mid-sagittal plane, and at least 80\% of the tumor region and of the mirror region lie inside the perfusion slab. The checks keep 45 of the 49 sessions of the test set and 46 of the 48 of the independent test set. The analytical verification is the Pearson correlation $r$ between the synthetic and the real ratio over these sessions.

The same ratio is also read on the enhancing tumor alone, against the mirror region of the enhancing tumor. A third reading is local. It takes the tenth of the slab's voxels with the highest rCBV in the synthetic map and in the real map, and scores their overlap by the Dice on the read sessions. A ratio is undefined where the mean of a model's mirror region is not positive. A model is scored on the ratio only when it is defined on at least half of the read sessions. DDIB's predictions lie so far below the real maps that its ratio is defined on 14 of the 45 read sessions of the test set and 6 of the 46 of the independent test set, so it has no scored ratio. CoCoLIT, whose training failed on perfusion (\cref{sec:sup-baselines}), has its ratio defined on 4 of 45 and 3 of 46 sessions and has no scored ratio either.

\begin{figure*}[t]
  \centering
  \includegraphics[width=\linewidth]{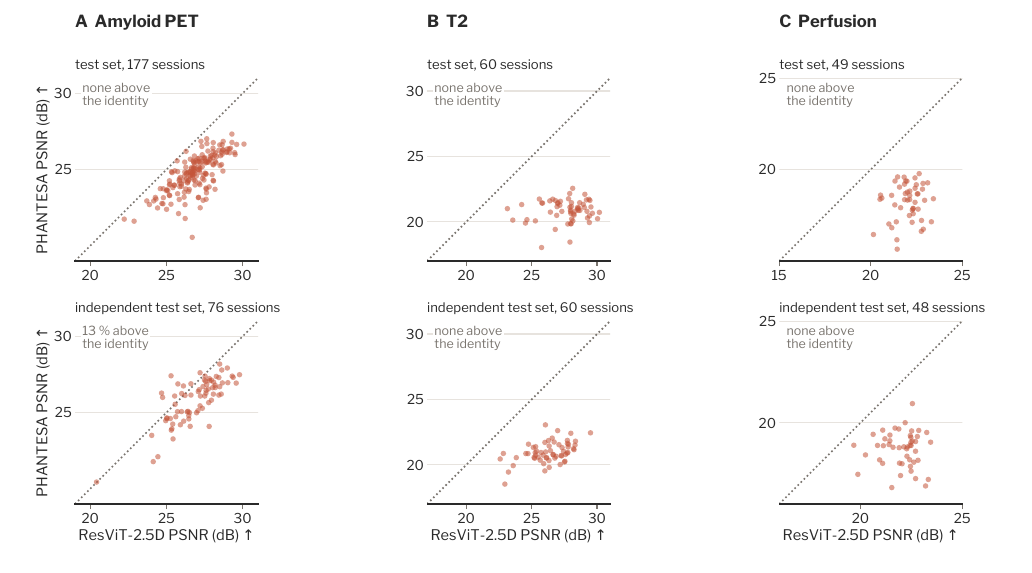}
  \caption{\textbf{PHANTESA against ResViT-2.5D per session.} The paired model leads on every session except 10 of the 76 of OASIS-3. The dotted line is the identity.}
  \label{fig:sup-persession}
\end{figure*}

\begin{figure*}[t]
  \centering
  \includegraphics[width=\linewidth]{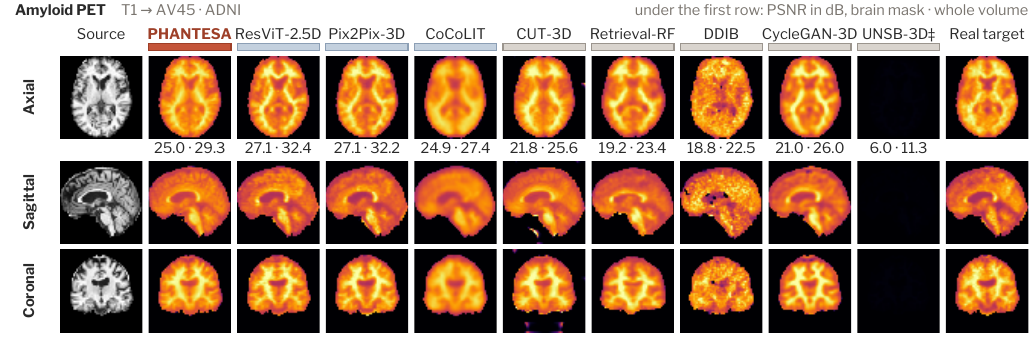}
  \caption{\textbf{One amyloid PET test session for every model, on the center slice of each view.} CoCoLIT blurs the anatomy, DDIB turns it to noise and UNSB-3D's failed run ($^{\ddagger}$) outputs the background value, whereas the other models keep the white-matter pattern of the real scan. The session is the one nearest the median PSNR of every model at once, as in \cref{M-fig:qualitative}; under the first row, each column's PSNR inside the brain mask and on the whole volume.}
  \label{fig:sup-qual-av45}
\end{figure*}

\begin{figure*}[t]
  \centering
  \includegraphics[width=\linewidth]{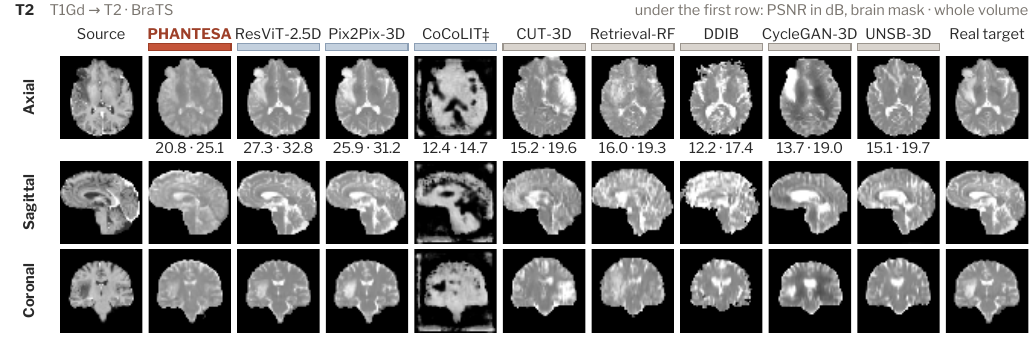}
  \caption{\textbf{One T2 test session for every model, on the center slice of each view.} ResViT-2.5D and Pix2Pix-3D show the tumor's bright edema as the real scan does and PHANTESA a fainter one, whereas Retrieval-RF blurs the anatomy and DDIB breaks it up. CoCoLIT's failed run ($^{\ddagger}$) shows the ventricles dark although they are bright in the real scan. The session is the one nearest the median PSNR of every model at once, as in \cref{M-fig:qualitative}; under the first row, each column's PSNR inside the brain mask and on the whole volume.}
  \label{fig:sup-qual-t2}
\end{figure*}

\begin{figure*}[t]
  \centering
  \includegraphics[width=\linewidth]{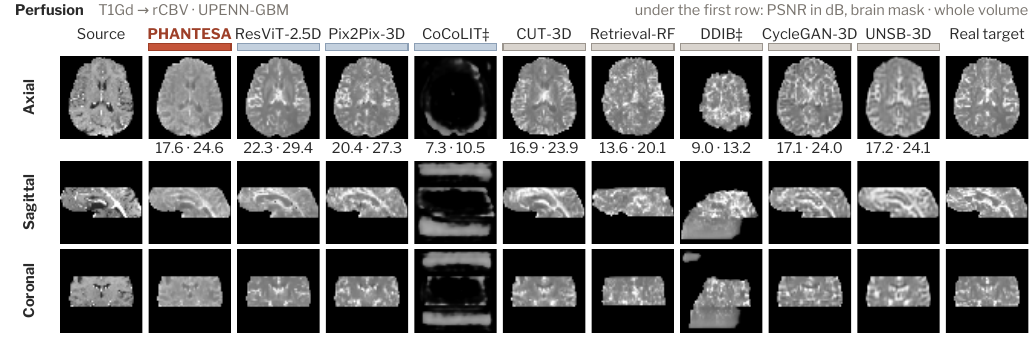}
  \caption{\textbf{One perfusion test session for every model, on the center slice of each view.} Every model keeps the anatomy of the scanned slab except the two that fail on perfusion ($^{\ddagger}$). CoCoLIT's failed run outputs none, and DDIB writes signal outside the slab. The session is the one nearest the median PSNR of every model at once, as in \cref{M-fig:qualitative}; under the first row, each column's PSNR inside the brain mask and on the whole volume.}
  \label{fig:sup-qual-rcbv}
\end{figure*}

\begin{figure*}[t]
  \centering
  \includegraphics[width=\linewidth]{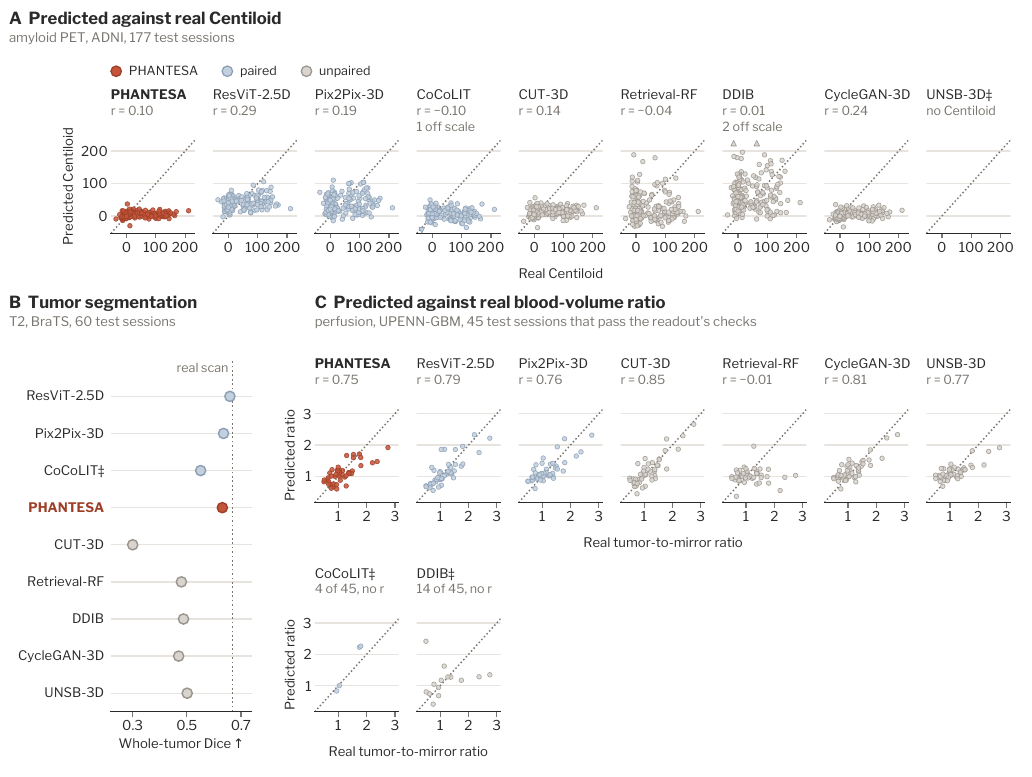}
  \caption{\textbf{The analytical verification of every model, test set.} No model reads amyloid, whereas on T2 PHANTESA, ResViT-2.5D and Pix2Pix-3D approach the real scan and on perfusion most models follow the whole tumor's blood volume. Dotted lines mark the identity or the real scan; values beyond an axis sit at its edge and are counted above the panel. The second row of panel C shows the two models that fail on perfusion ($^{\ddagger}$) on the sessions where their ratio is defined, too few for a correlation (\cref{sec:sup-verification}). In panel B the same mark is on CoCoLIT's failed T2 run, and in panel A on UNSB-3D's failed run, which has no Centiloid.}
  \label{fig:sup-clinical}
\end{figure*}

\subsection{The scaling experiment}
\label[suppsec]{sec:sup-scaling-settings}

\begin{table}[t]
\centering
\footnotesize
\setlength{\tabcolsep}{5pt}\setlength{\aboverulesep}{0pt}\setlength{\belowrulesep}{0pt}\renewcommand{\arraystretch}{1.18}
\caption{\textbf{The scaling experiment changes only the unpaired training scans.} Unpaired training sessions of each size, with the number of subjects in parentheses. The middle size is the benchmark's unpaired training data (\cref{tab:sup-splits}); the smallest keeps a third of its subjects, and the largest adds the additional subjects, subjects of the same datasets in neither group, in two halves. Perfusion has none. The validation sessions and both test sets are those of \cref{tab:sup-splits} at every size.}
\label{tab:sup-scaling-data}
\begin{tabular}{l ccc}
\toprule\addlinespace[2pt]
 & \shortstack{Smallest,\\a third} & \shortstack{Middle,\\all} & \shortstack{Largest, with the\\additional subjects} \\
\midrule[\heavyrulewidth]
\multicolumn{4}{l}{\textbf{Amyloid PET}} \\
\quad Source scans & 202\,(114) & 635\,(345) & 953\,(517) \\
\quad Target scans & 141\,(104) & 422\,(316) & 839\,(489) \\
\multicolumn{4}{l}{\textbf{T2}} \\
\quad Source scans & 218\,(124) & 656\,(376) & 757\,(420) \\
\quad Target scans & 211\,(64) & 587\,(195) & 674\,(238) \\
\multicolumn{4}{l}{\textbf{Perfusion}} \\
\quad Source scans & 62\,(58) & 189\,(175) & -- \\
\quad Target scans & 61\,(58) & 189\,(175) & -- \\
\bottomrule
\end{tabular}
\end{table}

The scaling experiment tests how PHANTESA and the five other unpaired models change when they train on more unpaired scans. It compares a smallest, a middle and a largest size of the unpaired training data (\cref{tab:sup-scaling-data}). The middle size holds all subjects of both groups, so it is exactly the benchmark's unpaired training data. The smallest holds a third of the subjects of each group, drawn once at random by subject with a fixed seed, and every model trains on the same draw. The largest adds the additional subjects, who belong to the same datasets and to neither group. They are further ADNI subjects on amyloid PET and the BraTS subjects of neither group on T2, so on amyloid PET the largest size's target scans come from both OASIS-3 and ADNI. The additional subjects are split at random by subject into two halves, so that one half adds source scans, the other adds target scans and no subject gives both. UPENN-GBM has no additional subjects, so perfusion has two sizes. Only these unpaired models are trained again at each size, and the paired models keep their benchmark training data. At the smallest size the paired models therefore train on about six times as many scans as the unpaired models on T2 and perfusion, and on almost four times as many on amyloid PET.

At each size every model is trained again from random weights on that size's source and target scans alone. PHANTESA trains both autoencoders, both flows and the projection, and the scale of its latents and the class levels of its pseudo-targets are computed again from that size's target scans. CUT-3D, CycleGAN-3D and UNSB-3D each train their networks. Retrieval-RF matches each source scan again to the nearest target scan of that size and trains its flow inside PHANTESA's autoencoders of that size. DDIB trains nothing and runs on the autoencoders and flows that PHANTESA trains at that size. The pseudo-target's construction and the reading of each task are those of the main paper at every size, the rank map read by one projection at flow time 0.95 on T2, and the painted class levels read through the chain of three projections with re-entry on amyloid PET and perfusion. They are not chosen again, so a change in a score cannot come from a new choice of construction or reading. The validation sessions, both test sets and the scoring are the same at every size. PHANTESA at the middle size is the model that the rest of the paper reports.

\section{Benchmark results}
\label[suppsec]{sec:sup-benchmark}

\begin{table*}[t]
\centering
\small
\setlength{\tabcolsep}{3.4pt}
\caption{\textbf{Unpaired training data at three sizes, both test sets.} With more unpaired data PHANTESA's PSNR rises on amyloid PET and T2 on both test sets. Of its biomarkers, only the perfusion ratio on the test set rises. The sizes are a third of the subjects, all of them, and all of them with the additional subjects. Perfusion has no additional subjects. An asterisk marks a value that the tests separate from the next smaller size. DDIB has no perfusion biomarker, because its ratio is read on too few sessions. $^{\ddagger}$A failed run. The failed runs with all subjects are those of \cref{tab:benchmark}, and UNSB-3D's on amyloid PET has no Centiloid. UNSB-3D's run on T2 with the additional subjects also failed and has no scores (\cref{sec:sup-scaling}).}
\label{tab:sup-scaling}
\begin{tabular}{l ccc ccc ccc ccc}
\toprule
& \multicolumn{6}{c}{Test set} & \multicolumn{6}{c}{Independent test set} \\
\cmidrule(lr){2-7}\cmidrule(lr){8-13}
& \multicolumn{3}{c}{PSNR (dB)} & \multicolumn{3}{c}{Biomarker} & \multicolumn{3}{c}{PSNR (dB)} & \multicolumn{3}{c}{Biomarker} \\
\cmidrule(lr){2-4}\cmidrule(lr){5-7}\cmidrule(lr){8-10}\cmidrule(lr){11-13}
Model & 1/3 & All & +Add. & 1/3 & All & +Add. & 1/3 & All & +Add. & 1/3 & All & +Add. \\
\midrule
\multicolumn{13}{l}{\textit{Amyloid PET}, Centiloid $r$; 177 and 76 sessions} \\
\rowcolor{oursrow}PHANTESA & 22.69 & 24.75$^{*}$ & 25.03$^{*}$ & 0.11 & 0.10 & 0.07 & 23.38 & 25.82$^{*}$ & 25.71$^{*}$ & 0.22 & 0.08 & 0.13 \\
CUT-3D & 22.18 & 21.46$^{*}$ & 21.86$^{*}$ & 0.18 & 0.14 & 0.06$^{*}$ & 22.61 & 21.91$^{*}$ & 22.13$^{*}$ & $-$0.03 & 0.21$^{*}$ & 0.24 \\
Retrieval-RF & 19.59 & 18.78$^{*}$ & 19.10$^{*}$ & 0.07 & $-$0.04 & 0.05 & 20.43 & 19.50$^{*}$ & 19.60 & $-$0.05 & 0.26$^{*}$ & $-$0.02$^{*}$ \\
DDIB & 18.83 & 18.21$^{*}$ & 19.27$^{*}$ & $-$0.06 & 0.01 & 0.07 & 18.91 & 18.88 & 18.82 & 0.24 & 0.26 & 0.23 \\
CycleGAN-3D & 21.79 & 21.00$^{*}$ & 22.65$^{*}$ & 0.11 & 0.24$^{*}$ & 0.13 & 22.31 & 22.22 & 23.10$^{*}$ & 0.12 & 0.13 & 0.12 \\
UNSB-3D & 18.42 & 6.08$^{\ddagger\,*}$ & 22.00$^{*}$ & 0.14 & -- & 0.25 & 18.97 & 6.10$^{\ddagger\,*}$ & 22.69$^{*}$ & 0.10 & -- & 0.11 \\
\addlinespace[2pt]
\multicolumn{13}{l}{\textit{T2}, Tumor Dice; 60 and 60 sessions} \\
\rowcolor{oursrow}PHANTESA & 20.28 & 20.86$^{*}$ & 21.59$^{*}$ & 0.64 & 0.63 & 0.62 & 20.99 & 20.96 & 21.37$^{*}$ & 0.69 & 0.69 & 0.69 \\
CUT-3D & 16.40 & 15.52$^{*}$ & 16.36$^{*}$ & 0.44 & 0.30$^{*}$ & 0.40$^{*}$ & 16.08 & 15.39$^{*}$ & 15.84$^{*}$ & 0.53 & 0.44$^{*}$ & 0.43 \\
Retrieval-RF & 14.93 & 15.59$^{*}$ & 16.57$^{*}$ & 0.43 & 0.48$^{*}$ & 0.48 & 14.90 & 15.69$^{*}$ & 16.33$^{*}$ & 0.55 & 0.57 & 0.60$^{*}$ \\
DDIB & 14.02 & 12.91$^{*}$ & 14.05$^{*}$ & 0.46 & 0.49 & 0.46 & 13.48 & 12.77$^{*}$ & 14.24$^{*}$ & 0.47 & 0.51 & 0.53 \\
CycleGAN-3D & 15.41 & 13.77$^{*}$ & 17.16$^{*}$ & 0.54 & 0.47$^{*}$ & 0.56$^{*}$ & 15.16 & 13.63$^{*}$ & 17.40$^{*}$ & 0.65 & 0.58$^{*}$ & 0.61 \\
UNSB-3D & 15.99 & 15.44$^{*}$ & --$^{\ddagger}$ & 0.39 & 0.50$^{*}$ & -- & 16.08 & 15.33$^{*}$ & --$^{\ddagger}$ & 0.48 & 0.50 & -- \\
\textit{Real scan} &  &  &  & \multicolumn{3}{c}{0.67} &  &  &  & \multicolumn{3}{c}{0.73} \\
\addlinespace[2pt]
\multicolumn{13}{l}{\textit{Perfusion}, Ratio $r$; 49 and 48 sessions, the ratio read on 45 and 46} \\
\rowcolor{oursrow}PHANTESA & 18.46 & 18.23$^{*}$ & -- & 0.61 & 0.75$^{*}$ & -- & 18.68 & 18.70 & -- & 0.68 & 0.75 & -- \\
CUT-3D & 17.96 & 17.06$^{*}$ & -- & 0.83 & 0.85 & -- & 18.24 & 17.32$^{*}$ & -- & 0.78 & 0.82 & -- \\
Retrieval-RF & 13.40 & 13.73 & -- & $-$0.09 & $-$0.01 & -- & 13.62 & 14.06$^{*}$ & -- & 0.05 & 0.02 & -- \\
DDIB & 12.17 & 8.68$^{\ddagger\,*}$ & -- & -- & -- & -- & 11.86 & 7.31$^{\ddagger\,*}$ & -- & -- & -- & -- \\
CycleGAN-3D & 16.85 & 17.23$^{*}$ & -- & 0.65 & 0.81 & -- & 17.30 & 17.46$^{*}$ & -- & 0.74 & 0.81 & -- \\
UNSB-3D & 17.77 & 17.43$^{*}$ & -- & 0.80 & 0.77 & -- & 17.94 & 17.61$^{*}$ & -- & 0.77 & 0.63 & -- \\
\bottomrule
\end{tabular}
\end{table*}

\paragraph{PHANTESA has the highest PSNR among the unpaired models on both test sets.} The tests separate PHANTESA's PSNR from that of each compared unpaired model on every task, on the test set and on the independent test set (\cref{tab:benchmark}). On the test set it reaches 24.7\,dB against 21.5 for CUT-3D on amyloid PET, 20.9 against 15.6 for Retrieval-RF on T2 and 18.2 against 17.1 for CUT-3D on perfusion, each the best of the main paper's other unpaired models on its task. Taken before rounding, PHANTESA's lead over these three is 1.2 to 5.3\,dB. Of the two further models, CycleGAN-3D reaches 21.0\,dB on the amyloid PET test set, and UNSB-3D's failed run reaches 6.1\,dB (\cref{sec:sup-baselines}). On T2 CycleGAN-3D and UNSB-3D reach 13.8 and 15.4\,dB. On perfusion they reach 17.2 and 17.4\,dB, so PHANTESA's smallest lead there is 0.8\,dB. PHANTESA also has the highest SSIM among the unpaired models on amyloid PET and T2. On perfusion CUT-3D's SSIM is higher on both test sets, at 0.28 against 0.23 on the test set. CycleGAN-3D's and UNSB-3D's SSIM is 0.25 on the perfusion test set, where the tests separate it from PHANTESA's, and 0.26 and 0.25 on the independent test set, where they do not. Among the paired models, CoCoLIT lies below PHANTESA in PSNR on amyloid PET, at 24.5 against 24.7\,dB on the test set and at 24.9 against 25.8\,dB on the independent test set. CoCoLIT's training failed on T2 and on perfusion (\cref{sec:sup-baselines}), where its PSNR on the test set is 12.3 and 8.9\,dB against PHANTESA's 20.9 and 18.2\,dB. On these two tasks ResViT-2.5D and Pix2Pix-3D are therefore the only paired models that synthesize the target scan. By contrast, both have a higher PSNR than PHANTESA inside the brain mask on every task and both test sets, ResViT-2.5D on every session except 10 of the 76 of OASIS-3 (\cref{fig:sup-persession}), so PHANTESA stays below the best paired models. \Cref{fig:sup-qual-av45,fig:sup-qual-t2,fig:sup-qual-rcbv} show one session of each task for every model, in three views. A higher PSNR does not show whether a synthetic scan yields the real scan's biomarker value, which the last three paragraphs of this section examine.

\paragraph{Scored on the whole volume, every model gains and PHANTESA keeps its place.} Every model's PSNR rises on the whole volume, by 2.5 to 7.8\,dB, and its SSIM by 0.17 to 0.67, because its synthetic scan is nearly as empty outside the brain as the real one (\cref{tab:benchmark}). On the test set PHANTESA's PSNR rises by 4.4\,dB on amyloid PET, 4.3 on T2 and 7.6 on perfusion. The whole-volume PSNR leaves PHANTESA where the brain-mask PSNR puts it, below ResViT-2.5D and Pix2Pix-3D and above every unpaired model on every task and both test sets. Three other orders change on both test sets. On T2 CUT-3D and UNSB-3D each pass Retrieval-RF, and on perfusion DDIB passes CoCoLIT's failed run. On the amyloid PET test set CoCoLIT's distance below PHANTESA, taken before rounding, grows from 0.3 to 1.9\,dB. The whole volume thus adds a background that every model reproduces, and the brain-mask scores, which compare the synthetic brains alone, are the ones the paper reports.

\paragraph{The projection adds to its pseudo-target image on amyloid PET and perfusion, and little on T2.} The pseudo-target image can be scored as a prediction with no network, by the readers that score every model (\cref{sec:sup-pseudo-info}). On amyloid PET and perfusion it is painted from the class levels and holds no detail within a class. PHANTESA exceeds it there by 2.0 and 1.2\,dB on the test set, at 24.7 against 22.7\,dB and 18.2 against 17.0\,dB, and by 2.8 and 1.1\,dB on the independent test set, and its SSIM is 0.59 against 0.45 and 0.23 against 0.15 on the test set. The tests separate each of these differences. On perfusion the projection also places the high blood volume better. The overlap of the top tenth of rCBV rises from 0.18 for the image to 0.25 for PHANTESA on the test set and from 0.22 to 0.29 on the independent test set, and the tests separate both gains. On the test set the whole-tumor ratio's $r$ rises from 0.69 to 0.75 and the ratio's $r$ inside the enhancing tumor from 0.18 to 0.36, and on the independent test set they rise from 0.59 to 0.75 and from 0.15 to 0.47. On the independent test set the tests separate both gains. On the test set they separate neither, and the image alone already reaches $r=0.69$ on the whole tumor. On amyloid PET the image's Centiloid has a mean of $-26.2$ on ADNI and $-29.4$ on OASIS-3, which PHANTESA raises to $4.4$ and $-2.9$, about 30 Centiloid closer to the real scans' $44.4$ and $27.4$. Its $r$ is 0.03 for the image and 0.10 for PHANTESA on ADNI, and 0.15 and 0.08 on OASIS-3, and the tests separate neither pair. The projection therefore moves the level of the uptake toward the real one and does not improve the order of the subjects. On T2 the rank map keeps the source's detail in the target's contrast, and the image alone scores as high as PHANTESA, at 20.6 against 20.9\,dB on the test set and 21.0 against 21.0\,dB on the independent test set, with an SSIM of 0.46 against 0.45. With the image in the real T2's place the judge's whole-tumor Dice is 0.62 on the test set and 0.69 on the independent test set, against 0.63 and 0.69 for PHANTESA, and the tests separate neither pair. On the tumor core they separate a gain of 0.03 on the independent test set alone. The projection therefore adds most where the pseudo-target image leaves the detail unspecified, and on T2 the image already carries what the judge reads.

\paragraph{No model reads amyloid burden.} Every scored model's Centiloid $r$ lies between $-0.10$ and $0.29$ on both test sets, far from the 1 of a model whose Centiloid follows the real one on a line (\cref{tab:benchmark}; \cref{fig:sup-clinical} shows the test set). The highest is ResViT-2.5D's 0.29 on ADNI. Its agreement with the real amyloid burden is too weak to indicate that burden. UNSB-3D's failed run has no Centiloid (\cref{sec:sup-baselines}). On $r$ the tests separate no model from PHANTESA. The two nearest to a separation are CoCoLIT, whose $r$ lies below PHANTESA's, and CycleGAN-3D, whose 0.24 lies above it, both on ADNI with a corrected p-value of 0.05. Among the other pairs the tests separate CoCoLIT from ResViT-2.5D, from CUT-3D and from CycleGAN-3D, and Retrieval-RF from ResViT-2.5D, all on ADNI. The balanced accuracy of the staging lies between 0.19 and 0.35, near its chance level of 0.25, with one exception (\cref{tab:sup-secondary}). DDIB reaches 0.56 on OASIS-3, where the tests separate it from PHANTESA, CoCoLIT, ResViT-2.5D and CycleGAN-3D. Yet DDIB scores 0.19 on ADNI with the same cut points. On OASIS-3 its 0.56 rests on 7 of the 8 sessions of the intermediate range, and two of the four stages hold only 8 and 7 sessions. Its Centiloid follows the real one with $r$ of only 0.01 on ADNI and 0.26 on OASIS-3, so this one accuracy does not change the finding that no model reads amyloid burden. On ADNI the tests separate Pix2Pix-3D, at 0.35, from PHANTESA and from DDIB. PHANTESA's synthetic PET shows what its Centiloid carries instead. Over the sessions its Centiloid has a mean of $4.4$ and a standard deviation of 7.9 on ADNI, and $-2.9$ and 9.8 on OASIS-3, whereas the real scans have $44.4$ and $53.8$, and $27.4$ and $47.3$. Its synthetic scans therefore vary far less across subjects than the real ones, at the level of a subject without amyloid. The cut points fitted on its own validation predictions lie at $0.4$, $5.2$ and $7.3$ Centiloid, all below the lowest published cut point of 10~\cite{collij2024centiloid}, so three of its four stages lie wholly inside the range that the published scale calls free of amyloid pathology, and the fourth begins inside it. For every model, paired or unpaired, a synthetic amyloid PET therefore cannot stand in for the real one where the question is the amyloid burden.

\paragraph{On T2 the judge segments the whole tumor from PHANTESA's scan almost as well as from a paired model's.} With PHANTESA's synthetic T2 in place of the real one, the judge's whole-tumor Dice is 0.63 on the test set, against 0.67 with the real T2, 0.66 and 0.64 with the paired models' and 0.30 to 0.49 with the main paper's other unpaired models' (\cref{tab:benchmark}). The tests separate it from the real T2 and from ResViT-2.5D's, but not from Pix2Pix-3D's. CoCoLIT's failed run gives 0.55 on both test sets, which the tests separate from PHANTESA's 0.63 and 0.69. CoCoLIT's output is not a T2 (\cref{sec:sup-baselines}), yet the judge reads three real sequences beside it (\cref{sec:sup-verification}), so the 0.55 cannot be credited to a synthesized T2. PHANTESA's Dice is higher than CUT-3D's by 0.331 (95\% interval 0.279 to 0.385), than Retrieval-RF's by 0.151 (0.104 to 0.199) and than DDIB's by 0.143 (0.095 to 0.193). On the independent test set the differences are 0.258, 0.127 and 0.184, and each interval lies above zero. With CycleGAN-3D's T2 the judge reaches 0.47 on the test set and with UNSB-3D's 0.50, below PHANTESA's by 0.161 (0.120 to 0.205) and by 0.130 (0.082 to 0.181). On the independent test set it reaches 0.58 and 0.50, below PHANTESA's by 0.110 and 0.199, and each interval again lies above zero. There PHANTESA's whole-tumor Dice is lower than ResViT-2.5D's by 0.015 (interval $-0.031$ to $+0.002$) and than Pix2Pix-3D's by 0.010, and the tests separate it from neither. On the inner regions PHANTESA's Dice stays below the paired models' (\cref{tab:sup-secondary}). On the test set it is 0.46 against 0.51 on the tumor core and 0.56 against 0.61 and 0.62 on the enhancing tumor. DDIB reaches 0.51 and 0.58 there and UNSB-3D 0.51 and 0.60, and every model's Dice lies within 0.12 of the real scan's on both regions and both test sets. Even CoCoLIT's failed run reaches 0.50 and 0.61 on the test set, within 0.02 of the real scan's 0.52 and 0.63. The inner regions therefore tell the models apart less than the whole tumor does. On the independent test set these gaps narrow to 0.02 on the tumor core, with intervals below zero, and to at most 0.015 on the enhancing tumor. For segmenting the whole tumor, PHANTESA's synthetic T2 therefore comes almost as close to the real one as a paired model's, although the real T2 still scores higher and PHANTESA's PSNR is 5 to 7\,dB lower than the paired models' on the test set.

\paragraph{On perfusion most models follow the whole tumor's blood volume, and on the test set every model follows that of its enhancing part less closely.} Over the 45 sessions of the test set whose ratio is read, the correlation between the synthetic and the real tumor-to-mirror ratio is 0.75 for PHANTESA, 0.79 for ResViT-2.5D, 0.76 for Pix2Pix-3D and 0.85 for CUT-3D, and the tests separate none of the four (\cref{tab:benchmark}). On the independent test set PHANTESA's correlation is again 0.75, against 0.82 for ResViT-2.5D, 0.81 for Pix2Pix-3D and 0.82 for CUT-3D, and the tests again separate none of the four. The perfusion maps of PHANTESA and CUT-3D therefore rank the subjects by the blood volume of the whole tumor as the paired models' do. Retrieval-RF's maps do not rank the subjects, with a correlation of $-0.01$ and 0.02. CycleGAN-3D's correlation is 0.81 on both test sets and UNSB-3D's is 0.77 and 0.63, and the tests separate neither from PHANTESA. Among the other pairs the tests separate UNSB-3D from ResViT-2.5D and from CycleGAN-3D on the independent test set. The two models that fail on perfusion, CoCoLIT and DDIB, have no scored ratio (\cref{sec:sup-verification}). On the enhancing tumor alone the correlation on the test set is 0.63 for ResViT-2.5D and 0.56 for Pix2Pix-3D, against 0.37 for CUT-3D, 0.36 for PHANTESA, 0.24 for CycleGAN-3D, 0.21 for UNSB-3D and $-0.17$ for Retrieval-RF, each below the same model's correlation on the whole tumor (\cref{tab:sup-secondary}). Inside the enhancing tumor only the paired models therefore keep a correlation above 0.5 on the test set. On the independent test set three unpaired models come closer, with 0.53 for CUT-3D and 0.47 for PHANTESA and for CycleGAN-3D against 0.58 and 0.57 for the paired models, whereas UNSB-3D stays at 0.27. The paired bootstrap test separates the paired models from Retrieval-RF on both test sets and from none of PHANTESA, CUT-3D, CycleGAN-3D and UNSB-3D on either. ResViT-2.5D's lead over PHANTESA is 0.28 on the test set (95\% interval $-0.01$ to $0.59$) and 0.11 on the independent test set ($-0.08$ to $0.29$). The two sets are halves of one cohort, and the 95\% bootstrap interval of each model's difference between them includes zero, so the smaller gap on the independent test set lies within the variation between samples of 41 and 45 sessions. By the overlap of the top tenth of rCBV the paired models place the highest blood volume better. It is 0.49 and 0.43 for them, 0.31 for CUT-3D and 0.25 for PHANTESA on the test set, in the same order on the independent test set (\cref{tab:sup-secondary}). On the test set UNSB-3D's overlap is 0.30 and CycleGAN-3D's is 0.26.

\section{Scaling results}
\label[suppsec]{sec:sup-scaling}

\paragraph{PHANTESA's PSNR rises with more unpaired data on amyloid PET and T2.} On the test set it rises from 22.7\,dB with a third of the subjects to 24.7 with all of them and 25.0 with the additional subjects on amyloid PET, and from 20.3 to 20.9 and 21.6\,dB on T2, and the tests separate every step (\cref{tab:sup-scaling}; \cref{M-fig:scaling}). Each size is compared with the next smaller size of the same model on the same sessions, by the Wilcoxon test on PSNR and the paired bootstrap on a biomarker, corrected by Holm's procedure over the comparisons of one score on one test set (\cref{sec:sup-scoring}). On the independent test set its PSNR rises from a third to the largest size by 2.3\,dB on amyloid PET and by 0.4 on T2, yet one step of each does not rise. On amyloid PET the additional subjects lower it by 0.1\,dB, which the tests separate, as expected when the target scans move from OASIS-3 toward ADNI, and on T2 the step from a third to all subjects is level. On perfusion, which has two sizes, its PSNR falls from 18.5 to 18.2\,dB on the test set and is level on the independent test set. At each size PHANTESA trains every network again, so a change with the size includes the autoencoders'. With a third of the subjects the amyloid PET target autoencoder trains on 141 scans and the perfusion autoencoders on 62 and 61, and they reconstruct held-out scans less accurately than the autoencoders of the larger sizes.

\paragraph{Of PHANTESA's biomarkers only the perfusion ratio rises.} Its Centiloid $r$ lies between 0.07 and 0.22 at every size on both test sets, and the tests separate no step. The judge's whole-tumor Dice with its T2 is 0.64, 0.63 and 0.62 on the test set and 0.69 at every size on the independent test set. The tests separate no two neighboring sizes, and on the test set they separate the largest from the smallest, whose Dice is higher by 0.015. The ratio's $r$ on perfusion rises from 0.61 to 0.75 on the test set, which the tests separate, and from 0.68 to 0.75 on the independent test set, which they do not. More unpaired scans therefore raise one biomarker. They do not raise the Centiloid $r$, because structural MRI does not display the amyloid uptake at any size (\cref{sec:sup-benchmark}), nor the whole-tumor Dice, which lies within 0.05 of the real scan's at every size.

\paragraph{The main paper's other unpaired models show no consistent gain.} Of CUT-3D, Retrieval-RF and DDIB, only Retrieval-RF's PSNR on T2 rises at every step on both test sets, from 14.9 to 15.6 and 16.6\,dB on the test set and from 14.9 to 15.7 and 16.3\,dB on the independent test set. From the smallest to the largest size its Dice rises from 0.43 to 0.48 on the test set and from 0.55 to 0.60 on the independent test set. On amyloid PET CUT-3D and Retrieval-RF have their highest PSNR with a third of the subjects on both test sets, and on perfusion CUT-3D's PSNR falls by 0.9\,dB. DDIB trains nothing and runs on PHANTESA's flows of each size, and its scores follow no order in the size. On perfusion its PSNR falls from 12.2\,dB to the 8.7\,dB of its failed run with all subjects (\cref{tab:benchmark}), and its ratio is defined on too few sessions to be scored at either size.

\paragraph{CycleGAN-3D's PSNR is highest at the largest size.} On amyloid PET and T2 the tests separate CycleGAN-3D's PSNR at the largest size from its PSNR at both smaller sizes on both test sets. Its PSNR does not rise at every step. On the test set it falls from 21.8\,dB with a third of the subjects to 21.0\,dB with all of them on amyloid PET, and from 15.4 to 13.8\,dB on T2, and it rises to 22.7 and 17.2\,dB with the additional subjects. On the independent test set it is level at 22.3 and 22.2\,dB on amyloid PET and falls from 15.2 to 13.6\,dB on T2, and it rises to 23.1 and 17.4\,dB with the additional subjects. From the smallest to the largest size CycleGAN-3D's PSNR therefore rises by 0.9 and 0.8\,dB on amyloid PET and by 1.8 and 2.2\,dB on T2. On perfusion, which has two sizes, it rises by 0.4 and 0.2\,dB, and the tests separate the step on both test sets. CycleGAN-3D's biomarkers follow no order in the size. The judge's whole-tumor Dice with its T2 is 0.54, 0.47 and 0.56 on the test set and 0.65, 0.58 and 0.61 on the independent test set. The tests separate the largest size from the smallest on both test sets. The largest size's Dice is higher by 0.02 on the test set and lower by 0.04 on the independent test set. Its Centiloid $r$ is 0.11, 0.24 and 0.13 on the test set, where the tests separate the first step alone, and it lies between 0.12 and 0.13 on the independent test set. The $r$ of its perfusion ratio is 0.65 and 0.81 on the test set and 0.74 and 0.81 on the independent test set, and the tests separate neither pair.

\paragraph{UNSB-3D shows no consistent gain.} UNSB-3D runs at three sizes on amyloid PET and on T2 and at two sizes on perfusion, and two of these eight runs failed. On T2 and on perfusion neither the smallest nor the middle size is a failed run, and between these two sizes UNSB-3D's PSNR falls. On T2 it falls from 16.0 to 15.4\,dB on the test set and from 16.1 to 15.3\,dB on the independent test set, and on perfusion it falls by 0.3\,dB on both test sets. The tests separate each of these steps. The first failed run is UNSB-3D's benchmark run on amyloid PET (\cref{sec:sup-baselines}). That run is the middle size, at 6.1\,dB, so each step on amyloid PET includes a failed run. From the smallest to the largest size UNSB-3D's PSNR rises from 18.4 to 22.0\,dB on the test set and from 19.0 to 22.7\,dB on the independent test set, and the tests separate the two sizes. The second failed run is on T2 at the largest size. In that run the generator's output inside the brain was at the background value at every validation of the training, with a PSNR of about 6\,dB on the validation sessions. The run left no test predictions, so it has no scores (\cref{tab:sup-scaling}). The tests separate one step of UNSB-3D's biomarkers. From a third of the subjects to all of them the judge's whole-tumor Dice with its T2 rises from 0.39 to 0.50 on the test set, whereas on the independent test set the Dice is 0.48 and 0.50. UNSB-3D's Centiloid $r$ is 0.14 at the smallest size and 0.25 at the largest on the test set, and 0.10 and 0.11 on the independent test set. The $r$ of its perfusion ratio is 0.80 and 0.77 on the test set and 0.77 and 0.63 on the independent test set.

\end{document}